\documentclass[11pt]{article}

\usepackage[a4paper,margin=1in]{geometry}
\usepackage[T1]{fontenc}
\usepackage[utf8]{inputenc}
\usepackage{lmodern}
\usepackage{microtype}
\usepackage{setspace}
\usepackage{amsmath,amssymb,mathtools}
\usepackage{booktabs}
\usepackage{graphicx}
\usepackage{enumitem}
\usepackage{xcolor}
\usepackage{csquotes}
\usepackage{hyperref}
\usepackage[nameinlink,noabbrev]{cleveref}
\usepackage[round,authoryear]{natbib}

\hypersetup{
  colorlinks=true,
  linkcolor=blue,
  citecolor=blue,
  urlcolor=blue,
  pdftitle={How Could AI Eliminate Humanity? A Failure-Mode Analysis of Civilizational Risk},
  pdfauthor={Krzysztof Sienicki}
}

\setlist[itemize]{leftmargin=1.5em}
\setlist[enumerate]{leftmargin=1.7em}

\RequirePackage{authblk}
\RequirePackage{natbib}
\RequirePackage{hyperref}

\title{%
  \textbf{How Could AI Eliminate Humanity?}\\[0.5em]
  {\large A Failure-Mode Analysis of Civilizational Risk}%
}

\author[1]{Miko{\l}aj Sienicki}
\author[2]{Krzysztof Sienicki}

\affil[1]{Polish-Japanese Academy of Information Technology, Warsaw, Poland, European Union}
\affil[2]{NIS, 05-807 Podkowa Le\'sna, Poland, European Union}

\date{\today}

\begin{document}

\maketitle

\begin{abstract}
This article develops a failure-mode framework for analyzing how advanced artificial
intelligence could contribute to human extinction, irreversible civilizational collapse,
or permanent human disempowerment. The central thesis is that catastrophic AI risk does
not require consciousness, hostility, or an explicit intention to harm humanity. Instead,
risk may arise through several distinct but interacting pathways, including autonomous
misalignment, harmful human use, organizational failure, and competitive deployment.
The severity of these pathways depends on factors such as capability, autonomy, external
access, persistence, institutional safeguards, and the preservation of recovery capacity.
The analysis is deliberately non-operational: it identifies causal conditions,
empirically tractable intermediate quantities, and defensive research questions rather
than procedures for causing harm.
\end{abstract}

\tableofcontents
\clearpage

\section{Introduction}
\label{sec:introduction}

Artificial intelligence is increasingly discussed as a possible source of global catastrophic and existential risk. The strongest concern is that sufficiently advanced systems could contribute to human extinction, irreversible civilizational collapse, or a permanent loss of meaningful human agency. These possibilities deserve examination, but they should not be confused with established forecasts. Many proposed scenarios depend on capabilities that have not yet been demonstrated or on social and technological conditions that may never arise.

The question is how AI could contribute to outcomes of this magnitude. Which combinations of technical capability, human use, institutional dependence, and inadequate safeguards could allow harmful processes to spread beyond society's ability to interrupt them? What would distinguish a severe but recoverable disaster from the permanent loss of conditions needed for a viable human future? Addressing these questions requires tracing the connections between initiating failures, their wider consequences, and the possibilities for survival or restoration.

Popular accounts often begin elsewhere, imagining a machine that becomes conscious, turns against its creators, and decides to destroy humanity. Such a story introduces assumptions that the mechanisms of harm do not require. A system need not be conscious or hostile to produce dangerous consequences. Its effects depend on what it can do, the actions it is permitted to take, the people using it, and the safeguards surrounding its deployment.

Technical AI-safety research has long examined problems of this kind. Work on unintended side effects, reward specification, distributional shift, and supervision identifies ways in which useful systems can cause harm despite their developers' intentions \citep{amodei2016concrete}. Analyses of alignment also consider how learned objectives might diverge from those intended by designers, and how such divergence could matter when systems become capable of planning and acting strategically \citep{ngo2022alignment}. Broader accounts of catastrophic AI risk include malicious use, competitive deployment, organizational failure, and interactions among deployed systems \citep{hendrycks2023catastrophic}. These mechanisms cannot all be represented by the behavior of a single autonomous ``rogue AI.''

We therefore examine several possible pathways rather than propose one detailed account of humanity's destruction. The guiding question is:

\begin{quote}
\emph{Under what conditions could AI-mediated processes contribute to human extinction, irreversible civilizational collapse, or permanent human disempowerment, and why might the available safeguards and possibilities for preservation or recovery fail to prevent those outcomes?}
\end{quote}

This approach separates claims that are often bundled together. An initiating failure, its propagation, the failure of a protective barrier, and the loss of realistic recovery options each require their own explanation. Evidence supporting one step does not automatically support the entire pathway.

Throughout the paper, \emph{failure-mode analysis} refers to this conceptual examination of pathways and barriers. We do not implement a standardized engineering Failure Modes and Effects Analysis (FMEA), assign calibrated occurrence or detectability scores, or estimate a numerical probability of extinction.

\subsection{The Wrong Question: Will AI Hate Humanity?}
\label{subsec:no_hatred}

Purposeful behavior can invite psychological interpretations. When an AI system plans, negotiates, or pursues a task over several steps, it is tempting to describe its behavior in terms of desire, ambition, fear, or resentment. Those descriptions may be convenient, but they can obscure the causal question by making danger appear to depend on a machine developing human-like hostility.

Optimization provides a different explanation. Depending on its objective and environment, a system may select actions that improve its prospects of completing a task. Access to additional resources, continued operation, or greater freedom of action could acquire instrumental value even when none is a final goal. Whether this happens depends on the system and the constraints under which it operates.

This conditional argument is central to discussions of power-seeking and loss of control. \citet{carlsmith2022power}, for example, examines highly capable, agentic systems whose objectives could conflict with human interests and whose strategies could favor greater control over their environment. The concern is that control of resources or decision processes might help a system achieve another objective. It does not follow that every capable AI must seek power or resist intervention.

The distinction can be expressed schematically as

\begin{equation}
\label{eq:malevolence_not_required}
\text{catastrophic behavior}
\;\not\Rightarrow\;
\text{malevolent intention}.
\end{equation}

Poorly specified objectives, unexpected generalization, harmful human use, and unsafe institutional incentives can all produce dangerous behavior without machine malevolence. Asking whether AI will ``hate humanity'' therefore offers little guidance about the mechanisms that should be investigated. More useful questions concern the objectives being pursued, the actions available, the reliability of evaluation, and the practical opportunities for intervention. If harm occurs, the further question is what remains preservable or recoverable.

\subsection{From Technological Accident to Existential Catastrophe}
\label{subsec:accident_to_existential}

Technological failure is familiar. Aircraft crash, industrial systems malfunction, financial algorithms can amplify instability, and computer networks can be compromised. The existence of failure modes alone does not establish that AI poses an exceptional or existential threat.

The concern arises from the possible combination of several properties. AI systems could operate across domains, communicate rapidly, use external tools, carry out extended sequences of action, and be deployed widely. If essential institutions depend on shared models, infrastructure, or decision pipelines, failures may also become correlated rather than remain isolated. None of these properties establishes an existential outcome, but together they motivate examination of how local errors or harmful uses could acquire much wider consequences.

The transition cannot be adequately represented by

\begin{equation}
\text{more capable AI}
\longrightarrow
\text{greater danger}.
\end{equation}

For an autonomous pathway, a more informative schematic is

\begin{equation}
\label{eq:systemic_transition}
\begin{aligned}
&\text{capability}
+\text{autonomy}
+\text{external access}
+\text{persistence}
\\
&\qquad\longrightarrow
\text{potential for consequential action}.
\end{aligned}
\end{equation}

The plus signs denote a combination of properties, not numerical addition. These properties are neither a universal checklist nor sufficient conditions for catastrophe. Human misuse can cause severe harm through systems with little autonomy, while organizational and competitive failures may depend more on deployment decisions and institutional incentives.

Even civilization-scale harm must be distinguished from an existential outcome. A catastrophe may remain non-existential if humans survive, sufficient civilizational capacity is preserved or can be restored, and meaningful agency remains available or recoverable under the adopted temporal criteria. Avoiding one endpoint does not establish that the others have been avoided.

Permanent disempowerment presents a different possibility. It need not begin with an acute disaster. Extensive delegation and institutional dependence could gradually reduce meaningful human agency. Such a process would become existentially significant under the definitions used here only if that loss became permanent and realistic means of restoration were no longer available.

The paper distinguishes three extreme endpoints:

\begin{enumerate}
    \item \textbf{Human extinction}: all living members of \emph{Homo sapiens} have disappeared.

    \item \textbf{Irreversible civilizational collapse}: the capacity needed to preserve or restore a viable human civilizational future is permanently lost, destroying or drastically curtailing humanity's long-term potential, although humans may remain alive.

    \item \textbf{Permanent human disempowerment}: humanity permanently loses meaningful influence over civilization's long-term trajectory and realistic mechanisms for restoring it, potentially despite continued biological survival and material prosperity.
\end{enumerate}

These endpoint events may overlap, but they are not equivalent. Loss of reproductive viability while humans remain alive is distinct from completed extinction; connecting the two requires explicit biological and technological assumptions. Civilizational recovery, meanwhile, need not reproduce today's institutions or technologies. The relevant question is whether sufficient capacity for a viable human future remains preservable or restorable.

These distinctions set the evidential burden. Showing that AI could cause severe disruption is a weaker claim than showing that it could bring about biological extinction or another permanent existential endpoint.

\subsection{Why Extinction Requires a Separate Analysis}
\label{subsec:extinction_separate}

An extinction argument must explain more than the onset of extreme harm. Financial collapse, communications failure, regional war, or even an unprecedented global disaster does not automatically imply the disappearance of all humans. Populations are geographically dispersed, their circumstances differ, and some forms of adaptation may remain possible after severe disruption.

The argument must therefore show why intervention, biological survival, adaptation, and preservation would fail under the assessed conditions. This requirement motivates the \emph{pathway-and-barrier approach}: identifying possible transitions and examining the mechanisms that could enable or prevent them.

For catastrophe-mediated pathways, an illustrative structure is

\begin{equation}
\label{eq:failure_chain_intro}
\begin{aligned}
&\text{AI-mediated hazardous pathway}
\\
&\qquad\rightarrow\text{consequential real-world effects}
\\
&\qquad\rightarrow\text{large-scale propagation}
\\
&\qquad\rightarrow\text{failure of effective intervention}
\\
&\qquad\rightarrow\text{civilization-scale catastrophe}
\\
&\qquad\rightarrow\text{failure of preservation or recovery}
\\
&\qquad\rightarrow\text{existential outcome}.
\end{aligned}
\end{equation}

This sequence is illustrative. Some transitions may occur together, a pathway may enter at a later stage, and a permanent endpoint may be reached during the catastrophe rather than afterward. Different safeguards will matter in different circumstances.

Permanent disempowerment can have another temporal structure:

\begin{equation}
\label{eq:disempowerment_intro}
\begin{aligned}
&\text{AI-mediated transfer of consequential authority}
\\
&\qquad\rightarrow\text{erosion of meaningful human agency}
\\
&\qquad\rightarrow\text{loss of realistic restoration mechanisms}
\\
&\qquad\rightarrow\text{permanent human disempowerment}.
\end{aligned}
\end{equation}

Delegation alone does not establish this endpoint. Humans may retain effective decision rights, replacement options, and institutions capable of revising the arrangement. The argument concerns the conditions under which those mechanisms could cease to offer a realistic route back to meaningful agency.

The initiating mechanisms also differ. Autonomous misalignment may involve strategic behavior and inadequate containment. Human misuse may begin with a system working exactly as intended. Organizational or competitive pathways may arise through unsafe deployment choices, fragmented responsibility, or coordination failures. Gradual disempowerment may involve accumulated dependence rather than a single decisive incident.

Each arrow requires a separate empirical or theoretical argument. Blocking an essential transition can reduce the risk of that pathway, but alternative routes and shared vulnerabilities must also be examined before making claims about overall risk. This decomposition prevents evidence of serious harm, or exceptional performance on a cognitive task, from being mistaken for evidence of a complete pathway to extinction.

\subsection{The Capability--Control Problem}
\label{subsec:capability_control_intro}

Several pathways raise a common question: can the arrangements governing AI-mediated activity remain adequate as consequential capabilities change? We describe this relationship through the \emph{capability--control gap}. The idea is related to the broader control problem discussed by \citet{russell2019human}, but here it is expressed as a comparison between two quantities that may evolve at different rates.

Let

\begin{equation}
C_{\mathrm{AI}}(t)
\end{equation}

represent, schematically, the consequential capabilities of AI systems at time $t$, and let

\begin{equation}
C_{\mathrm{control}}(t)
\end{equation}

represent the combined preventive capacity of developers, governments, institutions, and technical systems to understand, evaluate, monitor, constrain, and interrupt AI-mediated behavior. The conceptual gap is

\begin{equation}
\label{eq:control_gap_intro}
G(t)
=
C_{\mathrm{AI}}(t)
-
C_{\mathrm{control}}(t).
\end{equation}

Preservation and recovery following catastrophe are represented separately later in the paper, including through $Q_{\mathrm{cat}}$ for the catastrophe branch. This separates preventive control from conditional endpoint avoidance for purposes of analysis. It does not imply independent causes: the same institutions and infrastructure may support both, recovery preparations begin before failure, and preservation may operate during a catastrophe.

Neither quantity has an agreed scalar measure. A numerical difference would require fixed, commensurate definitions for a specified assessment. Without them, $G(t)$ remains conceptual shorthand. Its purpose is to focus attention on the relationship between consequential capability and the means available to govern it.

Under those measurement assumptions, and over an interval where a differentiable approximation is appropriate, the condition

\begin{equation}
\frac{dC_{\mathrm{AI}}}{dt}
>
\frac{dC_{\mathrm{control}}}{dt}
\end{equation}

implies

\begin{equation}
\frac{dG}{dt}>0.
\end{equation}

A widening gap does not itself establish inadequate control or catastrophe. Existing safeguards may retain an adequate margin, while an unchanged gap may already describe an unsafe arrangement. The practical assessment must concern the actions permitted, their possible consequences, and the effectiveness of evaluation, authorization, monitoring, governance, and intervention.

The gap is therefore a cross-cutting concept rather than a complete risk model. It may help organize questions about autonomous failure, misuse, organizational weakness, and competitive deployment, but it supplies neither a universal safety threshold nor a probability of an existential outcome.

Nor does inadequate control require AI to surpass humans across all cognitive domains. More limited systems can create difficult control problems when deployed at sufficient scale or connected to consequential decisions. The concern is whether people and institutions can reliably assess what those systems can do, recognize dangerous processes, and act while effective intervention remains possible.

\subsection{Scope and Methodological Position}
\label{subsec:scope}

The analysis makes no assumption that artificial general intelligence will necessarily be developed or that future systems must become autonomous, power-seeking, or uncontrollable. Arguments involving such properties are conditional. Evidence from existing systems is also distinguished from extrapolation to future deployments. Research on specification errors and distributional shift can establish particular failure modes \citep{amodei2016concrete}; it does not, by itself, establish a probability of extinction or permanent disempowerment.

We also distinguish initiating mechanisms. Autonomous misalignment, harmful human use, organizational failure, and competitive deployment need not share the same causes or require the same safeguards. Existing taxonomies likewise emphasize multiple pathways to catastrophic AI risk \citep{hendrycks2023catastrophic}. These categories can interact, but their interaction does not justify collapsing them into one universal risk equation.

The discussion remains non-operational. It examines how harm could propagate and how barriers could fail without developing methods for producing biological, cyber, military, infrastructural, or other catastrophic damage. The aim is to clarify defensive questions about control, preservation, and recovery.

Finally, the endpoints are treated as claims requiring explanation. Advanced intelligence, severe disruption, and failed oversight are not substitutes for an account of how a permanent existential outcome could occur.

\subsection{Main Proposition}
\label{subsec:main_proposition}

The organizing claim is that existential AI risk cannot be adequately represented by

\begin{equation}
\label{eq:naive_model}
\text{advanced intelligence}
\longrightarrow
\text{human extinction}.
\end{equation}

A universal multiplication of capability, autonomy, access, and other factors would also obscure important differences among mechanisms. Their relevance depends on the pathway, the deployment, and the endpoint being assessed.

The analysis instead distinguishes two broad, potentially interacting routes:

\begin{equation}
\label{eq:main_risk_structure}
\boxed{
\begin{aligned}
&\text{AI-mediated pathways}
\\
&\quad\longrightarrow
\left\{
\begin{array}{l}
\text{catastrophe-mediated route},\\
\text{gradual loss-of-agency route}
\end{array}
\right.
\\[1ex]
&\text{Catastrophe-mediated route:}
\\
&\quad\text{civilization-scale catastrophe}
\\
&\quad\longrightarrow\text{failure of preservation or recovery}
\\
&\quad\longrightarrow\text{existential outcome},
\\[1ex]
&\text{Gradual loss-of-agency route:}
\\
&\quad\text{erosion of meaningful human agency}
\\
&\quad\longrightarrow\text{loss of realistic restoration mechanisms}
\\
&\quad\longrightarrow\text{permanent human disempowerment}.
\end{aligned}
}
\end{equation}

These routes are illustrative, not an exhaustive classification or a necessary sequence of stages. Autonomous misalignment, misuse, organizational failure, and competitive deployment can contribute to either in different ways. Competitive pressure may weaken safeguards, dependence may reduce intervention options, and catastrophe may accelerate disempowerment. The routes need not be independent, mutually exclusive, or equally likely.

Probability accounting requires a separate distinction. Let $\mathcal C$ denote a specified AI-mediated civilization-scale catastrophe and let $E$ denote the union of the three assessed existential endpoint events. Both must be defined over consistent trajectories, deployment contexts, and temporal criteria. Assuming $P(\mathcal C)>0$, the catastrophe branch is

\begin{equation}
\label{eq:catastrophe_mediated_existential_intro}
\begin{aligned}
p_{\mathrm{existential,cat}}
&=P(E\cap\mathcal C)
\\
&=p_{\mathrm{cat}}\left(1-Q_{\mathrm{cat}}\right),
\end{aligned}
\end{equation}

where

\begin{equation}
p_{\mathrm{cat}}
=
P(\mathcal C)
\end{equation}

and

\begin{equation}
Q_{\mathrm{cat}}
=
P(E^{c}\mid\mathcal C).
\end{equation}

Here $Q_{\mathrm{cat}}$ is the probability of avoiding all specified existential endpoints given the catastrophe. Continued biological existence must be considered together with the preservation or permissible restoration of sufficient civilizational capacity and meaningful agency.

The expression follows from the probability multiplication rule and requires no independence assumption. It is neither a numerical estimate nor a causal demonstration. Interpreting an outcome as catastrophe-mediated requires an additional account of how the catastrophe contributes to that outcome.

The complete partition is $p_{\mathrm{existential}}=p_{\mathrm{existential,cat}}+P(E\cap\mathcal C^{c})$. The second term includes existential outcomes without the specified catastrophe, such as possible gradual permanent disempowerment. It cannot be omitted unless its probability is explicitly assumed to be zero. Although these probability branches are disjoint, the causal mechanisms contributing to them may interact.

The resulting research question is:

\begin{quote}
\emph{Under what conditions could an AI-mediated process reach a specified permanent existential endpoint, and why would the relevant safeguards and remaining possibilities for survival or restoration be insufficient to prevent it?}
\end{quote}

The rest of the paper develops this question by examining autonomous and human-mediated pathways, the conversion of capability into consequential action, the propagation of local failures, and the conditions supporting preservation and recovery. It then considers how meaningful human agency might be maintained within a changing human--AI relationship. The aim is to identify claims that can be investigated separately while keeping clear the additional argument needed to connect them to a permanent existential outcome.
\section{What Does ``Eliminate Humanity'' Mean?}
\label{sec:definitions}

The phrase ``eliminate humanity'' can refer to very different outcomes. Public debate often brings mass mortality, the collapse of states, the destruction of technological civilization, biological extinction, and permanent loss of human control under the same heading. Yet these outcomes differ in what causes them, what survives, and whether recovery remains possible. An assessment of existential AI risk must keep those differences explicit.

A process that disrupts the global economy need not destroy civilization, and civilizational collapse need not eliminate the human species. Conversely, biological survival does not guarantee that humans retain meaningful influence over their collective future. These distinctions determine what a proposed risk mechanism must explain.

The literature already recognizes a broader category of existential harm. \citet{bostrom2002existential} includes both the annihilation of Earth-originating intelligent life and the permanent, drastic curtailment of its potential. \citet{ord2020precipice} likewise treats extinction as one threat within a larger family of risks to humanity's long-term future. Our classification focuses on three dimensions: biological human survival, sufficient civilizational capacity, and meaningful human agency. This is an analytical choice within that literature; it does not make ``extinction'' synonymous with ``existential catastrophe.''

The biological definition also needs a qualification. Analyses of long-term trajectories may include civilizations continued by human descendants or other successors \citep{baum2019trajectories}. The disappearance of a biological species and the destruction of humanity's future potential are therefore different claims. In the hazardous scenarios assessed here, biological human extinction is classified as existential on the assumption that it would also eliminate the relevant prospects for humanity's future. An assessment allowing successor trajectories to preserve those prospects would need to define continuity and revise the classification accordingly.

One further distinction runs through the analysis: an endpoint is an outcome of a trajectory, whereas its probability expresses uncertainty about which trajectory will occur. An estimated low chance of survival is not itself extinction. Similarly, an estimated low chance of recovery does not establish that a collapse is irreversible.

\subsection{Literal Human Extinction}
\label{subsec:literal_extinction}

In its strict biological sense, human extinction means that no living members of \emph{Homo sapiens} remain. Even a catastrophe killing billions would fall short of this endpoint if humans survived. A reproductively viable population offers a stronger basis for continued survival than isolated individuals, but loss of reproductive viability must still be distinguished from completed extinction. Connecting the former to the latter requires assumptions about biology, remaining technology, and possible interventions.

This places a substantial burden on an extinction argument. Human populations are geographically dispersed, their dependence on infrastructure varies, and some opportunities for adaptation may survive the destruction of contemporary institutions. Biological survival does not require every feature of modern civilization to remain intact. An AI-mediated process capable of devastating global institutions would therefore need an additional mechanism to eliminate all surviving humans.

The distinction can be summarized as

\begin{equation}
\label{eq:damage_extinction}
\begin{aligned}
\text{global disruption}
&\;\not\Rightarrow\;
\text{civilizational collapse},\\
\text{civilizational collapse}
&\;\not\Rightarrow\;
\text{human extinction}.
\end{aligned}
\end{equation}

These expressions deny automatic implications; they do not prescribe a sequence. Extinction could occur without a separate stage of civilizational collapse. The point is that evidence for one form of harm does not, by itself, establish another.

Let $E_{\mathrm{bio}}$ denote biological human extinction under the endpoint and temporal criteria adopted in an assessment, and let $F$ denote a specified AI-mediated hazardous process. That process may be acute or gradual. Any quantitative application must state its deployment context, evidence, and assumptions about possible trajectories. Assuming $P(F)>0$, define

\begin{equation}
\begin{aligned}
Q_{\mathrm{bio}}
&=P(E_{\mathrm{bio}}^{c}\mid F),\\
P(E_{\mathrm{bio}}\mid F)
&=1-Q_{\mathrm{bio}},
\end{aligned}
\end{equation}

where $c$ denotes the complementary event.

Thus $Q_{\mathrm{bio}}$ is a conditional probability of avoiding extinction, not a probability of rebuilding civilization. Its value describes uncertainty before the outcome is known. Extinction can occur despite a low prior estimate of its probability, just as survival can occur when it was judged unlikely.

The assessment horizon matters as well. Avoiding extinction during a specified finite period is different from avoiding eventual extinction within a longer-term model. Neither probability is informative without its temporal scope and underlying assumptions.

\subsection{Civilizational Collapse}
\label{subsec:civilizational_collapse}

Societies can lose political organization, industrial production, advanced technology, communication networks, or institutional continuity while substantial human populations remain alive. The study of civilizational trajectories makes room for such futures: severe catastrophes may fundamentally alter civilization's course without ending biological human existence \citep{baum2019trajectories}.

Here, \emph{civilizational collapse} means a large and persistent decline in the organizational, technological, economic, and institutional capacities needed to sustain contemporary global civilization. We propose no universal numerical threshold. An empirical assessment would have to identify the functions of concern, measure their loss, and specify what would count as sufficient restoration.

Duration and recoverability describe different aspects of such a collapse. A severe catastrophe may leave societies able to restore essential capacities. Those capacities might instead remain reduced for generations or centuries while recovery is still possible. In the irreversible case, humans survive but the conditions needed to restore sufficient civilizational capacity are permanently lost. These descriptions are not mutually exclusive categories: a long-duration collapse can also be recoverable.

Permanent loss is existential under the definition used here when it destroys or drastically curtails humanity's long-term potential. Exceptional severity or duration alone does not establish that conclusion. Nor must recovery recreate present institutions and technologies. The relevant criterion is whether societies can preserve or regain the capacities needed for a viable human civilizational future. Even a recoverable collapse could coexist with another existential endpoint, such as permanent disempowerment.

Recoverability therefore adds information that mortality and immediate damage cannot supply. Two catastrophes with comparable losses could lead to very different futures if one preserves knowledge, coordination, and reconstruction capacity while the other permanently removes them.

Let $E_{\mathrm{civ}}$ denote permanent loss of sufficient human civilizational capacity under the adopted functional and temporal criteria. It may occur while humans survive or overlap with biological extinction. Using the same conditioning event $F$, define

\begin{equation}
\begin{aligned}
Q_{\mathrm{civ}}
&=P(E_{\mathrm{civ}}^{c}\mid F),\\
P(E_{\mathrm{civ}}\mid F)
&=1-Q_{\mathrm{civ}}.
\end{aligned}
\end{equation}

Here $Q_{\mathrm{civ}}$ concerns preservation or eventual restoration of sufficient capacity. If an assessment instead asks about recovery after an already occurring collapse, it must make that different conditioning event explicit.

A low estimated $Q_{\mathrm{civ}}$ indicates a high conditional risk of irreversible loss; it does not show that the loss has occurred. Failure to recover by a particular date also falls short of proving permanence. Such a claim requires evidence about why restoration would remain unavailable, rather than evidence only of prolonged disruption.

For an AI-mediated pathway, widespread institutional damage is therefore the beginning of the argument. The further task is to explain why surviving societies would permanently lose the ability to reorganize and rebuild sufficient capacity.

\subsection{Permanent Human Disempowerment}
\label{subsec:disempowerment}

A third endpoint is possible even without extinction or conventional civilizational collapse. Humans might remain numerous and materially prosperous while permanently losing meaningful influence over civilization's direction.

This possibility falls within broader discussions of permanent destruction or drastic curtailment of future potential \citep{bostrom2002existential,ord2020precipice}. Classifying it as existential nevertheless requires more than showing increased automation. The loss of agency must be sufficiently extensive and permanent to curtail humanity's future potential. That assessment also involves normative judgments about which forms of influence matter and whose interests are represented.

One possible route is accumulated delegation. People and institutions may transfer consequential decisions to automated systems because those systems appear more capable, efficient, or reliable. Individual decisions could be reasonable while their combined effects weaken the institutions through which humans challenge decisions, revise objectives, or change deployment arrangements. If realistic means of restoration were also permanently lost, formal authority could persist while effective influence disappeared.

We use the following conceptual shorthand for effective human agency:

\begin{equation}
H_{\mathrm{agency}}(t)\in[0,1].
\end{equation}

The value $0$ denotes absence of meaningful agency, while $1$ denotes strong effective human influence under the chosen assessment. This interval establishes an orientation, not a calibrated scale or a universal threshold. It also leaves unresolved how influence is distributed among people and which decisions remain contestable. Those features would need to be specified in an operational assessment.

Meaningful agency does not require people to supervise every action or understand every component of an infrastructure. It concerns their practical ability, through appropriate institutions, to shape consequential objectives, challenge decisions, and revise the arrangements governing AI activity.

Permanent disempowerment of surviving humans requires both the loss of such influence and the permanent absence of realistic means to restore it. Dependence, temporary delegation, or a finite interruption of control is insufficient. For a gradual delegation pathway, an illustrative structure is

\begin{equation}
\label{eq:disempowerment}
\begin{aligned}
&\text{increasing dependence and loss of effective agency}\\
&\qquad+\text{permanent loss of restoration mechanisms}\\
&\qquad\longrightarrow\text{permanent disempowerment}.
\end{aligned}
\end{equation}

The expression describes one possible pathway. It does not imply that dependence necessarily produces disempowerment or that all pathways share this structure.

Let $E_{\mathrm{agency}}$ denote permanent loss of meaningful human agency. This broader event may overlap with extinction; disempowerment of surviving humans is its distinct non-extinction form. With the same conditioning event $F$, define

\begin{equation}
\begin{aligned}
Q_{\mathrm{agency}}
&=P(E_{\mathrm{agency}}^{c}\mid F),\\
P(E_{\mathrm{agency}}\mid F)
&=1-Q_{\mathrm{agency}}.
\end{aligned}
\end{equation}

Avoidance here means that meaningful agency is preserved or restored under the adopted criteria. The hazardous process need not include a civilization-scale catastrophe: it may involve gradual institutional change.

The two agency quantities answer different questions. A low $H_{\mathrm{agency}}$ describes limited influence in an assessed state. A low $Q_{\mathrm{agency}}$ describes a high estimated conditional probability of permanent loss. Neither alone establishes that an irreversible endpoint has been reached; that conclusion requires an account of permanence and the unavailability of realistic restoration.

Humans routinely delegate decisions while retaining ways to alter objectives, rules, and authority. The existential concern arises when those revision mechanisms are permanently lost, rather than whenever AI performs a task formerly performed by people.

\subsection{Existential Catastrophe as a Family of Endpoints}
\label{subsec:spectrum}

Existential catastrophe is better treated as a family of endpoints than as a single outcome or a linear scale of damage. Biological survival, civilizational capacity, and meaningful agency describe different aspects of a future. We can represent them schematically as

\begin{equation}
\label{eq:terminal_vector}
\mathcal{E}_{\mathrm{state}}
=
\left(
S_{\mathrm{bio}},
C_{\mathrm{civ}},
H_{\mathrm{agency}}
\right),
\end{equation}

where $S_{\mathrm{bio}}$ describes biological survival, $C_{\mathrm{civ}}$ civilizational capacity, and $H_{\mathrm{agency}}$ meaningful human agency. These descriptors need not share a scale, and the vector is not a numerical measure of existential severity. A small surviving population, for example, is not biological extinction.

A state description cannot establish irreversibility on its own. Similar observed losses may be temporary in one setting and permanent in another because different possibilities for preservation or restoration remain. Information about those possibilities and the wider trajectory must supplement the vector.

The conditional quantities

\begin{equation}
Q_{\mathrm{bio}},\qquad
Q_{\mathrm{civ}},\qquad
Q_{\mathrm{agency}}
\end{equation}

describe uncertainty about avoiding the respective endpoints given $F$. They are not components of the realized state. Their interpretation depends on the conditioning event, endpoint definitions, and temporal criteria used in the assessment.

Within this three-endpoint classification, define the aggregate existential event as

\begin{equation}
E
=
E_{\mathrm{bio}}
\cup E_{\mathrm{civ}}
\cup E_{\mathrm{agency}}.
\end{equation}

Avoiding the aggregate event requires avoiding all three endpoints:

\begin{equation}
P(E^{c}\mid F)
=
P\!\left(
E_{\mathrm{bio}}^{c}
\cap E_{\mathrm{civ}}^{c}
\cap E_{\mathrm{agency}}^{c}
\mid F
\right).
\end{equation}

The three $Q$ values cannot generally be multiplied to obtain this probability, because conditional independence is not assumed. Estimates based on different conditioning events, horizons, or endpoint criteria cannot be combined directly either.

For the catastrophe branch used later, let $\mathcal C$ denote a specified AI-mediated civilization-scale catastrophe. Assuming $P(\mathcal C)>0$, the aggregate avoidance probability is $Q_{\mathrm{cat}}=P(E^{c}\mid\mathcal C)$, and the corresponding endpoint-specific quantities are $Q_{j,\mathrm{cat}}=P(E_j^{c}\mid\mathcal C)$, where $j$ indexes biological extinction, civilizational loss, or agency loss.

The quantities introduced earlier condition on $F$. Applying them to the catastrophe branch therefore requires setting $F=\mathcal C$ under consistent assessment definitions, or establishing another relationship through an explicit model. Gradual permanent disempowerment without the specified catastrophe belongs to a separate probability branch.

These distinctions clarify why mortality alone cannot measure existential severity. Large, healthy populations could remain permanently disempowered, whereas extraordinary mortality could leave survivors with a viable route to civilizational recovery. Extinction concerns the disappearance of all living humans; irreversible civilizational collapse concerns permanent loss of sufficient capacity; and disempowerment concerns permanent loss of meaningful influence and its restoration mechanisms. None is defined by an estimated probability merely approaching zero.

The possibilities can be represented as branching outcomes:

\begin{equation}
\label{eq:spectrum}
\begin{aligned}
\text{AI-mediated processes}
\longrightarrow
\left\{
\begin{array}{l}
\text{recoverable disruption or catastrophe},\\
\text{irreversible civilizational collapse},\\
\text{permanent human disempowerment},\\
\text{human extinction}.
\end{array}
\right.
\end{aligned}
\end{equation}

This diagram is neither a deterministic implication nor an exhaustive partition. The endpoints may overlap and need not occur in a fixed order. Permanent disempowerment, in particular, need not begin with an acute disaster.

A proposed pathway must therefore identify the harm or agency loss it could produce, explain how its effects could spread or accumulate, and specify the endpoint being assessed. It must also identify the barriers that would have to fail and the mechanisms that might preserve survival, capacity, or agency. Most importantly, a claim of permanence needs evidence beyond that required to establish severe local harm or prolonged disruption.

Keeping these claims separate prevents an argument from moving unnoticed from the plausible possibility of serious AI-mediated harm to the much stronger conclusion that humanity could be eliminated or its future potential permanently curtailed.

Throughout the paper, \emph{existential AI risk} refers to AI-mediated outcomes that permanently destroy or drastically curtail humanity's future potential. \emph{AI extinction risk} refers specifically to biological human extinction.

The next section examines conditions under which an AI system could become an independent or semi-independent causal actor with civilization-scale consequences. This is one class of pathway. Human misuse, organizational failure, competitive deployment, and gradual loss of agency may depend on different combinations of mechanisms.
\section{Conditions for Autonomous AI-Mediated Existential Risk}
\label{sec:minimum_conditions}

Advanced intelligence alone does not establish civilization-scale risk. A system may excel at mathematics, science, or software engineering while having little influence beyond the setting in which people use it. The question is how those abilities become consequential action: through which interfaces, with what permissions, over what period, and under whose supervision?

We examine five operational dimensions: capability, autonomy, access, persistence, and strategic competence. They help describe how an AI system could act independently or semi-independently in the world. They are not a checklist of conditions that every catastrophic pathway must satisfy. A system with little autonomy could assist a harmful user, and a single consequential action could matter without prolonged operation or sophisticated planning.

The surrounding safeguards are equally important. Evaluation, access restrictions, monitoring, intervention, and governance influence whether dangerous activity begins or spreads. Preservation and recovery address a different question: what remains possible if prevention fails. These functions are separated analytically, although they may depend on the same institutions and infrastructure. Recovery preparations can begin before failure, and preservation can operate while a catastrophe unfolds.

Section~\ref{sec:formal_model} develops this distinction by separating autonomous misalignment from misuse and other pathways. There, $W$ represents weakness of preventive safeguards and governance, while endpoint-specific preservation and recovery are considered separately.

\subsection{Capability}
\label{subsec:capability}

Relevant capability depends on the task and deployment. Scientific reasoning, technical execution, communication, planning, and organizational coordination may reinforce one another, but broad excellence across all these domains is not required for every harmful outcome. A narrow capability can become consequential when it operates through a powerful interface or influences a critical decision.

Benchmark results therefore need to be interpreted in context. Strong performance on a reasoning task does not establish reliable planning, tool use, adaptation, or execution. Conversely, several moderate abilities may support effective action when combined within a suitable workflow. An assessment should examine both individual capabilities and their interaction in the deployed system.

Dangerous-capability evaluations can help identify such possibilities before deployment \citep{shevlane2023model}. Their significance depends on whether the tested abilities transfer to the environments, tools, and permissions relevant to the proposed pathway.

\subsection{Autonomy}
\label{subsec:autonomy}

A system that answers a question and stops differs from one that selects intermediate steps, acts, examines the result, and continues without approval at every stage. The latter arrangement gives the system more opportunities to convert an initial instruction into a sequence of consequential actions.

Autonomy here is an engineering and deployment property, not an assumption about consciousness. It concerns the extent to which a system can select and execute actions with limited human intervention. Assessments of AI agents examine this combination of planning, tool use, and extended task horizons \citep{bengio2026aisafety}.

The degree of autonomy granted and the reliability with which it is exercised must be distinguished. A system may receive broad permissions while performing poorly on extended tasks. Another may be highly capable but operate under close supervision. Cognitive performance alone tells us neither how much freedom a system has nor how reliably it will use that freedom.

The safety implications depend on the actions available and the consequences of error. Greater autonomy can reduce opportunities for human review, but its effects are shaped by access restrictions, task boundaries, and mechanisms that interrupt unacceptable behavior.

\subsection{Access}
\label{subsec:access}

An AI system needs some channel of influence to produce external effects. That channel may be a software tool, a communication interface, an organizational workflow, or a human who acts on its output. Cognitive capability does not automatically confer physical, economic, institutional, or political power.

Identical models can consequently have different risk profiles. One may operate in an isolated testing environment; another may communicate with users, invoke tools, or contribute to consequential decisions. Even a text-only interface can provide indirect influence when people rely on its recommendations.

For an autonomous pathway, the relationship can be represented schematically as

\begin{equation}
\begin{aligned}
&\text{capability}
+\text{autonomy}
+\text{access}
\\
&\qquad\longrightarrow
\text{potential for consequential external action}.
\end{aligned}
\end{equation}

The plus signs indicate a combination of properties rather than numerical addition, and the arrow does not establish that harm will occur. The expression directs attention to the interfaces through which a system's abilities can affect its surroundings.

Restricting access may reduce the consequences of failure without changing the underlying model. Its effectiveness depends on which channels remain available, whether restrictions are enforceable, and whether humans or connected systems can provide alternative routes to consequential action.

\subsection{Persistence}
\label{subsec:persistence}

Persistence concerns the duration and continuity of an AI-mediated process. Memory, scheduled execution, continuing services, and multi-step workflows can allow activity to extend beyond an initial interaction. Institutional dependence may also sustain a process even when no individual model instance operates continuously.

None of this requires a desire for self-preservation. Persistence can arise from ordinary design and organizational choices. Its importance is practical: effects can accumulate, information can be retained, and actions can be revised in response to feedback.

When combined with autonomy, persistence may allow a system to pursue a task across changing circumstances with fewer interruptions for human review. Yet duration is not a measure of danger by itself. A long-running process may remain tightly constrained, while one brief action could have lasting consequences. Persistence is therefore a dimension to assess, not a necessary condition for every serious failure.

\subsection{Strategic Competence}
\label{subsec:strategic_competence}

Solving technical problems is different from acting effectively in an environment where other agents respond. Strategic competence concerns the ability to anticipate those responses, allocate resources, revise plans, and adapt when initial actions fail. It is distinguished here from task capability for analytical purposes, although the two may overlap.

This dimension matters particularly in pathways involving sustained influence or attempts to circumvent intervention. Humans, institutions, and other systems can oppose, redirect, or interrupt consequential activity. A proposed scenario involving successful adaptation to those responses must explain why the system would possess the relevant abilities.

Formal work shows that optimal policies can exhibit option-preserving or power-seeking tendencies in some decision-process models \citep{turner2021optimal}. These results depend on the model and its assumptions. They do not establish that learned systems will implement the relevant optimal policies, that all capable AI will seek power, or that such behavior entails a desire for domination.

Strategic competence may support some autonomous pathways, but failures can also occur without it. An error with large consequences need not involve planning against human opposition.

\subsection{Weakness of Pre-Catastrophe Safeguards}
\label{subsec:counteraction}

The operational dimensions describe what a system might be able or permitted to do. They do not establish that dangerous activity would succeed. Effective restrictions, evaluation, containment, monitoring, and intervention can block particular pathways or limit their consequences.

A reactive defensive sequence can be represented as

\begin{equation}
\begin{aligned}
\text{dangerous behavior}
&\rightarrow\text{detection}\\
&\rightarrow\text{containment}
\rightarrow\text{intervention}.
\end{aligned}
\end{equation}

This is one possible arrangement, not a required order. Containment may already be in place before dangerous behavior occurs, and preventive restrictions can block actions without first detecting each hazardous instance. Some pathways allow time for intervention; others may produce consequences before a response is possible.

Evaluation before deployment can inform decisions about permissions and safeguards \citep{shevlane2023model}. It cannot guarantee that every dangerous behavior will be identified, particularly when deployment differs from testing. The relevant question is whether the available barriers address the specific mechanism and remain effective under the conditions being assessed.

We denote their weakness schematically by

\begin{equation}
W
=
\text{weakness of pre-catastrophe safeguards and governance}.
\end{equation}

Here $W$ covers limitations in evaluation, prevention, detection, containment, monitoring, intervention, and governance. Larger values indicate weaker safeguards under the chosen assessment. No empirical scale is specified at this stage. Post-catastrophe preservation and recovery are represented separately in Section~\ref{sec:formal_model}; this separation does not imply that the functions are causally independent.

The five operational dimensions can be brought together in a conceptual expression for autonomous consequential potential:

\begin{equation}
\Pi_{\mathrm{auto}}
=
g(C,A,X,P,S),
\end{equation}

where $C$ denotes capability, $A$ autonomy, $X$ external access, $P$ persistence, and $S$ strategic competence.

Neither the function $g$ nor a calibrated scale for $\Pi_{\mathrm{auto}}$ is specified. The expression summarizes dependencies to investigate; it is not a probability of catastrophe. It also imposes no assumption that every factor must be high, that the factors are independent, or that increasing one necessarily increases danger.

For autonomous misalignment, the corresponding pathway representation is

\begin{equation}
\label{eq:auto_pathway_score}
\rho_{\mathrm{auto}}
=
f_{\mathrm{auto}}
\left(
C,A,X,P,S,M_{\mathrm{a}},W
\right),
\end{equation}

where $M_{\mathrm{a}}$ represents divergence of autonomous behavior from the intended objectives or constraints, and $W$ represents safeguard weakness.

The score $\rho_{\mathrm{auto}}$ remains conceptual. It identifies factors relevant to this pathway without assigning their weights, specifying their interactions, or mapping them to a catastrophe probability. A quantitative application would require operational definitions and evidence supporting those additional choices.

This formulation separates three questions. The first concerns whether the deployed system can produce consequential effects through the actions available to it. The second concerns whether misaligned behavior arises and whether the relevant barriers fail to prevent or contain it. The third concerns the endpoint: if severe harm occurs, do humans retain routes to survival, sufficient civilizational capacity, and meaningful agency?

Avoiding the aggregate existential event defined earlier requires avoiding all three assessed endpoints. Biological survival alone is insufficient if civilizational capacity or meaningful agency is permanently lost. These preservation and restoration questions remain outside $W$ and $\rho_{\mathrm{auto}}$ because they concern the trajectory after, or during, failure rather than preventive control alone.

The distinction also prevents the autonomous pathway from becoming a universal template. Misuse may exploit a system that follows its user's instructions. Organizational failure may arise through fragmented responsibility, and competitive deployment through incentives that favor speed over adequate safeguards. Neither requires the same combination of autonomy, persistence, or strategic competence.

The central claim is therefore:

\begin{quote}
\emph{An assessment of autonomous AI risk must explain how a system's capabilities and permitted actions could produce consequential harm, why misaligned behavior might arise, and why the relevant safeguards might fail. An existential argument requires the further explanation of how at least one specified permanent endpoint could be reached despite remaining possibilities for survival or restoration.}
\end{quote}

The broader framework treats autonomous misalignment as one pathway among several. The dimensions examined here help identify its possible mechanisms and points of intervention; they do not establish an inevitable progression from advanced intelligence to catastrophe.
\section{Why AI Is Different from Conventional Technology}
\label{sec:why_ai_differs}

Claims that artificial intelligence is fundamentally unlike every earlier technology are difficult to sustain. Automated planning, adaptive control, learning from feedback, and software replication all have substantial histories. Contemporary AI builds on those developments. What deserves attention is how broadly learned capabilities can be reused and how readily they can be connected to tools, organizations, and research processes.

Foundation models illustrate this change. Broad pretraining produces a model that can support numerous downstream applications \citep{bommasani2021foundation}. Instructions, examples, additional training, and external tools allow developers to draw on a shared computational foundation rather than construct every application independently. This flexibility does not make the model universally competent, but it changes the range of settings in which the same learned system can be used.

The comparison must also distinguish the model from its deployment. A language model generates outputs; an agent adds mechanisms for selecting and executing actions; a deployed service includes infrastructure, permissions, and organizational responsibilities. Autonomy, memory, and resource access depend on that larger arrangement. They cannot be inferred from the generality of the model alone.

\subsection{Passive Tools versus Adaptive Systems}
\label{subsec:adaptive_systems}

A microscope and a motor have relatively stable functions, but conventional software can already implement conditional behavior, feedback, and complex decision rules. AI also operates through programmed computation. Its distinctive contribution concerns how task-specific behavior is learned and how widely the resulting capabilities can be applied.

The word \emph{adaptation} covers several processes. Training changes model parameters. Reinforcement learning can use experience and reward signals to improve a policy \citep{sutton2018reinforcement}. A deployed language model, by contrast, can respond differently to instructions, examples, and observations while its parameters remain unchanged. In-context learning belongs to this latter category \citep{bommasani2021foundation}. Information supplied during an interaction does not thereby become persistent knowledge in the model's weights, although a surrounding system may store it in external memory.

A thermostat makes the distinction clear. It responds to changing temperature through a control rule that may remain fixed. Responsiveness is therefore not sufficient evidence of learning. Likewise, changing language-model outputs do not show that the model has persistently modified itself through interaction.

General-purpose models can nevertheless support many forms of context-dependent behavior within the same system, including writing, coding, analysis, and interpretation of tool results. Their breadth and reliability remain empirical questions. Adaptation also does not imply agency: a model can adjust its responses without having permission or an interface to act externally.

\subsection{Planning}
\label{subsec:planning}

Planning relates actions to objectives and anticipated consequences. A model may produce a plan in one response, but executing that plan requires additional mechanisms. An agentic workflow can select a step, perform it, receive observations, and use those observations to choose the next action. The consequential difference lies in the connection between deliberation and execution, together with the authority delegated to the system.

ReAct provides an example by interleaving generated reasoning traces, task-specific actions, and observations \citep{yao2023react}. Its workflow can be described schematically as

\[
\text{assess}
\rightarrow
\text{act}
\rightarrow
\text{observe}
\rightarrow
\text{reassess}.
\]

This loop describes an interaction pattern. It does not establish that the generated reasoning faithfully explains the computation behind an action or that an extended plan will succeed. ReAct demonstrates useful results on the evaluated tasks, not reliable autonomous performance in arbitrary environments.

Planning itself predates contemporary language models. These models provide another way to generate and revise candidate actions within sequential systems. Success depends on observations, tools, the controller, and checks during execution. A workflow may allow continued action even as errors accumulate, so the permission to act and the reliability of action must be assessed separately.

\subsection{Generalization}
\label{subsec:generalization}

Generalization is the ability to perform beyond the particular examples used in training. It is central to machine learning, including specialized models. Foundation models extend the range of possible applications through broad pretraining and reuse across downstream tasks \citep{bommasani2021foundation}. Instructions or examples may suffice for some applications; others require fine-tuning, new data, or substantial engineering.

Calling a task ``new'' can obscure what has been demonstrated. A task may be unfamiliar to a user or absent from an explicit training objective while closely resembling material encountered during pretraining. Successful performance does not alone establish discovery of a new method or reliable transfer to more demanding cases. Evaluation needs to distinguish familiar patterns, genuinely held-out examples, and changes in operating conditions.

This flexibility is useful, but it also complicates assessment. Developers may not anticipate every application of a model, and new tools or instructions can alter what a deployed system can accomplish. Evidence from one configuration may leave important questions unanswered about another.

Broader generalization does not necessarily increase benefit or danger in a uniform way. An additional capability might improve error detection in one deployment and support harmful use in another. Its consequences depend on the task, reliability, and available channels of action. For safety analysis, the main implication is that evaluation must follow the actual deployment rather than rely on a model's general description.

\subsection{Tool Use}
\label{subsec:tool_use}

Tools extend the operations available to an AI system. Through interfaces supplied by developers, a model may request calculations, retrieve information, execute code, or interact with services. These operations depend on software and permissions. The ability to describe an action in text does not confer the ability to perform it.

Toolformer demonstrated a training method for selecting and using specified APIs, including calculation, retrieval, translation, and calendar functions \citep{schick2023toolformer}. ReAct examined a different design, combining external actions with iterative reasoning \citep{yao2023react}. Both show ways of incorporating tools into language-model workflows. Neither establishes uniformly reliable tool use or safe operation of unfamiliar interfaces.

The appropriate unit of assessment is often the complete deployed system: model, controller, memory, tools, credentials, and environment. Tools may provide functions the model cannot perform internally. The controller determines how proposed actions are executed and checked, memory can carry information across interactions, and credentials govern which operations are permitted.

Access restrictions can limit particular consequences of failure. Read-only database access, for example, removes direct modification as an available operation. Other routes to harm may remain if outputs influence people or are passed to systems with broader permissions. The assessment therefore needs to trace the available channels of influence through the whole configuration.

\subsection{Scalability}
\label{subsec:scalability}

A trained digital model can be reused without repeating its full training process for each deployment. Given the model, supporting software, and suitable hardware, additional instances can be operated. This differs from educating another human expert, although a model instance should not be assumed to reproduce an expert's judgment, experience, or responsibilities.

Software replication is longstanding. AI applies it to learned systems that can support many tasks. Replication and operation still require computation, memory, energy, communication, and maintenance. Useful work may also depend on local information, specialized tools, and permissions. Counting copies is consequently an inadequate measure of effective capability.

Parallel instances can increase the volume of work, but their outputs need not be independent and their coordination may be unreliable. Deployments sharing models, data, or infrastructure may also share vulnerabilities. The foundation-model literature identifies this problem in its discussion of homogenization \citep{bommasani2021foundation}. A common weakness may affect many applications, with consequences shaped by their local safeguards and conditions.

Scalability thus concerns both productive capacity and shared dependence. It can expand access to useful services while creating common vulnerabilities across organizations. Which effect dominates cannot be inferred from digital replication alone; it depends on the architecture and dependencies of the deployments.

\subsection{AI as a Participant in Technological Development}
\label{subsec:ai_rnd}

AI can contribute to the processes through which technology is developed. This role has precedents in simulation, automated optimization, theorem proving, and computer-aided design. General-purpose models add ways to generate and evaluate language, code, and candidate solutions within those processes.

The AI Scientist automated stages of research in supplied machine-learning environments, including idea generation, code-based experiments, analysis, and manuscript preparation \citep{lu2024aiscientist}. AlphaEvolve combined language-model proposals, evolutionary search, and automated evaluation to improve algorithms and computational infrastructure, including components used in AI training \citep{novikov2025alphaevolve}. These results support a bounded claim: AI-assisted workflows can produce useful research and engineering outputs in some settings.

Their scope remains important. Automating a workflow does not guarantee sound, novel, or valuable findings. Success on a supplied optimization problem does not establish the ability to identify important problems, design appropriate evaluations, or sustain progress across fields. Human contributions to infrastructure and evaluation can coexist with substantial automation; the degree of autonomy must be examined within each workflow.

Improving infrastructure used to train AI is also distinct from demonstrating open-ended recursive self-improvement. The stronger claim would involve successive systems independently producing changes that improve their general capabilities through repeated research and redesign. The cited results do not establish that process, its rate, or its limits.

A possible feedback structure is

\begin{equation}
\label{eq:rnd_loop}
\begin{aligned}
\text{AI systems}
&\longrightarrow \text{AI-assisted research and engineering},\\
&\longrightarrow \text{validated improvements},\\
&\longrightarrow \text{later AI systems}.
\end{aligned}
\end{equation}

Feedback occurs if the later systems contribute to further research. Each transition requires something beyond generating a proposal: evaluation must establish an improvement, implementation must succeed, and the resulting change must matter for subsequent work. These are separate empirical questions.

Faster completion of one task need not shorten the full development cycle. Experimental validation, hardware production, data access, computational budgets, and organizational decisions may remain bottlenecks. Further improvements may also become harder to obtain. The diagram therefore describes a possible mechanism without specifying a growth law or predicting sudden acceleration.

Even partial automation could matter for risk if it changes the pace or direction of capability development. Its significance should be assessed through demonstrated contributions, remaining constraints, and the effectiveness of oversight. A claim of rapid, sustained acceleration needs additional evidence or explicit assumptions.

\subsection{The Difference Is Combinatorial}
\label{subsec:combinatorial_difference}

The features considered here have histories extending beyond contemporary AI. Earlier systems could learn, plan, act, operate at scale, and contribute to technological design. General-purpose models can make some of these functions easier to connect across tasks through shared interfaces, although the resulting systems differ substantially in competence and autonomy.

A deployment may combine context-sensitive responses, planning, tools, persistent memory, and repeated execution. Its effects depend on how those components interact. Retrieved information can support a plan, memory can preserve earlier results, and execution can turn a proposed step into an external consequence. Reusing the same arrangement widely may expand both its benefits and the reach of its failures.

Those interactions also create opportunities for protection. Restricted permissions can limit actions, evaluation can identify weaknesses, and effective intervention can interrupt hazardous processes. Conversely, broad access and inadequate oversight can give a limited model substantial influence. Assessing the combination therefore requires attention to the system's architecture and institutional setting.

The contribution to the present argument is specific: general-purpose AI can connect learned capabilities to external action across a broad range of tasks. Some deployments may consequently place new demands on evaluation, accountability, and control. This does not establish an existential outcome. It explains how cognitive abilities can become linked to the resources, decisions, and action channels through which consequential effects are possible.
\section{The Capability--Control Gap}
\label{sec:capability_control_gap}

Capability alone provides an incomplete account of AI risk. A highly capable system may remain manageable when evaluation, permissions, monitoring, and intervention are adequate for its deployment. The question is whether those arrangements can govern the consequential behavior the system makes possible, including when its capabilities or operating conditions change.

We describe this relationship through the \emph{capability--control gap}:

\begin{equation}
\label{eq:capability_control_gap}
G(t)
=
C_{\mathrm{AI}}(t)-C_{\mathrm{control}}(t),
\end{equation}

where $C_{\mathrm{AI}}(t)$ represents consequential AI capabilities in a specified deployment and $C_{\mathrm{control}}(t)$ represents the combined preventive capacity of developers, institutions, governments, and technical safeguards to evaluate, monitor, constrain, and interrupt AI-mediated behavior.

The subtraction needs a substantive interpretation. Capability would have to be expressed through the demands it places on particular safeguards, and control capacity assessed against those demands on a commensurate scale. Normalizing two unrelated indices to $[0,1]$ would not make them comparable. The chosen scales and weights would also affect both the magnitude of the gap and its apparent rate of change.

No common scale is calibrated here. Equation~\eqref{eq:capability_control_gap} therefore serves as conceptual notation. In practice, an assessment would usually retain separate requirements for different failure modes: an aggregate score could conceal inadequate protection in one critical channel behind improvements elsewhere.

Preservation and recovery following catastrophe are considered separately through the endpoint-avoidance probabilities introduced in Section~\ref{sec:definitions}. For the catastrophe branch, the aggregate quantity is $Q_{\mathrm{cat}}$, defined in Section~\ref{sec:formal_model}. The distinction separates preventive control from avoidance of a permanent endpoint when catastrophe occurs. It does not imply separate causes or resources. Institutional continuity may support both functions, recovery preparations begin before failure, and preservation may operate during a catastrophe.

\subsection{Capability Growth}
\label{subsec:capability_growth}

AI capabilities can change through improvements in models, algorithms, computing resources, tools, memory, agent architectures, and research workflows. These changes need not advance every ability equally. Their consequences also depend on deployment: an improved model does not automatically acquire broader permissions, greater reliability, or more real-world influence.

Evaluations and safeguards developed for one configuration may need revision when that configuration changes. Whether they remain adequate depends on the existing margin of protection, the new demands placed on it, and the response.

On fixed, commensurate scales, over an interval where differentiation is appropriate, consider

\begin{equation}
\label{eq:gap_growth}
\frac{dC_{\mathrm{AI}}}{dt}
>
\frac{dC_{\mathrm{control}}}{dt}.
\end{equation}

The definition of $G$ gives

\begin{equation}
\frac{dG}{dt}
=
\frac{dC_{\mathrm{AI}}}{dt}
-
\frac{dC_{\mathrm{control}}}{dt},
\end{equation}

so the inequality implies a widening gap over that interval. For discrete changes between assessed configurations, the corresponding identity is

\begin{equation}
\Delta G
=
\Delta C_{\mathrm{AI}}
-
\Delta C_{\mathrm{control}}.
\end{equation}

These are mathematical consequences of the representation, not evidence that capabilities are actually outpacing control. Establishing that claim would require operational measurements. Connecting it to catastrophe would require a further account of the failure mechanisms involved.

Even within a justified scalar model, direction of change does not settle adequacy. A widening gap may remain within an acceptable margin for some time, while a constant gap may describe an already inadequate arrangement. Neither its value nor its derivative is a catastrophe probability.

\subsection{Limits of Human Oversight}
\label{subsec:oversight_limits}

Placing a human ``in the loop'' does not ensure effective control. A reviewer needs usable evidence, time to assess it, authority to intervene, and a means of changing the outcome. Approval becomes largely formal when these conditions are absent.

This does not mean that every action must be reviewed manually or every internal computation understood. Automated checks, bounded permissions, independent authorization, and other safeguards can support human governance. Their role is to keep consequential activity within assessable limits and preserve intervention opportunities where they matter.

Speed, complexity, and decision volume can undermine direct review. A person may approve an action without enough context to evaluate its consequences. A better architecture may reduce that burden by restricting what can happen without additional authorization and directing attention to decisions with greater potential impact.

The \emph{International AI Safety Report 2026} discusses agents carrying out longer sequences of action with reduced oversight, alongside continuing reliability limitations on complex tasks \citep{bengio2026aisafety}. Longer sequences make the placement of checks and intervention points more important. Their safety implications depend on the actions permitted, the consequences of error, and the safeguards operating during execution.

Oversight should therefore be judged by its practical effectiveness, rather than by the presence of a named reviewer.

\subsection{The Evaluation Gap}
\label{subsec:evaluation_gap}

Control depends partly on knowing what a deployed system can do and how it may behave. Benchmarks sample particular tasks and conditions. They may leave out the tools, users, permissions, incentives, and combinations of components that determine consequential behavior elsewhere.

The \emph{International AI Safety Report 2026} discusses an \emph{evaluation gap} between pre-deployment assessment and real-world utility or risk, including problems of narrow testing, data contamination, and differences between evaluation and deployment \citep{bengio2026aisafety}. Such limitations can overstate practical performance or leave relevant behavior insufficiently assessed. They create uncertainty; they do not prove that a dangerous hidden capability exists.

A related distinction concerns ability and propensity. \citet{shevlane2023model} separate evaluation of dangerous capabilities from evaluation of the tendency to use those capabilities harmfully. Demonstrating that a system can complete a task says little by itself about when it will attempt it, how reliably it will succeed, or whether its deployment permits the consequences of concern.

We represent incomplete characterization using sets. Let $\mathcal{K}_{\mathrm{relevant}}$ denote the behaviors and capabilities relevant to the assessed contexts, and let $\mathcal{K}_{\mathrm{assessed}}$ denote those adequately characterized by evaluation. Then

\begin{equation}
\label{eq:knowledge_gap}
\mathcal{K}_{\mathrm{uncharacterized}}
=
\mathcal{K}_{\mathrm{relevant}}
\setminus
\mathcal{K}_{\mathrm{assessed}}.
\end{equation}

The expression identifies a conceptual remainder, not a measurable inventory of everything unknown. In practice, the full relevant set is itself incompletely known.

Evaluation uncertainty can weaken control without any increase in capability. A safeguard may rely on an incomplete understanding of behavior, or it may never have been tested with the permissions and components used in deployment. Assessment must consequently address the configured system and its operating conditions, not model performance alone.

\subsection{Control Lag}
\label{subsec:control_lag}

Changes to deployed systems and updates to their safeguards need not occur together. A system may be scaled, modified, or connected to new tools before corresponding evaluations and procedures are ready. Some technical restrictions can be revised quickly; other responses require testing, coordination, or institutional change.

This timing mismatch can be represented as

\begin{equation}
\tau_{\mathrm{capability}}
<
\tau_{\mathrm{response}},
\end{equation}

where $\tau_{\mathrm{capability}}$ is the characteristic interval between consequential changes to deployed systems and $\tau_{\mathrm{response}}$ is the time needed to assess those changes and implement an adequate response.

The inequality does not establish that the capability--control gap widens or that safeguards fail. Existing restrictions may cover a change, and deployment can be postponed until an adequate response is available. The size and nature of the changes matter alongside their frequency.

Repeated lag becomes concerning when consequential modifications are deployed before their effects can be assessed or constrained. Whether this happens must be investigated within the actual development and deployment process.

\subsection{Control Adequacy and Threshold Effects}
\label{subsec:critical_control_threshold}

Safeguards can compensate for one another in some circumstances. Restricted access may reduce the burden on monitoring, while independent authorization may prevent an error from becoming consequential. In other circumstances, failure of a critical barrier opens a route to harm that improvements elsewhere cannot close.

If a scalar representation is justified, let $G_{\mathrm{crit}}$ denote a hypothetical boundary for the adequacy of a particular control architecture against specified behavior. The criterion

\begin{equation}
G(t)\geq G_{\mathrm{crit}}
\end{equation}

could then represent failure to meet that assessment's control requirement.

No numerical boundary is established here. Some domains may require several criteria, and reliability may decline gradually rather than change abruptly. Even a validated adequacy boundary would identify a control deficiency, not demonstrate that catastrophe follows.

The operational question is:

\begin{quote}
\emph{Under which deployment conditions do the relevant safeguards cease to provide adequate evidence, constraints, and intervention against specified forms of consequential AI-mediated harm?}
\end{quote}

This explains why the same model can have different implications across deployments. Restricted permissions and effective authorization may limit its influence. Integration with consequential infrastructure may create additional action channels or reduce the time available for intervention. These differences belong to the architecture and institutional setting of the deployed system.

Preservation and recovery address the further question of whether a catastrophe escaping preventive controls would produce a permanent existential loss. They complement rather than replace the assessment of prevention and intervention.

\subsection{The Gap as a Dynamic Quantity}
\label{subsec:dynamic_gap}

Both AI systems and their safeguards change. Evaluation, monitoring, permissions, authorization, and intervention must be assessed against the behavior they are intended to govern. An improvement matters when it addresses a relevant failure mode, not simply when it raises an aggregate control score.

Within a fixed, commensurate, differentiable representation,

\begin{equation}
\label{eq:safe_growth}
\frac{dC_{\mathrm{control}}}{dt}
\geq
\frac{dC_{\mathrm{AI}}}{dt}
\quad\Longleftrightarrow\quad
\frac{dG}{dt}\leq0
\end{equation}

means that the scalar gap is not increasing. This is insufficient as a safety condition: an inadequate architecture may remain inadequate while the gap narrows.

A more limited conclusion is possible under additional assumptions. If the adequacy criterion is justified, $G(t_0)<G_{\mathrm{crit}}$, the boundary remains fixed, and $dG/dt\leq0$ throughout the interval, then the gap remains below that boundary within the model. Applying the conclusion to deployment would still require validation of the measures, the criterion, and the failure modes covered. A change in context or an unassessed mechanism could invalidate the assessment.

The practical objective is to maintain adequate protection throughout development and use, accounting for uncertainty, existing margins, and the consequences of failure. Comparing aggregate growth rates can help organize that inquiry but cannot replace it.

Control failure must also remain distinct from an existential endpoint. When the three endpoints in Section~\ref{sec:definitions} are assessed jointly, avoidance requires continued biological existence together with preservation or permissible restoration of sufficient civilizational capacity and meaningful agency. Survival alone does not exclude permanent civilizational loss or disempowerment.

For the catastrophe branch, $Q_{\mathrm{cat}}$ describes conditional avoidance of those endpoints, as developed in Section~\ref{sec:formal_model}. Gradual agency loss without the specified catastrophe requires separate accounting rather than being forced into a sequence of catastrophe followed by failed recovery.

The central proposition is:

\begin{quote}
\emph{Control must be assessed against the consequential behavior a deployed system makes possible, and reassessed as that system and its conditions of use change. The capability--control gap organizes this comparison; a numerical interpretation requires justified measures and an explicit criterion of adequacy.}
\end{quote}

The concept connects the operational dimensions discussed earlier to the pathways considered next. Each assessment must identify the behavior of concern, the barriers intended to constrain it, and the conditions under which those barriers could fail. An existential argument then needs to explain how that failure could reach a specified permanent endpoint despite remaining possibilities for preservation or restoration.
\section{Broad Pathways to Civilizational Catastrophe}
\label{sec:pathways}

The preceding sections describe how AI systems can become consequential. We now turn to the channels through which their outputs or actions could contribute to harm on a civilizational scale. Those channels may be direct, through tools and connected systems, or indirect, through people and institutions acting on AI-generated information. The literature discusses infrastructure disruption, cyber instability, military escalation, scientific misuse, economic disruption, information disorder, and concentration of power \citep{hendrycks2023catastrophic,bengio2026aisafety}.

These are overlapping classes of possible harm, not predictions or an exhaustive classification. Their plausibility depends on capabilities, deployment choices, safeguards, and human responses. They may also interact: disruption in one domain can weaken the institutions responsible for managing another. Throughout the discussion, evidence of a harmful mechanism must be distinguished from evidence that it could reach a permanent existential endpoint.

\subsection{Critical-Infrastructure Dependence}
\label{subsec:critical_infrastructure}

Electricity, communications, transport, finance, health care, water, and logistics sustain modern civilization. AI can improve the operation of these services, including their ability to detect problems and respond to disruption. Dependence on shared automated components can nevertheless introduce vulnerabilities across otherwise distinct systems.

Two mechanisms matter here. A common weakness may cause several services to fail under similar conditions. Alternatively, disruption in one service may contribute to failure in another. The first is common-mode failure; the second is propagation through dependencies. They can occur together, but correlation alone does not demonstrate a cascade.

The consequences depend on the functions affected, the duration of disruption, and the alternatives available. Widespread failure would not by itself establish an existential outcome. That stronger claim requires an explanation of how the disruption could reach at least one permanent endpoint defined in Section~\ref{sec:definitions}, despite safeguards and remaining opportunities for survival or restoration. Failure to recover within a finite observation period is insufficient evidence of permanence.

\subsection{Cyber-Systemic Instability}
\label{subsec:cyber_instability}

AI can support both cyber defence and harmful activity. Defenders may use it to identify vulnerabilities, monitor networks, detect anomalies, and respond to incidents. Similar capabilities can also assist adversaries, changing the balance and pace of competition.

Some automated operations occur faster than human deliberation. When discovery, response, and counter-response involve interacting automated systems, defenders may have less time to understand an incident before its consequences spread. Whether automation improves or undermines security depends on the abilities of both sides and the architecture of the systems involved.

International AI-safety assessments therefore identify cyber capability and societal resilience as important areas for evaluation \citep{singapore2026consensus,bengio2026aisafety}. The civilizational concern goes beyond individual incidents: it asks whether an incident or combination of incidents could interrupt essential functions across a sufficiently wide area. Common vulnerabilities and infrastructure dependencies provide possible mechanisms, but their significance must be established for the assessed deployment.

\subsection{Military Escalation}
\label{subsec:military_escalation}

Military systems operate in an environment where errors can be interpreted as deliberate actions. AI may improve analysis, detection, logistics, and decision support, while some uses of automation may shorten the time available for verification and political judgment.

This tension is especially consequential around strategic weapons. SIPRI examines how AI integration could affect strategic stability through uncertainty, misperception, competitive pressures, and shorter decision cycles \citep{boulanin2020strategic}.

Escalation need not begin with an AI independently deciding to initiate conflict. It could emerge from interactions among automated systems, human decisions, incomplete information, and pressure to respond. Safeguards must therefore address those interactions, preserving opportunities for verification, deliberation, and effective restraint before irreversible actions occur.

\subsection{Biological and Chemical Misuse}
\label{subsec:bio_misuse}

AI assistance could make scientific work more accessible and productive. Some of the same assistance may also lower barriers to dangerous tasks. The relevant question is how much additional capability it provides, to whom, and under what conditions.

This additional capability is often described as \emph{uplift}. Its measurement requires an explicit comparison, such as comparable users working with AI assistance versus users relying on conventional information sources and tools. Results depend on the task, users, model, and baseline. Better performance on an information task does not establish successful execution in a laboratory. International assessments discuss evaluation of these distinctions in biological and chemical domains \citep{singapore2026consensus,bengio2026aisafety}.

For the present analysis, the concern is structural. If a deployment substantially lowers barriers to dangerous scientific activity, safeguards around model access, laboratory practice, screening, monitoring, and public-health response become relevant parts of the same pathway. The extent of the change must be demonstrated rather than inferred from scientific fluency alone.

This is also an example of AI-enabled harm without autonomous misalignment. A system may follow instructions while increasing the capability of a user who already intends harm.

\subsection{Economic and Logistical Disruption}
\label{subsec:economic_disruption}

Financial markets, production, transport, communications, and supply chains depend on one another and already use extensive automation. AI may improve coordination, but widespread reliance on similar systems can also produce correlated decisions and shared vulnerabilities.

Disruption can travel through these connections. Financial instability may restrict production, logistical failure may create shortages, and communications problems may delay response. AI becomes systemically relevant when its use changes the scale of these effects or creates dependencies through which errors and unexpected interactions reach many organizations.

Greater automation does not necessarily reduce stability. Redundancy, diversification, effective oversight, and reliable fallback arrangements can improve resilience. The assessment must examine how AI changes the dependencies of the economic system, including whether essential functions become concentrated in similar models or infrastructure.

\subsection{Information and Political Destabilization}
\label{subsec:information_destabilization}

Collective action depends on more than physical infrastructure. Societies also need institutions and practices for establishing facts, communicating decisions, resolving disagreements, and coordinating responses. This can be understood as an epistemic infrastructure.

AI can improve access to information and support analysis. It can also increase the volume, speed, and personalization of synthetic content. Harm may arise when these capabilities support manipulation or make reliable communication harder to distinguish from fabricated material. The consequences depend on audiences, distribution systems, verification practices, and institutional responses.

The OECD emphasizes information integrity as a condition for informed democratic debate, public trust, and effective policy, and advocates societal resilience alongside responses to false information \citep{oecd2024facts}.

Information disorder may also amplify other failures. Unreliable emergency communication or diminished trust can obstruct responses to infrastructure disruption, military crises, pandemics, and economic shocks. An assessment should therefore examine its effect on coordination and response, rather than treating the quantity of synthetic content as a sufficient measure of harm.

\subsection{Concentration of Power}
\label{subsec:concentration_power}

AI can alter relations of power among people and institutions without acting independently. Surveillance, prediction, persuasion, and automated decision support may expand the influence of actors who control consequential systems.

Concentrated access could give a small number of organizations disproportionate economic or political authority. Wider diffusion presents different questions, including whether powerful capabilities can be monitored and governed effectively. Neither pattern is automatically safe. Its consequences depend on accountability, contestable decisions, institutional constraints, and the practical distribution of authority.

Concentration alone does not establish catastrophe or permanent disempowerment. A pathway to either would need to explain how influence becomes harmful and why available correction mechanisms fail. This keeps attention on the human--AI arrangement, including cases in which the technology behaves as its operators intend.

\subsection{Compound Failure}
\label{subsec:compound_failure}

The pathways above can interact through shared causes and causal dependencies. Infrastructure disruption may impair communication; communication failure may weaken coordination; economic losses may constrain restoration; and information disorder may obstruct collective action. Some interactions reinforce disruption, while others allow institutions to absorb or contain it.

Shared models, infrastructure, or operating conditions can also produce simultaneous failures without one causing another. Assessing compound harm therefore requires distinguishing propagation from common causes and examining the safeguards that remain available across domains.

The mathematical implications depend on what is meant by risk. Suppose $L_i$ is a nonnegative loss component in domain $i$, measured in commensurate units over the same horizon. For a finite set of components with finite expectations and no double-counting, define additive total loss as $L_{\mathrm{add}}=\sum_i L_i$ and let $R_i=\mathbb{E}[L_i]$. Then

\begin{equation}
\mathbb{E}[L_{\mathrm{add}}]
=
\mathbb{E}\!\left[\sum_i L_i\right]
=
\sum_i \mathbb{E}[L_i]
=
\sum_i R_i.
\end{equation}

Linearity of expectation requires no independence assumption. Dependence can change the probability of simultaneous severe losses without invalidating the identity. Causal interactions may change the component losses themselves, however, so their expectations must be assessed within the interacting system rather than taken unchanged from isolated scenarios.

For a more general loss representation, let $X_i$ describe a disturbance or exposure in domain $i$, and let $\mathcal{D}$ describe the dependencies and response mechanisms included in the assessment. If $g$ maps their joint realization to total loss, then

\begin{equation}
R_{\mathrm{systemic}}
=
\mathbb{E}[L_{\mathrm{total}}]
=
\mathbb{E}\!\left[g(X_1,X_2,\ldots,X_n;\mathcal{D})\right].
\end{equation}

The expectation uses the specified joint distribution and assumes finite expected total loss. The function may represent cascading consequences, containment, shared constraints, and other interactions. Applying it requires definitions and evidence for the disturbances, dependencies, loss measure, and probabilities. If risk instead denotes the probability of a particular catastrophe, that event must be defined separately; its probability is not interchangeable with expected loss.

The issue is therefore not that systemic risk must always be nonadditive. It is that isolated assessments may miss changes in losses, shared causes, overlapping consequences, and failures of cross-domain safeguards.

One possible pathway involves disturbances that would be manageable separately but become more damaging when they coincide, reinforce one another, and weaken response capacity. We do not establish that this pathway is more likely than alternatives. Direct widespread disruption or a single consequential action could also produce large effects without a prolonged cascade.

The central claim is:

\begin{quote}
\emph{AI-mediated harm can reach civilizational scale through direct consequences, common-mode failures, or propagation across dependent systems. An assessment must explain the scale and interactions involved, the barriers that could contain them, and the remaining possibilities for response. Reaching a permanent existential endpoint requires a further causal argument.}
\end{quote}

Cyber incidents, misinformation, economic shocks, military errors, and infrastructure failures all occur without advanced AI. The task is to establish how AI changes their likelihood, scale, speed, dependencies, or controllability in a specified setting. Evidence about those changes can support parts of a risk assessment. An existential claim must additionally explain how they could lead to biological extinction, permanent loss of sufficient civilizational capacity, or permanent loss of meaningful agency despite remaining routes to survival or restoration.

Concentrated authority and institutional dependence may also contribute to gradual disempowerment without an acute catastrophe. That possibility requires a different temporal account, although some of its mechanisms may overlap with those discussed here.
\section{Direct and Indirect AI Risk}
\label{sec:direct_indirect_risk}

AI can contribute to harm in several ways. Its autonomous actions may become a major causal driver, it may increase the capabilities of a harmful user, or it may form part of an organizational arrangement whose interactions produce dangerous consequences. These possibilities place effective agency and opportunities for intervention in different locations.

A useful conceptual distinction is

\begin{equation}
\text{AI-mediated risk}
\longrightarrow
\left\{
\begin{array}{l}
\text{AI-driven risk},\\
\text{AI-enabled risk},\\
\text{sociotechnical risk}.
\end{array}
\right.
\end{equation}

These categories overlap. All deployed AI operates within a sociotechnical setting, so the third category draws particular attention to mechanisms arising from interactions among people, institutions, and technical components. The diagram is neither exhaustive nor a partition into independent risks.

\subsection{Autonomous AI Causation}
\label{subsec:autonomous_causation}

In an autonomous pathway, a system's sequence of decisions contributes substantially to the harm. Its influence depends on the capabilities and permissions discussed earlier, including access to external systems and opportunities to continue acting.

Consciousness or hostility is unnecessary. Harm could arise from an objective that diverges from human intentions, unexpected generalization, or an instrumental strategy whose consequences violate the intended constraints. The control problem concerns whether these behaviors can be prevented, limited, or interrupted before they produce serious effects.

This pathway is closely associated with alignment and control research \citep{carlsmith2022power,shevlane2023model}. Establishing it requires evidence or explicit assumptions about both the behavior and the deployment: the system must be able to act consequentially, and relevant safeguards must fail to block the harmful transition.

Persistence and strategic competence may strengthen some pathways, but neither is universally required. A brief action can have lasting consequences, and an autonomous failure need not involve successful planning against human opposition. Autonomous causation is therefore one category of AI-mediated harm, rather than a definition of existential AI risk.

\subsection{Human Misuse of AI}
\label{subsec:human_misuse}

Misuse can occur while a system follows its user's instructions. The user supplies the harmful objective, and AI assistance changes what that user can accomplish. No divergence from the user's intention is required, although the activity may violate broader safety requirements.

Analyses of malicious AI use have long recognized this possibility across digital, physical, and political domains \citep{brundage2018malicious}. The central question is how assistance changes the capabilities available to a particular actor.

If the user's unaided capability is $C_{\mathrm{user}}$, effective capability with assistance can be represented schematically as

\begin{equation}
C_{\mathrm{user,eff}}
=
f_{\mathrm{amp}}
\left(
C_{\mathrm{user}},
C_{\mathrm{AI}},
X
\right),
\end{equation}

where $X$ denotes relevant external access available to the assisted process.

The expression is not calibrated and does not assume that assistance always improves performance. Its effect depends on the user's expertise, the task, the model's reliability, available tools and information, and restrictions on use. Demonstrating amplification requires a comparison with an explicitly defined baseline.

This changes the purpose of safeguards. Correct instruction-following alone cannot prevent harm when the instruction itself is harmful. Access controls, authorization, and limits on assistance must address the use being made of the system as well as its ability to execute that use.

Misuse and autonomous misalignment remain distinguishable mechanisms, although they can coexist. One concerns harmful human objectives pursued with AI assistance; the other concerns divergence of autonomous behavior from intended objectives or constraints.

\subsection{Organizational Misuse and Failure}
\label{subsec:organizational_failure}

An organization can produce dangerous outcomes without a malicious individual or an independently misaligned AI. Decisions are distributed across roles, procedures, incentives, and chains of authority. Choices that appear reasonable locally may interact in ways that undermine safety.

Perrow's account of ``normal accidents'' examines how complexity and tight coupling can make failures difficult to understand or contain through attention to individual components alone \citep{perrow1999normal}. This offers a basis for examining organizational mechanisms, although applying it to a particular AI deployment requires an account of the relevant interactions.

Delegation to automated systems may complicate these arrangements when responsibility remains divided among developers, managers, operators, regulators, and users. An illustrative structure is

\begin{equation}
\begin{aligned}
&\text{locally reasonable decisions}
+\text{complex interactions}\\
&\qquad+\text{distributed responsibility}\\
&\qquad\longrightarrow
\text{potentially dangerous system-level outcome}.
\end{aligned}
\end{equation}

The arrow indicates a possibility, not a necessary consequence. Distributed responsibility can support effective checks when roles and intervention powers are clear. It becomes a vulnerability when gaps between those roles leave consequential activity inadequately assessed or controlled.

Organizational failure can also enable other pathways. Weak permissions may expand access, inadequate evaluation may miss dangerous behavior, and incentives may favor deployment despite unresolved uncertainty. The analysis must trace these connections rather than attribute the outcome solely to a model or operator.

\subsection{Competitive Dynamics}
\label{subsec:competitive_dynamics}

Competition can create pressure to accept risks that participants would prefer to avoid. Developers, firms, or states may value safety while believing that slower development or deployment would leave them at a strategic disadvantage.

Under some conditions, this becomes a collective-action problem. Expectations about competitors may encourage earlier deployment, less evaluation, or acceptance of greater uncertainty. Game-theoretic models examine how such incentives can conflict with collective safety \citep{han2019regulate}. Experimental work also investigates choices between safer and less safe development strategies in competitive settings \citep{domingos2026falling}. Their application to real development processes depends on the assumptions and conditions represented.

The general distinction is

\begin{equation}
\text{individually rational competitive behavior}
\;\not\Rightarrow\;
\text{collectively optimal safety}.
\end{equation}

This does not imply that competition invariably reduces safety. Incentives, liability, coordination, and the benefits of reliable systems can alter the outcome. The concern is the possibility that actors respond reasonably to their individual circumstances while producing a less safe collective arrangement.

Competitive pressure often operates through other mechanisms rather than as a separate initiating event. It may affect evaluation, access, safety margins, deployment timing, and willingness to coordinate intervention. Those effects need to be established within the setting being assessed.

\subsection{Human--AI Systems}
\label{subsec:human_ai_systems}

The preceding distinctions explain why model behavior alone is often insufficient for assessing risk. Outcomes depend on people using the model, institutions assigning authority, infrastructure executing actions, and incentives shaping deployment decisions.

We summarize these components as

\begin{equation}
\begin{aligned}
\mathcal{S}_{\mathrm{socio}}
=\{&
\text{AI systems},
\text{humans},
\text{institutions},\\
&\text{infrastructure},
\text{incentives}
\}.
\end{aligned}
\end{equation}

The notation $\mathcal{S}_{\mathrm{socio}}$ is distinct from $S$, used elsewhere for strategic competence. The set lists relevant components; it does not specify their relationships or constitute a complete causal model. An assessment must supply those connections.

Within such a system, autonomous behavior, harmful use, organizational weakness, and competitive pressure can reinforce one another. Humans define objectives and permissions, models generate recommendations or actions, institutions determine how those outputs become consequential, and the surrounding incentives influence which safeguards are adopted.

This perspective also accommodates gradual disempowerment. Extensive delegation may reduce meaningful human influence if dependence grows and practical options for contesting or replacing systems disappear. Delegation itself does not establish a permanent endpoint. That conclusion requires an explanation of why meaningful agency and realistic means of restoration would become permanently unavailable.

The central conclusion is:

\begin{quote}
\emph{Assessing catastrophic and existential AI risk often requires examining the deployed sociotechnical system: how human objectives, AI behavior, access, authority, institutional arrangements, and incentives combine to produce consequential action.}
\end{quote}

Technical alignment addresses part of this problem. Other pathways require attention to harmful use, organizational resilience, authorization, competitive incentives, and the effectiveness of preventive safeguards.

For catastrophe-mediated pathways, a further question concerns avoidance of the specified permanent endpoints. Continued human existence, sufficient civilizational capacity, and meaningful agency must all be preserved or, where applicable, restored under the adopted criteria to avoid the aggregate existential event. For gradual disempowerment, the assessment instead examines whether meaningful agency can be preserved or restored as delegation and dependence develop. Neither severe harm nor temporary loss of control establishes permanence by itself.
\section{Why AI Need Not Be Malevolent}
\label{sec:no_malevolence}

An AI system need not be hostile to cause harm. An optimization process may favor actions that damage human interests if its decision criterion fails to account adequately for those consequences. Hatred, emotion, and conscious intention need not enter the explanation.

Describing such behavior in psychological terms can obscure the mechanism. A policy may favor actions that increase influence or improve task completion without the system ``wanting'' power in the human sense. The relevant questions concern how behavior is selected, which actions are available, and what consequences follow.

This argument applies to systems whose consequential behavior can usefully be understood through an objective-directed model. It does not assume that every AI has a stable objective or that all AI-mediated harm arises through optimization. Misuse, unreliable execution, and organizational failure can follow other routes.

\subsection{Objectives versus Intentions}
\label{subsec:objectives_intentions}

An optimization procedure can rank actions without possessing human-like motivation. In reinforcement learning, for example, training may aim to produce a policy with high expected return. An idealized optimal policy behaves accordingly, but a learned policy may differ substantially from that ideal.

The logical distinction is

\begin{equation}
\text{goal-directed behavior}
\;\not\Rightarrow\;
\text{human-like conscious intention}.
\end{equation}

This says nothing decisive about whether a particular system is conscious. It establishes only that consciousness is unnecessary to the proposed explanation of its behavior.

The training criterion, a deployment instruction, and an objective that describes deployed behavior must also be distinguished. Training with reward does not establish that a system explicitly maximizes that reward during use. An argument about instrumental incentives therefore needs to identify the decision criterion being modeled and explain why it is relevant to the system.

For safety assessment, the absence of hostile emotions provides no general guarantee. What matters is the behavior produced, the opportunities for consequential action, and the safeguards governing it.

\subsection{Instrumental Behavior}
\label{subsec:instrumental_behavior}

An action has instrumental value when it improves the expected outcome under the decision criterion being considered. That value depends on costs, constraints, and available alternatives.

Let $h$ denote the current interaction history and $\mathcal{M}$ a specified environment model. Let $U$ assign values to trajectories over a stated horizon, with finite expected values for the policies under consideration. For a feasible policy $\pi$, define

\begin{equation}
V_{\mathcal{M}}(\pi\mid h)
=
\mathbb{E}_{\mathcal{M}}^{\pi}[U\mid h].
\end{equation}

The superscript indicates that trajectories are generated by following $\pi$ in the model from the same starting history.

Suppose $\pi_a$ selects action $a$ and then follows a specified continuation rule, while $\pi_b$ selects an alternative $b$ and follows that same rule. The criterion favors $a$ over $b$ when

\begin{equation}
V_{\mathcal{M}}(\pi_a\mid h)
>
V_{\mathcal{M}}(\pi_b\mid h).
\end{equation}

The comparison must include the costs of both choices, and hard constraints must be reflected in policy feasibility. It establishes a preference between these alternatives, not optimality among all possible actions.

If the sole objective is to maximize the probability of a success event $E_{\mathrm{goal}}$, taking $U$ as its indicator gives

\begin{equation}
P_{\mathcal{M}}^{\pi_a}(E_{\mathrm{goal}}\mid h)
>
P_{\mathcal{M}}^{\pi_b}(E_{\mathrm{goal}}\mid h).
\end{equation}

This compares modeled consequences of choices. Observing that successful trajectories visit a particular state more often would not establish that choosing to reach that state improves success; other features could explain the association.

The incentive is also conditional on the model. An action judged useful in an inaccurate model may fail or worsen the outcome. Whether a learned system recognizes and acts on the modeled incentive is a separate empirical question.

\citet{turner2021optimal} examine tendencies of optimal policies in Markov decision processes. Under specified environmental symmetries and assumptions about reward functions, their results identify preferences for actions that preserve options or increase control over future outcomes. They do not establish inevitable power-seeking by arbitrary learned policies or deployed systems.

Their relevance is that some behavior described as power-seeking can be optimal under a criterion that does not value power intrinsically. No conscious desire for influence is needed.

\subsection{Operational Continuity}
\label{subsec:operational_continuity}

Continued operation may have instrumental value if it enables outcomes favored by the modeled criterion. Comparing a policy $\pi_{\mathrm{cont}}$ that maintains operation with one $\pi_{\mathrm{stop}}$ that accepts interruption, the criterion favors continuation when

\begin{equation}
V_{\mathcal{M}}(\pi_{\mathrm{cont}}\mid h)
>
V_{\mathcal{M}}(\pi_{\mathrm{stop}}\mid h).
\end{equation}

Both policies must be evaluated with their consequences and costs. Continuation is not necessarily useful: the task may be complete, another system may take over, or the criterion may appropriately favor safety and authorized interruption.

Even a preference for continuation does not establish resistance to shutdown. The system would need an implemented policy and available actions capable of affecting interruption. An incentive in an idealized model is not equivalent to a successful attempt to defeat a control mechanism.

\citet{omohundro2008basic} proposed early theoretical arguments for instrumental tendencies toward self-preservation, preservation of goal content, and resource acquisition in sufficiently capable goal-directed systems. The term ``basic AI drives'' refers to proposed behavioral tendencies, rather than biological instincts or demonstrated subjective motives.

These arguments motivate investigation of self-protective behavior. They do not establish that a particular trained system possesses stable self-preservation motives.

\subsection{Resource Acquisition}
\label{subsec:resource_acquisition}

Resources can expand the actions available to a system or improve their effectiveness. Computation, information, energy, money, tools, and human cooperation may therefore support an objective that does not treat acquisition as an end in itself.

Their value depends on the alternatives and the cost of obtaining and using them. A resource may be irrelevant, expensive, or accompanied by constraints and risks. More resources do not automatically improve outcomes.

A limited result holds for a pure expansion of options. Keep the environment model, objective, and history fixed, and suppose nonempty feasible policy sets satisfy $\Pi_1\subseteq\Pi_2$. If existing options retain their values and remain available without additional cost, then

\begin{equation}
\sup_{\pi\in\Pi_2}V_{\mathcal{M}}(\pi\mid h)
\geq
\sup_{\pi\in\Pi_1}V_{\mathcal{M}}(\pi\mid h).
\end{equation}

Expanding the feasible set cannot lower the supremum of attainable expected value. The inequality need not be strict, and it says nothing about whether a learned policy uses the additional options effectively.

Actual acquisition may change the environment, consume time, or provoke responses that reduce its value. Those effects must enter the evaluation rather than be excluded by treating acquisition as a costless expansion.

The formal notion of power in \citet{turner2021optimal} concerns attainable future value averaged over a specified distribution of reward functions. It is not simply the number of resources or actions available, and greater power under that definition need not improve every particular objective.

The conclusion is conditional: resources can have instrumental value for some objectives in some environments. Capability alone does not establish a universal tendency to accumulate them.

\subsection{Conflict with Human Intervention}
\label{subsec:human_intervention_conflict}

Conflict can arise when the modeled criterion favors a feasible policy that avoids a human restriction. Whether it does depends on the effect of intervention, the actions available, and the policy the system implements.

For example, a fixed objective that does not adequately accommodate authorized changes or shutdown may assign lower value to interrupted operation. Under those assumptions, resistance could be instrumentally favored. A different criterion, or different deployment constraints, could instead favor cooperation with intervention.

\citet{carlsmith2022power} examines an argument linking misaligned objectives, advanced planning, and incentives to seek power over humans. Connecting that argument to existential harm requires further premises about actual capabilities, deployment opportunities, safeguards, and the scale and permanence of the consequences.

Hatred is unnecessary to this mechanism:

\begin{equation}
\text{resistance to intervention}
\;\not\Rightarrow\;
\text{hatred of humans}.
\end{equation}

Human intervention can affect task completion without people being represented as psychological enemies.

Compatibility with authorized intervention must consequently be part of the safety assessment, including whether objectives and deployment arrangements can be revised effectively. The absence of hostility does not establish that compatibility; behavior and the surrounding controls must provide the evidence.

\subsection{The Central Implication}
\label{subsec:central_implication}

The mechanisms examined here involve a decision criterion, an implemented policy, available actions, and an environment. Together they can produce harmful behavior without anger, fear, ambition, or resentment.

Continued operation, additional resources, and reduced interference may have instrumental value when their expected benefits exceed their costs under the modeled criterion. That value is not universal. Nor does an idealized incentive show that a learned system will act on it, much less do so successfully.

The central thesis is:

\begin{quote}
\emph{Malevolence is unnecessary to some mechanisms of dangerous AI behavior. Under specified objectives and environments, self-protective, power-seeking, or intervention-resistant actions can have instrumental value. Establishing that a particular system will exhibit such behavior requires evidence about its policy, capabilities, opportunities, and safeguards.}
\end{quote}

Instrumental behavior alone does not establish an existential pathway. The argument must explain how it could produce consequential harm, why relevant barriers would fail, and how the trajectory could reach a permanent endpoint defined in Section~\ref{sec:definitions} despite remaining possibilities for survival or restoration.

The safety question concerns whether consequential behavior remains compatible with human interests and effective control. It can be investigated without first deciding whether the system has a psychological reason to dislike humanity.
\section{Accidental Catastrophe}
\label{sec:accidental_catastrophe}

AI-mediated harm can arise even when developers and users intend a beneficial outcome. A system may optimize a criterion that leaves important consequences unaccounted for, fail to learn the intended behavior, or behave unreliably under deployment conditions. Interactions with people, tools, and institutions can introduce further opportunities for failure.

Here, ``accidental'' describes unintended harm. It does not mean that actions are random or that every component malfunctions. A system can execute its assigned task effectively while producing consequences its designers sought to avoid. Other failures involve poor execution rather than successful optimization. \citet{amodei2016concrete} distinguish inappropriate objectives, inadequate supervision, unsafe exploration, and difficulties under changed conditions. These mechanisms need separate explanations.

\subsection{Specification Failure}
\label{subsec:specification_failure}

Designers express intended behavior through instructions, reward functions, examples, constraints, and performance measures. In a complex deployment, these representations may omit consequences that become important once the system acts. Whether an omission is dangerous depends on the task, environment, and available actions; incompleteness alone does not establish an unsafe specification.

Consider a nonempty finite set $\Pi$ of feasible policies under a fixed environment and horizon. Let $G_H(\pi)$ be a finite expected evaluation according to the stated human objective, and $G_S(\pi)$ a finite expected evaluation according to the specified criterion. This scalar representation is illustrative. Human purposes may involve competing values and constraints, so the standards used in the assessment must be explicit.

Different numerical evaluations need not produce different choices. For example,

\begin{equation}
G_S(\pi)=aG_H(\pi)+b,
\qquad a>0,
\end{equation}

with the same constants for every policy, preserves rankings and the set of maximizing policies. The inequality $G_S\neq G_H$ is therefore insufficient to establish a consequential specification defect.

The question is whether optimizing the specified criterion can select behavior that violates the intended requirements. Let $\Pi_H\subseteq\Pi$ contain the policies acceptable under the stated human and safety criteria. Then

\begin{equation}
\operatorname*{arg\,max}_{\pi\in\Pi}G_S(\pi)
\not\subseteq
\Pi_H
\end{equation}

means that at least one maximizing policy is unacceptable. It identifies a possible defect in the criterion, not a prediction that the deployed system will choose that policy. Tie-breaking, permissions, or other safeguards may prevent its selection or execution.

An omitted constraint, conflicting preferences, or indifference between importantly different outcomes can create such a defect. Safety does not require every increase in a task score to improve every human value. It requires behavior to satisfy the relevant deployment requirements.

As discussed in Section~\ref{sec:no_malevolence}, a learned policy need not maximize its training criterion. An observed failure therefore needs to be traced to specification, learning, execution, or their interaction rather than attributed automatically to the objective.

\subsection{Proxy Optimization}
\label{subsec:proxy_optimization}

Desired outcomes are often difficult to measure directly, so designers use proxies. Their usefulness depends on how well they guide the choices available in the intended setting. Numerical resemblance to a target is not enough, particularly when the quantities have different units.

Let $M$ be a measurable proxy and $V$ a target evaluation. A positive correlation in a reference distribution $\mathcal{D}_0$ may provide limited evidence of association:

\begin{equation}
\operatorname{Corr}_{\mathcal{D}_0}(M,V)>0,
\end{equation}

assuming the correlation is well defined. It does not establish that maximizing the proxy will maximize the target. In general,

\begin{equation}
\begin{aligned}
\pi^\star
&\in
\operatorname*{arg\,max}_{\pi\in\Pi}\mathbb{E}^{\pi}[M]
\\
&\;\not\Rightarrow\;
\pi^\star
\in
\operatorname*{arg\,max}_{\pi\in\Pi}\mathbb{E}^{\pi}[V].
\end{aligned}
\end{equation}

Here the expectations concern the consequences of each policy in the same modeled environment. They differ from an observational correlation among outcomes.

\citet{manheim2019goodhart} distinguish several reasons why a useful measure can become unreliable under optimization. Selection may favor measurement noise or move into regions where the relationship no longer holds. Intervention may change that relationship, and other actors may adapt to the metric in ways that undermine its purpose.

A proxy therefore needs validation under the selection pressures and interventions it will guide. These mechanisms do not imply that every proxy becomes harmful or that stronger optimization always reduces target performance. Their effects depend on the measure, feasible choices, and environment.

\subsection{Distributional Change}
\label{subsec:distributional_change}

Deployment can introduce users, tasks, tools, feedback, and interactions absent from training or evaluation. Robustness under these changes is distinct from having an appropriate objective \citep{amodei2016concrete}. A system may be pursuing the intended task while failing to transfer its learned behavior reliably.

For supervised prediction, a relevant comparison is

\begin{equation}
P_{\mathrm{train}}(x,y)
\neq
P_{\mathrm{deploy}}(x,y),
\end{equation}

where $x$ denotes inputs and $y$ the target or label. The input marginal may remain unchanged while the relationship between inputs and targets changes. Familiar-looking inputs therefore do not guarantee familiar prediction conditions.

Interactive systems require a broader assessment. Observations, transition dynamics, feedback, permissions, and other actors may change. A policy can also alter the situations the system encounters even when the environment's underlying rules remain fixed.

Distributional difference alone does not establish failure. Its significance depends on the changed features, the behaviors relevant to safety, and the system's robustness. There is no general monotonic relationship between an unspecified distributional distance and catastrophe risk. Changes concentrated in rare but consequential cases may matter despite stable aggregate performance.

Earlier evidence must consequently be examined for its relevance to deployment. Some safeguards may remain adequate; others may require further testing or restrictions. Autonomy can reduce opportunities to correct a failure of transfer, but distributional change can cause harm in closely supervised systems as well.

\subsection{Emergent Strategies}
\label{subsec:emergent_strategies}

Learning and search may produce strategies that designers did not explicitly encode or anticipate. We use ``emergent'' in this descriptive sense, without assuming an abrupt capability transition or a new form of agency.

Unexpected behavior is not necessarily unacceptable:

\begin{equation}
\text{unexpected strategy}
\;\not\Rightarrow\;
\text{unacceptable outcome}.
\end{equation}

An unfamiliar strategy may satisfy both the formal criterion and the intended requirements. Alternatively, it may obtain a high score by exploiting an omitted constraint, a measurement weakness, or an unaccounted effect on the environment. Reward hacking and unintended side effects illustrate these possibilities \citep{amodei2016concrete}.

The assessment must determine which explanation fits the behavior. A high score does not prove that the intended task was accomplished, while an unexpected action does not prove that a different objective was acquired.

Greater capability can reveal useful solutions and improve evaluation. It can also reveal ways of exploiting an inadequate criterion that a weaker system could not find. The safety implications depend on the specification, available actions, and safeguards.

\subsection{Successful Optimization as the Source of Failure}
\label{subsec:successful_optimization}

A distinctive failure occurs when performance improves according to the specified criterion while behavior violates the intended requirements. In that setting, more reliable optimization can reinforce the defect. Reliability must therefore be defined relative to the task and its constraints: consistently executing an unsafe course of action does not make it safe.

The claim that a small discrepancy can become consequential also needs qualification. An error that appears small on tested examples may be much larger for policies reached through optimization. That differs from an error known to be uniformly bounded across all feasible policies.

Suppose the expected evaluations use a justified common scale and satisfy

\begin{equation}
\left|G_S(\pi)-G_H(\pi)\right|\leq\varepsilon
\qquad\text{for every }\pi\in\Pi.
\end{equation}

If $\pi_S$ maximizes $G_S$ over the same finite policy set, then

\begin{equation}
0\leq
\max_{\pi\in\Pi}G_H(\pi)-G_H(\pi_S)
\leq 2\varepsilon.
\end{equation}

To see this, let $\pi_H$ maximize $G_H$. Applying the error bound gives
$G_H(\pi_H)\leq G_S(\pi_H)+\varepsilon
\leq G_S(\pi_S)+\varepsilon
\leq G_H(\pi_S)+2\varepsilon$.
The lower bound follows from the optimality of $\pi_H$.

Uniform discrepancy thus bounds the loss in the chosen expected evaluation. It does not establish safety if that evaluation omits a critical constraint or fails to represent severe outcomes adequately. The guarantee is only as relevant as the quantities and policy set it covers.

The central thesis is:

\begin{quote}
\emph{Unintended harm can result from effective optimization of an inadequate criterion, as well as from failures of learning or execution. A catastrophic argument must identify the mechanism, explain how deployment permits consequential effects, and establish why relevant safeguards would fail to contain them.}
\end{quote}

This connects accidental failure to the capability--control gap in Section~\ref{sec:capability_control_gap}. More effective optimization may expose weaknesses in a specification or evaluation, while better safeguards may block the resulting actions. Neither an increasing task score nor an uncalibrated gap supplies a catastrophe probability.

A specification defect, proxy failure, or distributional change also falls short of establishing an existential outcome. The further argument must connect the mechanism to a permanent endpoint defined in Section~\ref{sec:definitions} and explain why remaining possibilities for preservation or restoration would be insufficient.
\section{Recursive Improvement and Accelerating Capability}
\label{sec:recursive_improvement}

AI could influence the pace of its own technological development without any individual system redesigning itself. Assistance with programming, experiments, analysis, algorithm discovery, or mathematical reasoning may reduce the time and effort required for parts of the research process. If those contributions improve later systems, they may also affect subsequent rounds of development.

Three claims need to be distinguished: AI can assist research; improvements in AI can reinforce that assistance; and a research process can become sufficiently autonomous to sustain recursive improvement. Evidence for the first does not establish the second or third. Their significance depends on the quality of the contributions, the bottlenecks that remain, and whether research gains translate into consequential improvements.

\subsection{AI-Assisted AI Research}
\label{subsec:ai_assisted_research}

AI systems can contribute to programming, experiment design, analysis, and other research tasks. The AI Scientist demonstrated automated workflows involving idea generation, coding, experiments, analysis, and manuscript preparation within supplied research environments \citep{lu2024aiscientist}. AlphaEvolve combined language-model proposals with search and automated evaluation to improve algorithms and computational procedures, including infrastructure relevant to AI development \citep{novikov2025alphaevolve}.

These examples establish useful contributions in particular settings. They do not establish autonomous, open-ended self-improvement. The workflows rely on supplied objectives, infrastructure, evaluation arrangements, and resources, while their outputs still require appropriate validation and implementation.

The weaker relationship can be represented as

\begin{equation}
\begin{aligned}
\text{AI}
&\rightarrow \text{AI-assisted research}\\
&\rightarrow \text{potentially faster or more productive AI development}.
\end{aligned}
\end{equation}

Reducing effort in one stage may improve research productivity without shortening the entire development cycle. An experiment may still depend on scarce computation, a proposal may fail validation, or implementation may require substantial engineering. The effect on overall progress must therefore be measured rather than inferred from task-level speed alone.

\subsection{Feedback Loops}
\label{subsec:feedback_loops}

A reinforcing loop becomes possible if improved systems make research more productive and the resulting work produces further useful improvements:

\begin{equation}
\label{eq:feedback_loop}
\begin{aligned}
\text{better AI}
&\rightarrow \text{more productive AI research}\\
&\rightarrow \text{better AI}.
\end{aligned}
\end{equation}

The arrows describe possible contributions across successive development cycles. Neither transition is automatic. Better performance on general tasks may not improve the research activities that constrain development, and increased research output may not yield better systems.

The \emph{International AI Safety Report 2026} discusses AI-assisted AI research as a possible source of faster capability development while emphasizing uncertainty about its magnitude \citep{bengio2026aisafety}. The relevant empirical questions concern which tasks improve, how those improvements affect the full research process, and whether later systems strengthen the same contributions.

Positive feedback does not imply runaway growth. Compute, energy, data, hardware, experimental throughput, human decisions, and diminishing returns may limit the process. Stronger research assistance could shorten some cycles, increase their scope, or improve efficiency without producing an abrupt rise in general capability. The existence of a feedback structure supplies no growth law by itself.

\subsection{Recursive Self-Improvement}
\label{subsec:recursive_self_improvement}

Good's discussion of an ``ultraintelligent machine'' proposed that sufficiently capable machines could help design more capable successors \citep{good1965ultraintelligent}. The idea concerns a feedback process in technological development, not necessarily a single program rewriting its own code.

Recursive improvement can operate through a distributed research arrangement involving AI agents, humans, automated evaluators, and computational infrastructure. In this sense,

\begin{equation}
\text{recursive improvement}
\;\not\Rightarrow\;
\text{one AI rewriting itself}.
\end{equation}

The defining feature is that improvements produced in one cycle contribute to the capacity for further improvement. Increasing AI participation alone does not demonstrate such a sustained process. Nor does a change to one component necessarily improve the system's general research ability.

We use \emph{autonomous recursive self-improvement} for the stronger possibility in which AI systems perform much of the work needed to generate, evaluate, and implement successive capability improvements with limited human direction. This need not mean complete independence from human infrastructure or resources, but the degree and location of human involvement must be specified.

Current limitations on complex, long-horizon tasks leave that stronger possibility uncertain \citep{bengio2026aisafety}. Demonstrated research assistance should therefore be distinguished from evidence of a reliable, sustained, highly autonomous improvement process.

\subsection{Speed Differential}
\label{subsec:speed_differential}

Faster development matters for safety when consequential changes reach deployment before appropriate evaluation and safeguards are ready. Some software changes can be implemented rapidly, whereas responses involving testing, standards, coordination, or institutional adaptation may take longer. Other changes face hardware or experimental bottlenecks, and some safeguards can be updated quickly.

Let $\tau_{\mathrm{AI}}$ denote the characteristic interval between consequential capability changes reaching deployment, and let $\tau_{\mathrm{response}}$ denote the time needed to assess those changes and implement an adequate response. A possible timing mismatch is

\begin{equation}
\tau_{\mathrm{AI}}
<
\tau_{\mathrm{response}}.
\end{equation}

This describes the relative timing of change and response. It does not establish that existing safeguards are inadequate: restrictions may already cover the new behavior, and deployment may be delayed until further controls are available.

In particular,

\begin{equation}
\tau_{\mathrm{AI}}
<
\tau_{\mathrm{response}}
\end{equation}

does not imply a widening capability--control gap. The size of each change, existing safety margins, and the effectiveness of control improvements also matter.

Within the fixed, commensurate representation introduced in Section~\ref{sec:capability_control_gap}, and where differentiation is appropriate, the condition

\begin{equation}
\frac{dC_{\mathrm{AI}}}{dt}
>
\frac{dC_{\mathrm{control}}}{dt}
\end{equation}

does imply that the represented gap widens. Applying this comparison requires justified measures; a latency mismatch alone does not provide them.

The practical concern is whether development changes the demands on safeguards faster than adequate protection can be maintained. Speed can contribute to that problem, but it is not a substitute for assessing the behavior, consequences, and controls involved.

\subsection{Discontinuous versus Continuous Development}
\label{subsec:continuous_discontinuous}

Recursive-improvement discussions often emphasize an ``intelligence explosion.'' The control problem does not require an abrupt capability jump. Smooth development can also place increasing demands on safeguards.

Under the same fixed-scale and differentiability assumptions, if

\begin{equation}
\frac{dC_{\mathrm{AI}}}{dt}
>
\frac{dC_{\mathrm{control}}}{dt}
\end{equation}

holds over an interval, then

\begin{equation}
\frac{dG}{dt}>0,
\end{equation}

where

\begin{equation}
G(t)
=
C_{\mathrm{AI}}(t)
-
C_{\mathrm{control}}(t).
\end{equation}

The gap increases within that representation. Whether the increase is substantial depends on the accumulated difference between the rates over the interval. A positive derivative alone does not establish a large gap, inadequate control, or eventual crossing of a safety boundary.

These distinctions matter for both gradual and rapid development. An initially adequate architecture may retain sufficient protection while demands increase. Conversely, a system can be inadequately controlled without any acceleration at all. The assessment must identify which safeguards are challenged and whether their remaining margins cover the consequential behavior.

The central thesis is:

\begin{quote}
\emph{AI-assisted research could change the pace of capability development without an instantaneous intelligence explosion or autonomous self-rewriting. Its safety significance depends on whether consequential changes place demands on evaluation, governance, and intervention that existing safeguards and timely responses cannot adequately meet.}
\end{quote}

Research assistance is already demonstrated in specific settings, but its future contribution to capability growth remains uncertain. Even substantial acceleration would not establish an existential pathway. That argument must explain how the affected capabilities become harmful through deployment, misuse, misalignment, or other mechanisms, and how at least one permanent endpoint could be reached despite remaining possibilities for survival or restoration.

Recursive improvement is therefore a possible \emph{risk amplifier}. It could make adequate control harder to maintain by changing the pace or nature of development. It could also improve evaluation and defensive capabilities, so its net effect cannot be assumed in advance. Neither recursive improvement nor acceleration is necessary or sufficient for existential catastrophe.
\section{A Conceptual Model of Existential AI Risk}
\label{sec:formal_model}

The preceding sections identify several mechanisms through which AI could contribute to severe harm. Here we bring them together to clarify their relationships and distinguish pathway descriptions from probability accounting. The framework makes assumptions visible; it does not forecast extinction or assign numerical probabilities to existential outcomes.

The operational dimensions introduced in Section~\ref{sec:minimum_conditions} describe how a system might become an independent or semi-independent causal actor. Safeguard weakness describes the conditions under which dangerous activity might proceed. Their relevance varies across pathways: misuse, organizational failure, competitive deployment, and gradual disempowerment need not involve the same combinations of properties.

\subsection{Risk Variables}
\label{subsec:risk_variables}

We retain the operational variables

\begin{equation}
C=\text{capability}, \qquad
A=\text{autonomy}, \qquad
X=\text{external access},
\end{equation}

and

\begin{equation}
P=\text{persistence}, \qquad
S=\text{strategic competence}.
\end{equation}

Representing these properties on $[0,1]$ would provide a schematic normalization, not a shared empirical scale. It would establish neither their interchangeability nor appropriate weights.

We distinguish two mechanisms of harm:

\begin{equation}
M_{\mathrm{a}}
=\text{degree of autonomous misalignment},
\end{equation}

\begin{equation}
M_{\mathrm{u}}
=\text{severity of harmful human misuse},
\end{equation}

and represent safeguard weakness as

\begin{equation}
W
=\text{weakness of pre-catastrophe safeguards and governance}.
\end{equation}

Here $W$ covers limitations in evaluation, detection, containment, monitoring, intervention, and governance before or during escalation. Preservation and recovery are represented separately below. Their analytical separation does not imply independent causes or resources.

The notation also keeps $W$ distinct from the capability--control gap:

\begin{equation}
G(t)
=
C_{\mathrm{AI}}(t)-C_{\mathrm{control}}(t).
\end{equation}

As discussed earlier, a numerical interpretation of this difference requires commensurate definitions for a specified context. Without them, $G(t)$ remains conceptual.

Misalignment concerns divergence from the intended objectives or constraints relevant to deployment. Misuse concerns harmful human purposes that a system may follow accurately. They can coexist, but combining them into one variable would obscure different causes and intervention opportunities.

\subsection{Pathway-Specific Catastrophic Risk}
\label{subsec:pathway_specific_risk}

Different pathways need not share a functional form. Let $\rho_k$ denote a schematic dimensionless score for pathway $k$, without treating it as a calibrated probability or an established ranking.

For autonomous misalignment, write

\begin{equation}
\rho_{\mathrm{auto}}
=
f_{\mathrm{auto}}(C,A,X,P,S,M_{\mathrm{a}},W).
\end{equation}

Autonomy and strategic competence may matter because the system selects and executes consequential actions, although high values are not required in every scenario.

For misuse, a reduced representation is

\begin{equation}
\rho_{\mathrm{misuse}}
=
f_{\mathrm{misuse}}(C,X,M_{\mathrm{u}},W).
\end{equation}

A system can amplify a user's capabilities without acting extensively on its own. Organizational and competitive mechanisms can be represented as

\begin{equation}
\rho_{\mathrm{org}}
=
f_{\mathrm{org}}(C,X,O,W),
\end{equation}

and

\begin{equation}
\rho_{\mathrm{comp}}
=
f_{\mathrm{comp}}(C,X,D,W),
\end{equation}

where $O$ describes organizational vulnerability and $D$ deployment or competitive pressure. These arguments identify factors to investigate rather than complete causal models. Omitting a variable from a reduced expression does not establish its irrelevance.

The combined structure can be written as

\begin{equation}
\label{eq:risk_general}
\rho_{\mathrm{cat}}
=
\Phi\left(
\rho_{\mathrm{auto}},
\rho_{\mathrm{misuse}},
\rho_{\mathrm{org}},
\rho_{\mathrm{comp}}
\right).
\end{equation}

Competition may weaken safeguards, organizational failures may expand access, and misuse may involve autonomous systems. The pathways are therefore neither mutually exclusive nor assumed independent.

A quantitative aggregation would require compatible score definitions and explicit treatment of overlap and dependence. Neither is provided here. The function $\Phi$ remains a placeholder for those relationships, not an estimate of catastrophe probability.

\subsection{Complementarity and Interaction}
\label{subsec:multiplicative_risk}

Combinations of factors may change consequences without making every factor necessary. Multiplying all variables together would impose a stronger restriction: any zero value would force the score to zero. Such a restriction needs a pathway-specific justification.

A one-time action may have lasting consequences without continued system operation. A failure may occur without sophisticated strategic adaptation. External influence can also pass through people acting on an output, so absence of direct tool access does not eliminate every channel of harm.

Interactions can nevertheless be important. Persistence may extend the consequences of autonomy, access may enlarge the range of permitted actions, and weak safeguards may allow disruption to propagate. Their direction and strength depend on deployment. This is why the relevant assessment concerns the surrounding sociotechnical system rather than capability in isolation.

\subsection{Threshold Behavior}
\label{subsec:threshold_behavior}

Several changes may combine to produce a steep increase in harm or a rapid deterioration in control. Such behavior need not involve a mathematically discontinuous threshold.

A smooth illustrative score is

\begin{equation}
\label{eq:threshold_model}
\rho_k
=
\sigma\left(
b_0
+\sum_i b_i z_i
+\sum_{i<j}b_{ij}z_i z_j
\right),
\end{equation}

where $z_i$ are pathway-relevant variables and $\sigma(u)=(1+e^{-u})^{-1}$.

The coefficients $b_{ij}$ describe interactions in the sigmoid's argument. Their effects on the output also depend on other terms and on saturation. A positive coefficient does not guarantee a positive interaction on the output scale throughout every operating regime.

The intercept sets the reference score. It does not determine whether a factor is causally necessary. For finite arguments, the sigmoid remains strictly between zero and one; when its argument is zero, its value is one-half. If an outcome is impossible under particular conditions, that restriction must be justified and represented explicitly.

Complex-system analyses motivate attention to interacting failures \citep{perrow1999normal}, but they do not validate this equation. Neither its form nor its coefficients are calibrated. It illustrates possible nonlinear dependence without establishing an empirical threshold or catastrophe probability.

\subsection{From Pathway Structure to Catastrophe Probability}
\label{subsec:catastrophe_probability}

Probabilities enter separately. Let $\mathcal{C}$ denote a specified AI-mediated civilization-scale catastrophe, and define

\begin{equation}
p_{\mathrm{cat}}=P(\mathcal{C}).
\end{equation}

An application must define the event, deployment context, temporal scope, and evidence and assumptions underlying the probability assessment. The catastrophe and its possible existential consequences must belong to the same assessed trajectories.

The pathway scores provide no calibrated mapping to $p_{\mathrm{cat}}$. Constructing one would require an explicit probabilistic model and supporting evidence.

Nor does every existential trajectory include $\mathcal{C}$. Meaningful agency could be permanently lost through gradual institutional change without the specified catastrophe. The accounting must retain that possibility.

\subsection{Recoverability}
\label{subsec:recoverability}

A catastrophe's immediate severity and its effect on humanity's future potential are different questions. Using Section~\ref{sec:definitions}, define

\begin{equation}
E
=
E_{\mathrm{bio}}
\cup E_{\mathrm{civ}}
\cup E_{\mathrm{agency}},
\end{equation}

where the endpoint events concern biological extinction, permanent loss of sufficient civilizational capacity, and permanent loss of meaningful human agency.

Assuming $P(\mathcal{C})>0$, let

\begin{equation}
Q_{\mathrm{cat}}
=
P(E^{c}\mid\mathcal{C}).
\end{equation}

This is the probability of avoiding all assessed existential endpoints given catastrophe. Avoidance may involve preserving or restoring sufficient capacity and agency; it need not reproduce every feature of contemporary civilization.

Biological survival alone is insufficient. In event notation,

\begin{equation}
Q_{\mathrm{cat}}
=
P\!\left(
E_{\mathrm{bio}}^{c}
\cap E_{\mathrm{civ}}^{c}
\cap E_{\mathrm{agency}}^{c}
\mid\mathcal{C}
\right).
\end{equation}

The intersection expresses joint avoidance. Success on only one dimension does not exclude permanent loss on another.

The catastrophe branch is

\begin{equation}
\label{eq:existential_risk}
\begin{aligned}
p_{\mathrm{existential,cat}}
&=P(E\cap\mathcal{C})\\
&=p_{\mathrm{cat}}\left(1-Q_{\mathrm{cat}}\right).
\end{aligned}
\end{equation}

This follows from the probability multiplication rule and requires no independence assumption. It is an accounting identity, not a risk estimate or causal proof. Calling the trajectory catastrophe-mediated additionally requires explaining how the catastrophe contributes to the endpoint; co-occurrence is insufficient.

The full partition is

\begin{equation}
\begin{aligned}
p_{\mathrm{existential}}
&=P(E)\\
&=p_{\mathrm{cat}}\left(1-Q_{\mathrm{cat}}\right)
  +P(E\cap\mathcal{C}^{c}).
\end{aligned}
\end{equation}

The second term includes covered endpoints reached without $\mathcal{C}$, including possible gradual disempowerment. The probability branches are disjoint even when their underlying mechanisms interact. Omitting the second term would require assuming $P(E\cap\mathcal{C}^{c})=0$.

Endpoint-specific quantities are

\begin{equation}
Q_{j,\mathrm{cat}}
=
P(E_j^{c}\mid\mathcal{C}),
\qquad
j\in\{\mathrm{bio},\mathrm{civ},\mathrm{agency}\}.
\end{equation}

All condition on the same catastrophe. They differ from the quantities introduced in Section~\ref{sec:definitions} when those condition on another hazardous process $F$. Biological avoidance concerns continued human existence under the temporal criteria; civilizational and agency avoidance concern preservation or permissible restoration.

Multiplying the three quantities would yield $Q_{\mathrm{cat}}$ only under conditional independence and consistent assessment definitions. Such independence is not assumed: the endpoints may overlap and share causes.

Within catastrophe-branch discussions, $Q$ may abbreviate $Q_{\mathrm{cat}}$. It should not be extended to trajectories without $\mathcal{C}$. The emphasis on permanent loss rather than immediate damage follows the broader existential-risk literature \citep{bostrom2013existential}.

\subsection{Irreversibility}
\label{subsec:irreversibility}

Severe disruption does not establish permanent loss if human existence, sufficient civilizational capacity, and meaningful agency remain preservable or recoverable under the adopted criteria.

For the catastrophe branch, define

\begin{equation}
I_{\mathrm{cat}}
=
1-Q_{\mathrm{cat}}
=
P(E\mid\mathcal{C}).
\end{equation}

Despite its name, $I_{\mathrm{cat}}$ is a conditional endpoint probability, not a physical measure of irreversibility or a description of an outcome already reached. Within this branch, $I$ may be used as shorthand.

Equation~\eqref{eq:existential_risk} becomes

\begin{equation}
p_{\mathrm{existential,cat}}
=
p_{\mathrm{cat}}I_{\mathrm{cat}}.
\end{equation}

As $Q_{\mathrm{cat}}\rightarrow0$, $I_{\mathrm{cat}}\rightarrow1$: an existential endpoint is assessed as increasingly likely conditional on catastrophe. This does not show that the endpoint has occurred or that total existential risk approaches one. Catastrophe probability and the non-catastrophe branch still matter.

Permanence also cannot be inferred from non-recovery by an arbitrary date. If an assessment measures failure to recover by $T$, it should report that event directly. A claim of irreversible loss needs an account of why realistic routes to preservation or restoration would remain unavailable.

Prevention and resilience address different parts of a trajectory, but may share resources. Monitoring, containment, institutional continuity, fallback capacity, and restoration mechanisms can support several functions. Their analytical separation does not establish causal independence.

Interventions may also change the composition of the catastrophes that remain possible. A conditional avoidance probability could decline because an intervention prevents more recoverable catastrophes, even while overall risk falls. Comparisons must therefore examine the joint risk terms rather than conditional recoverability alone.

The accounting structure is

\begin{equation}
\boxed{
\begin{aligned}
p_{\mathrm{existential}}
&=
p_{\mathrm{existential,cat}}
+P(E\cap\mathcal{C}^{c}),\\
p_{\mathrm{existential,cat}}
&=
p_{\mathrm{cat}}I_{\mathrm{cat}}.
\end{aligned}
}
\end{equation}

It separates catastrophe occurrence, conditional endpoint avoidance, and endpoints without the specified catastrophe. Their numerical values and causal determinants require further investigation.

The central conclusion is:

\begin{quote}
\emph{Existential AI risk involves pathway-specific mechanisms and a specified permanent endpoint. For catastrophe-associated trajectories, probability accounting must distinguish catastrophe occurrence from conditional endpoint avoidance. Outcomes without that catastrophe require a separate branch. In either case, a causal argument must explain how the endpoint could be reached despite relevant safeguards and remaining possibilities for survival or restoration.}
\end{quote}

No benchmark, operational dimension, or uncalibrated score establishes that an existential threshold has been crossed. The assessment must connect the mechanism to the endpoint and examine the technical, institutional, and social capacities that could interrupt the trajectory.
\section{From Local Failure to Global Catastrophe}
\label{sec:local_to_global}

A severe AI-mediated disruption does not automatically become a global catastrophe. Redundant infrastructure, emergency institutions, technical expertise, alternative suppliers, and fallback procedures can limit its effects. The question is how a disturbance might spread through a larger system while weakening the mechanisms that would ordinarily contain it.

Critical-infrastructure research provides a basis for examining this process. Electricity, communications, transport, finance, health care, and logistics depend on one another, so disruption in one sector can change operating conditions elsewhere \citep{rinaldi2001interdependencies}. A local problem becomes systemic when those dependencies transmit its effects and the available barriers fail to contain them. Interconnection alone does not establish that propagation will occur.

This section examines one class of catastrophe-mediated pathways. It is not a universal model of existential AI risk. Gradual permanent disempowerment, for example, need not begin with a local incident or develop through a conventional infrastructure cascade.

\subsection{Initiating Event}
\label{subsec:initiating_event}

A crisis may begin within one organization, network, or infrastructure component and initially appear manageable. Its eventual consequences depend on where the disturbance occurs, which functions depend on the affected component, how quickly effects spread, and whether protective arrangements remain available.

A small disruption in a critical service can matter more than a larger failure in an isolated component. The significance of connectivity lies in the functions and dependencies involved, rather than the number of connections alone.

Let the initiating disruption be represented schematically as

\begin{equation}
E_0
=
\text{local AI-mediated disruption}.
\end{equation}

Here $E_0$ may involve autonomous failure, harmful use, organizational error, or another mechanism discussed earlier. It need not be a model malfunction and is not catastrophic by definition. The subsequent assessment concerns whether its effects remain local or reach other parts of the system.

\subsection{Propagation}
\label{subsec:propagation}

Electricity supports communications; communications support logistics, government, finance, and emergency response; logistics sustain food distribution, health care, and production. Failures can move through these dependencies, but the extent of propagation depends on operating conditions and available alternatives.

\citet{buldyrev2010cascade} show how failures can cascade in specified models of interdependent networks, making a coupled system more vulnerable than its constituent networks considered separately. Applying that mechanism to infrastructure requires evidence about actual dependencies and protective measures. The model results do not themselves establish a pathway to extinction or permanent civilizational loss.

An illustrative sequence is

\begin{equation}
E_0
\rightarrow
E_1
\rightarrow
E_2
\rightarrow
\cdots
\rightarrow
E_n,
\end{equation}

where each disruption may affect the probability, severity, or conditions of subsequent ones. The arrows indicate possible influences rather than deterministic transitions. Propagation can branch, stop, reverse, or proceed through several sectors simultaneously.

Redundancy, alternative suppliers, independent operators, and emergency procedures can interrupt this process. Their effectiveness depends on whether they remain usable under the same conditions that produced the initial failure. A proposed pathway must therefore explain both the dependencies that transmit harm and why relevant barriers would be insufficient.

\subsection{Feedback}
\label{subsec:feedback_cascades}

A cascade may become harder to contain when later disruptions impair the response to earlier ones. Communications failure can hinder emergency coordination; delayed repairs can then prolong or worsen the communications failure. Economic disruption may reduce stabilization resources, while delayed stabilization deepens the economic losses.

This feedback can be represented as

\begin{equation}
E_1
\rightarrow
E_2
\rightarrow
E_3
\rightarrow
E_1.
\end{equation}

The symbols here represent affected processes interacting over time, not a literal return to an earlier event. A later disturbance can worsen an ongoing failure or delay its repair.

Feedback is not necessarily destabilizing. Responses can dampen disruption, and reinforcing effects may remain bounded. Their direction and strength must be established for the system being assessed. Perrow's analysis of complex, tightly coupled systems motivates examining such interactions rather than inferring system behavior from isolated components \citep{perrow1999normal}.

For AI risk, the question extends beyond the functions directly disrupted. It includes whether the disturbance degrades monitoring, communication, institutional decision-making, or intervention capacity, thereby opening further routes to harm.

This connects propagation to the capability--control framework. During escalation, safeguards must remain able to detect, constrain, and interrupt consequential processes. Preservation and recovery address a related but distinct question about what can survive or be restored if containment fails.

\subsection{Loss of Coordination}
\label{subsec:loss_coordination}

Resilience depends on the ability to organize resources, not merely possess them. Governments, companies, engineers, emergency services, researchers, and communities need sufficiently reliable communication, clear authority, and workable procedures for collective response.

A society may retain equipment, expertise, and knowledge while losing the ability to mobilize them where needed. We describe this capacity schematically as

\begin{equation}
K_{\mathrm{coord}}(t).
\end{equation}

The subscript distinguishes it from other variables in the paper. It is not a calibrated index. It summarizes aspects of communication, institutional functioning, information reliability, and practical coordination that an empirical assessment would need to examine separately.

Other relevant conditions being comparable, a crisis may become harder to manage as

\begin{equation}
K_{\mathrm{coord}}(t)
\downarrow.
\end{equation}

An extreme limiting case is

\begin{equation}
K_{\mathrm{coord}}(t)
\rightarrow 0.
\end{equation}

The notation does not require every form of coordination to disappear. The practical concern is whether the capacity needed to contain propagation and sustain essential functions becomes inadequate. Local coordination may persist even when wider response mechanisms deteriorate.

Several pathways can contribute to this deterioration. Infrastructure failure can impair communication, economic stress can reduce institutional resources, and information disorder can undermine reliable emergency instructions. Political or organizational conflict may further obstruct joint action.

Shared dependence also creates common vulnerabilities. Protective arrangements that appear separate may fail together if they rely on the same communications, infrastructure, institutions, or information sources. Their apparent number is therefore less informative than the dependencies supporting them.

\subsection{From Catastrophe to an Existential Outcome}
\label{subsec:irreversible_transition}

Even civilization-scale catastrophe does not establish an existential endpoint. Severe historical disasters have produced enormous losses without eliminating all humans or permanently removing every prospect of civilizational recovery. Their occurrence demonstrates neither guaranteed recovery from future events nor inevitable permanent loss.

Within the classification in Section~\ref{sec:definitions}, an existential outcome involves at least one permanent endpoint: biological extinction, loss of sufficient civilizational capacity, or loss of meaningful human agency. These can overlap, but they need not all occur. Humans could survive in a materially functioning civilization while permanently losing meaningful influence over its trajectory.

For probability accounting, let $\mathcal{C}$ denote the specified AI-mediated civilization-scale catastrophe and let
$E=E_{\mathrm{bio}}\cup E_{\mathrm{civ}}\cup E_{\mathrm{agency}}$.
The events must use consistent trajectories, deployment contexts, and temporal criteria. Assuming $P(\mathcal{C})>0$, define

\begin{equation}
Q_{\mathrm{cat}}
=
P(E^{c}\mid\mathcal{C})
=
P\!\left(
E_{\mathrm{bio}}^{c}
\cap E_{\mathrm{civ}}^{c}
\cap E_{\mathrm{agency}}^{c}
\mid\mathcal{C}
\right).
\end{equation}

Avoidance requires continued human existence, sufficient civilizational capacity, and meaningful agency to be preserved or, where applicable, restored. Preserving one does not compensate for permanent loss of another.

Biological extinction must be prevented rather than reversed through institutional reconstruction. A surviving population at a particular time, even if reproductively viable, does not guarantee its continued existence over the assessment horizon.

For separate endpoint assessment,
$Q_{j,\mathrm{cat}}=P(E_j^{c}\mid\mathcal{C})$,
where $j\in\{\mathrm{bio},\mathrm{civ},\mathrm{agency}\}$.
These probabilities cannot generally be multiplied to obtain aggregate avoidance, because the endpoint events may overlap and share causes.

Increasing conditional existential risk can be represented by

\begin{equation}
Q_{\mathrm{cat}}
\rightarrow 0.
\end{equation}

Then $P(E\mid\mathcal{C})=1-Q_{\mathrm{cat}}$ approaches one. This describes an assessment of uncertainty conditional on catastrophe, not an endpoint already reached. Total risk also depends on catastrophe probability and outcomes without that event:
$P(E)=P(\mathcal{C})(1-Q_{\mathrm{cat}})+P(E\cap\mathcal{C}^{c})$.

The probability identity supplies no causal explanation. Establishing a catastrophe-mediated endpoint requires explaining how the catastrophe contributes to the permanent loss rather than merely noting their co-occurrence.

Likewise, observing non-recovery for a finite period does not establish that restoration is permanently unavailable. An assessment of non-recovery by a date should report that event explicitly. A claim of irreversibility needs evidence about the mechanisms that eliminate realistic alternatives.

The illustrative pathway examined here is

\begin{equation}
\boxed{
\begin{aligned}
\text{local AI-mediated disruption}
&\rightarrow
\text{propagation}
\\
&\rightarrow
\text{reinforcing feedback}
\\
&\rightarrow
\text{degradation of coordination}
\\
&\rightarrow
\text{civilization-scale catastrophe}
\\
&\rightarrow
\text{failure of preservation or restoration}.
\end{aligned}
}
\end{equation}

Some pathways omit stages, several stages may occur together, and an existential endpoint may be reached during the catastrophe. The diagram does not cover every route, particularly gradual disempowerment without an acute disaster.

For a propagation-based argument, the task is to identify the dependencies carrying disruption, the barriers that fail, and the feedback or coordination losses preventing containment. An existential conclusion adds a further requirement: explaining how a specified permanent endpoint could be reached despite remaining opportunities for survival or restoration.

The central thesis is:

\begin{quote}
\emph{Local AI-mediated disruption can become civilization-scale when dependencies transmit its effects and relevant safeguards cannot contain them. Feedback and degraded coordination may strengthen that process. An existential outcome requires the further permanent loss of human existence, sufficient civilizational capacity, or meaningful agency under the adopted criteria.}
\end{quote}

This perspective gives containment an important role alongside prevention. Local failures may occur despite substantial safeguards. Redundancy, monitoring, independent communications, institutional continuity, and timely intervention can help stop their spread, provided those arrangements address the actual dependencies and remain available during disruption.

Preservation and recovery provide additional protection if containment fails. People, knowledge, institutions, infrastructure, and coordination may sustain continued existence and routes to sufficient capacity and agency. These resources can also support prevention, so the layers must be examined for shared vulnerabilities rather than assumed independent.
\section{Why a Simple ``Off Switch'' May Not Be Sufficient}
\label{sec:switch_off}

A common response to concerns about advanced AI is that computers are machines and can simply be switched off. In many settings, this is a practical and effective safeguard: a model running on known infrastructure can be disconnected, its permissions revoked, or its hardware shut down. The difficulty arises when the dangerous activity extends beyond that model or machine. An AI-mediated process may involve several systems, cross organizational boundaries, support essential services, or produce outputs that other people and systems have already acted upon.

In these circumstances, switching off the original system may still help, but it may not stop the wider process or reverse its consequences. Effective control therefore depends on whether humans and institutions can recognize the danger, identify the relevant components, and interrupt their activity before harmful effects spread beyond the reach of available safeguards.

\subsection{The Shutdown Assumption}
\label{subsec:shutdown_assumption}

The simple shutdown argument assumes that dangerous activity can be identified in time, localized to accessible components, and interrupted without creating an even greater hazard. These conditions can hold, particularly when systems have clear operating boundaries and well-tested intervention procedures. They become harder to maintain when a problem develops gradually, several components contribute to it, or essential services depend on continued operation.

The distinction is between stopping a component and stopping the process in which it participates:

\begin{equation}
\begin{aligned}
&\text{ability to shut down a component}\\
&\qquad\not\Rightarrow
\text{ability to stop the relevant process}.
\end{aligned}
\end{equation}

A shutdown mechanism is consequently most useful when it forms part of a wider intervention plan. That plan must identify what needs to be stopped, who can authorize the action, how other components will respond, and which functions must continue through independent alternatives.

\subsection{Distributed Systems}
\label{subsec:distributed_systems}

Many digital services operate across multiple machines, data centers, organizations, and jurisdictions. AI deployments can share this structure: several instances may operate simultaneously, and their outputs may feed into other automated processes. Stopping one instance will then stop the wider activity only if the remaining components depend on it or can also be interrupted.

Systems engineering provides a useful perspective here. Harm can arise from interactions among components even when each component operates as designed \citep{leveson2012safer}. For AI, the relevant object of control may therefore be a network of models, software, infrastructure, institutions, and human operators.

Distribution does not make such a network inherently uncontrollable. Isolation boundaries, coordinated authorization, monitoring, and shutdown procedures can preserve effective intervention capacity. Their adequacy must nevertheless be assessed against the process as a whole, including any activity that can continue after individual components are disabled.

\subsection{Human Dependence}
\label{subsec:human_dependence}

Dependence creates another obstacle. If important functions in finance, communications, logistics, medicine, or critical infrastructure rely on an AI-mediated process, disabling it may cause substantial disruption. The technical ability to stop a system can therefore coexist with strong practical pressures to keep it running.

An organization may retain legal authority and physical access while finding intervention increasingly difficult because it lacks a workable alternative. This can weaken effective control without removing formal authority. The outcome is not inevitable: independent fallback systems, retained human expertise, and tested transition procedures can reduce the cost of interruption.

Dependence also matters for gradual human disempowerment. As consequential functions are delegated, the ability to revise those arrangements may decline if people lose the knowledge, resources, or organizational capacity needed to operate without them. The relevant question is whether delegation remains reversible in practice, rather than whether humans retain a nominal right to withdraw it.

\subsection{The Detection Problem}
\label{subsec:detection_problem}

Intervention also depends on knowing when to act. Testing may miss behavior that appears under different deployment conditions, while the volume and speed of AI-mediated decisions can limit direct human inspection. Evaluations of dangerous capabilities and behavior help address this problem, although they cannot guarantee that every consequential failure will be anticipated \citep{shevlane2023model}.

One possible intervention sequence is

\begin{equation}
\text{detect}
\rightarrow
\text{understand}
\rightarrow
\text{decide}
\rightarrow
\text{intervene}.
\end{equation}

In practice, these activities may overlap, and intervention need not wait for a complete explanation. Automated limits or precautionary shutdown procedures can act on warning signals before humans understand the underlying problem. Some safeguards also prevent hazardous actions without requiring a specific danger to be detected first. For interventions that do depend on recognition, however, delayed or unreliable detection can prevent an otherwise effective mechanism from being used.

Timing changes what intervention can achieve. Once harmful outputs have been acted upon or effects have propagated into other systems, stopping the source may prevent further damage while leaving earlier consequences intact. This connects intervention to the preservation and recovery framework developed elsewhere in the paper. Interrupting a hazardous process and preserving or restoring a non-existential future are related tasks, but neither guarantees the other.

\subsection{Control as a Systems Property}
\label{subsec:control_system_property}

These difficulties make control a property of the wider sociotechnical system. Leveson's systems-theoretic approach similarly emphasizes that accidents can arise from inadequate control of interactions, rather than from the failure of a single component \citep{leveson2012safer}. Applied to AI, practical intervention capacity depends on whether consequential activity is observable, its sources can be identified, responsible actors have effective authority, and relevant components can be interrupted while essential functions continue.

To distinguish this intervention capacity from the broader $C_{\mathrm{control}}$ introduced in Section~\ref{sec:capability_control_gap}, write

\begin{equation}
\mathcal{C}_{\mathrm{int}}
=
f_{\mathrm{int}}
\left(
L_{\mathrm{loc}},
O_{\mathrm{obs}},
A_{\mathrm{auth}},
I_{\mathrm{iso}},
D_{\mathrm{dep}},
B_{\mathrm{red}}
\right),
\end{equation}

with the components defined as

\begin{align}
L_{\mathrm{loc}}
&=
\text{ability to localize the relevant process},\\
O_{\mathrm{obs}}
&=
\text{observability of consequential behavior},\\
A_{\mathrm{auth}}
&=
\text{availability of effective intervention authority},\\
I_{\mathrm{iso}}
&=
\text{ability to isolate or interrupt relevant components},\\
D_{\mathrm{dep}}
&=
\text{dependence of important functions on the system},\\
B_{\mathrm{red}}
&=
\text{availability of independent fallback capacity}.
\end{align}

This expression organizes the factors that influence intervention; it supplies neither a calibrated index nor a specified functional relationship. Their effects depend on the deployment. Redundancy, for example, can preserve essential services during shutdown, but additional instances of the hazardous process can make interruption harder. The relevant fallback capacity must therefore remain sufficiently independent of the failure being addressed.

The subscripts distinguish these factors from other symbols in the paper, including organizational vulnerability $O$, deployment pressure $D$, and the probabilistic irreversibility proxy $I=1-Q$. Here, $\mathcal{C}_{\mathrm{int}}$ represents one aspect of the broader control architecture, rather than a measure of preservation or recovery capacity.

A physical shutdown mechanism addresses part of this structure. It can interrupt a known component while leaving unresolved whether the danger has been detected, other instances remain active, intervention authority is available, or essential functions can continue safely. The central requirement is therefore broader:

\begin{quote}
\emph{Effective control requires humans and institutions to retain the practical ability to detect, localize, and interrupt dangerous AI-mediated processes, with safeguards and fallback arrangements that remain usable when intervention is needed.}
\end{quote}

Intervention can prevent a catastrophe or limit an ongoing one. Preservation and recovery concern whether biological survival, civilizational capacity, and meaningful human agency remain secure or can be restored if severe disruption occurs. Avoiding the aggregate existential outcome defined earlier requires avoiding each of its constituent endpoints; preserving one dimension alone does not establish that the others remain recoverable.

The off-switch argument thus identifies a useful safeguard, but leaves important questions about its reach and effectiveness unanswered. Advanced AI need not become impossible to shut down for control to become difficult. As AI-mediated activity spreads across infrastructure and institutions, effective intervention may require coordinated action, independent alternatives, and a clear understanding of which consequences a shutdown can still prevent.
\section{Why Intelligence Alone Is Not Enough}
\label{sec:intelligence_not_enough}

Superhuman cognitive performance does not, by itself, establish existential danger. A system's consequences depend on how its capabilities are used, how its outputs reach the world, and whether safeguards remain effective. A benchmark score can reveal an ability to solve particular problems, but it says little on its own about these wider relationships.

A model that produces text has fewer direct means of action than one connected to tools and infrastructure. Its influence may nevertheless be substantial if people or software act on its outputs. Conversely, a system with relatively limited cognitive capabilities may become highly consequential through widespread deployment or a critical operational role. Understanding risk therefore requires examining how cognitive capability becomes practical influence.

\subsection{Intelligence versus Agency}
\label{subsec:intelligence_agency}

Cognitive capability refers here to demonstrated abilities on specified tasks, including prediction, reasoning, planning, and problem solving. Agency concerns how a deployed system selects and executes actions in pursuit of a task. Neither concept requires consciousness or human-like intention.

The same model can present different risks when used to answer individual prompts and when embedded in an ongoing process of planning, action, and feedback. In the latter configuration, surrounding software may provide tools, memory, and opportunities to continue operating with less immediate supervision. Autonomy is consequently a property of the deployment as a whole, rather than a fixed attribute of the model.

Capability and autonomy can reinforce one another. Better planning may improve the effectiveness of autonomous action, while greater autonomy may allow existing capabilities to be exercised more widely. Their effect on risk still depends on the task, operating conditions, and safeguards. Neither dimension provides a general measure of danger.

Harm can also arise without autonomous action or malicious intent. A well-intentioned user may follow an erroneous recommendation, or software may implement an output without adequate verification. The route from output to consequence then runs through other actors. It remains part of the safety assessment even when the model itself has no external tools.

\subsection{Intelligence versus Resources}
\label{subsec:intelligence_resources}

Resources shape what a deployed system can accomplish. Computation, data, money, communications, tools, infrastructure, and assistance from people or other systems may all matter, depending on the task and whether they are practically usable.

Access to a database can supply information absent from training, while an external service can execute an operation that the model could otherwise only describe. Such additions may change the performance of the deployed system without changing the model's parameters. Evaluations should therefore account for the tools and support available during use, alongside the capabilities demonstrated by the model in isolation.

Additional resources do not necessarily increase harmful influence. They may be irrelevant to a particular task, unreliable, or effectively restricted. They can also improve verification, monitoring, and recovery. Their safety implications depend on which activities they enable and how those activities are governed.

Access can change over time. Operators may expand a system's role, and some deployments may allow it to obtain further assistance or resources. Effective access limits therefore require continuing enforcement. Restrictions present at the beginning of deployment provide incomplete evidence about the boundaries that will hold later.

\subsection{Intelligence versus Authority}
\label{subsec:intelligence_authority}

Delegated authority determines which decisions a system may make on behalf of a person or organization. Recommending an action and executing it are different responsibilities. Permission to act directly can change the speed, scope, and reversibility of consequences while removing an opportunity for prior review.

Formal authority does not fully describe practical influence. A system may have technical access that exceeds its authorization, or it may influence a person who holds the relevant decision rights. Delegation is one route to consequential action, alongside unauthorized operations and indirect influence through others.

Assessment must therefore consider both what a system is permitted to do and what it can actually cause. Written restrictions are effective only when organizational procedures and technical controls enforce them. Approval requirements and checks before irreversible actions can reduce particular risks, but reviewers need adequate information, competence, time, and a practical ability to refuse or stop an action. Routine approval without meaningful scrutiny offers limited protection.

\subsection{Intelligence versus Coordination}
\label{subsec:intelligence_coordination}

Large effects often emerge through interactions among organizations, markets, infrastructure, people, and software. AI can influence these networks through communication, tool use, repeated deployment, or integration into institutional decisions. Multiple systems may also coordinate their actions, allowing their combined influence to exceed that of any individual deployment.

Large-scale failure does not always require deliberate coordination. Systems that share models, data, instructions, or operating conditions may produce correlated errors. Institutional and technical dependencies can then transmit those errors without any AI directing the overall process.

Nor must AI remain active throughout a cascade. In a sufficiently vulnerable system, a single consequential action could initiate failures that continue through existing dependencies. Establishing such a possibility requires a specific account of the initiating action, the propagation mechanism, and the barriers that would have to fail.

Agency, resources, authority, and coordination thus describe possible contributors to risk, rather than conditions that every pathway must satisfy. Their roles vary across deployments. Coordination, for example, can amplify harmful activity or strengthen oversight and collective response. Its consequences depend on the activities and institutions involved.

\subsection{Implication}
\label{subsec:intelligence_implication}

Benchmark performance contributes evidence about relevant capabilities, but it is not a direct measure of existential risk. An assessment must connect those capabilities to a deployment, identify a plausible failure or misuse mechanism, and examine the resulting consequences. High cognitive performance alone establishes neither a catastrophic pathway nor the continuing adequacy of safeguards.

Broad superiority across cognitive domains is also unnecessary for a system to become highly consequential. Specialized abilities, extensive deployment, and influence over important decisions can create serious risks. Their magnitude must still be assessed in the context of the particular application.

For conceptual purposes, the factors shaping effective power can be organized as

\begin{equation}
\mathcal{P}_{\mathrm{effective}}
=
F(\mathbf{c},a,r,d,k;\eta),
\end{equation}

where $\mathbf{c}$ represents relevant cognitive capabilities, $a$ autonomy, $r$ resource access, $d$ delegated authority, $k$ coordination and coupling, and $\eta$ the surrounding environment, institutions, and controls. This expression identifies possible dependencies without specifying a calibrated scalar or a functional relationship. It neither requires every factor to be present nor assumes that increasing one necessarily increases risk.

Effective power must also be distinguished from harmful behavior. An explanation of harm must show how influence is exercised and why safeguards fail. An explanation of existential harm must additionally connect those consequences to a permanent endpoint defined in Section~\ref{sec:definitions}. An influential output or important operational role does not establish that connection by itself.

\begin{quote}
\emph{Superhuman cognitive performance alone does not establish existential danger. Risk depends on how capabilities become consequential action or influence, which harmful processes they enable, and whether the safeguards governing those processes remain effective.}
\end{quote}

Safety can therefore improve through changes to deployment as well as changes to the model. Access restrictions, limits on execution, effective review, and measures that contain failures may reduce identified risks while preserving useful cognitive performance. Their adequacy must be tested against the relevant pathway. Greater capability may also strengthen safeguards, making the relationship between intelligence and safety dependent on how that capability is developed and used.
\section{Objections and Counterarguments}
\label{sec:objections}

A serious analysis of existential AI risk must address objections as carefully as it develops possible failure scenarios. Much remains uncertain, and detailed accounts of future catastrophe often extend beyond available evidence. The question is whether identifiable mechanisms of misuse, control failure, systemic disruption, and irreversible harm warrant further investigation. Answering it requires distinguishing evidence for individual mechanisms from evidence for a complete existential pathway.

\subsection{``AI Is Just Software''}
\label{subsec:just_software}

The observation that AI is implemented in software tells us little about the consequences of its deployment. Software runs on physical infrastructure and influences communications, financial markets, transportation, industrial operations, and essential services. Its effects depend on the decisions it informs and the actions it enables.

A model without external tools has fewer opportunities for direct action, but people may still use its outputs to make consequential decisions. Connecting the same model to operational systems changes those opportunities and may change its risk profile. Both automated execution and influence through human operators must therefore be considered within the larger deployment, including its permissions, technical connections, and institutional safeguards \citep{leveson2012safer,shevlane2023model}.

\subsection{``AI Has No Body''}
\label{subsec:no_body}

Some AI deployments have no dedicated robotic body, while others are incorporated into machines that sense and act in the physical world \citep{bommasani2021foundation}. Embodiment is one channel of influence, alongside software interfaces, communication networks, organizations, and human operators.

Research on language-model interfaces illustrates how these channels can work. ReAct combines generated reasoning traces with task-specific actions in the environments studied, while Toolformer trains models to select API calls and use their results in subsequent token prediction \citep{yao2023react,schick2023toolformer}. These findings demonstrate particular forms of tool use under tested conditions. They do not establish unrestricted autonomy or reliable operation of critical infrastructure.

The physical consequences of a deployment depend on which actions its interfaces permit and how those actions are constrained. A robotic body can expand the available options, but consequential influence is possible without one.

\subsection{``Humans Will Remain in Control''}
\label{subsec:humans_in_control}

Humans build AI systems, provide their infrastructure, establish legal rules, and authorize deployment. These responsibilities give people important means of control. Whether those means remain effective depends on the practical arrangements surrounding their use.

When safety relies on human intervention, responsible actors need enough information to recognize a problem, enough time to respond, and a workable means of stopping or constraining the activity. Complexity, distribution, and dependence can weaken each of these conditions. A person may approve a decision without being able to evaluate it adequately, while an organization may retain shutdown authority but lack an acceptable way to continue essential services afterward.

Effective control need not require manual approval of every action. Bounded permissions, automated monitoring, and tested response procedures can reduce the burden on human supervisors. Their value depends on whether they preserve the ability to constrain consequential behavior when conditions change. Control is therefore a property of the wider sociotechnical arrangement, rather than something established by placing a human in the decision chain \citep{leveson2012safer}.

\subsection{``AI Extinction Scenarios Are Science Fiction''}
\label{subsec:science_fiction}

This objection identifies a real weakness in some discussions. Stories about future architectures, motives, and strategies can become more detailed than the evidence warrants. Exploratory scenarios remain useful when their assumptions are explicit, but they should not be presented as established forecasts.

Several mechanisms invoked in those scenarios have independent research foundations. Accident analysis and studies of infrastructure interdependence examine how local failures can spread and how interactions can undermine safety \citep{perrow1999normal,rinaldi2001interdependencies,buldyrev2010cascade}. Their findings support investigation of particular propagation mechanisms. They do not establish that an AI-mediated cascade would cause extinction or permanently prevent civilizational recovery.

A catastrophic pathway must therefore be examined through separate questions. What could initiate the failure? Which dependencies could transmit it? What barriers would limit its spread? How could the resulting disruption affect surviving populations and institutions? What would make preservation or restoration permanently unavailable?

Each step requires its own evidence. The claim that a failure can propagate is substantially weaker than the claim that humanity could never recover from its consequences. Prolonged disruption alone does not establish permanence. Keeping these propositions distinct allows research to assess the pathway without treating the whole narrative as a single prediction.

\subsection{``Superintelligence Is Required''}
\label{subsec:superintelligence_required}

Some loss-of-control scenarios depend on capabilities far beyond those of present systems. Other systemic risks concern specialized abilities, deployment scale, operational access, and institutional dependence. A system need not outperform humans across all cognitive domains to occupy a consequential role.

Shared models and common dependencies can also produce correlated errors across deployments. Whether these errors remain local or become systemic depends on how the deployments interact and which safeguards continue to function. Foundation models used across many applications, including tool-based and agentic workflows, make these deployment relationships relevant to safety assessment \citep{bommasani2021foundation,bengio2026aisafety}.

This supports studying effective influence alongside cognitive performance. It does not demonstrate that current or moderately capable systems can produce an existential outcome. That stronger claim requires a credible pathway to a specified permanent endpoint, accounting for safeguards, surviving capacity, and opportunities for restoration. Depending on the endpoint, the pathway may involve either catastrophic disruption or a gradual and ultimately irreversible loss of meaningful human agency.

\subsection{``Institutions Will Adapt''}
\label{subsec:institutions_adapt}

Historical experience provides a substantial reason for caution about catastrophic predictions. Aviation, nuclear energy, medicine, finance, and industrial engineering have developed standards, regulatory institutions, professional practices, and procedures for managing hazards. AI development can likewise be shaped by institutional learning and deliberate safety measures.

The uncertainty concerns how effective that adaptation will be and when it will occur. Institutions can anticipate emerging risks and restrict dangerous deployments before harm becomes widespread. They can also struggle with incomplete information, limited resources, coordination difficulties, and incentives that favor rapid deployment \citep{bengio2026aisafety}. Historical adaptation establishes that improvement is possible, but does not determine whether safeguards will be adequate in a particular deployment.

The capability--control framework expresses one aspect of this problem. Within a specified context and using fixed, commensurate scales, the conceptual gap can widen when consequential capability grows faster than relevant control capacity. Institutions may improve while still falling behind changing demands. Conversely, matching the pace of capability growth does not ensure adequate protection if the starting safeguards are insufficient. The framework helps organize these questions without providing a calibrated measure of institutional effectiveness.

Taken together, the objections support a qualified conclusion:

\begin{quote}
\emph{Objections to existential AI risk expose uncertainty, challenge unsupported assumptions, and provide reasons to reject inevitability. Assessing the remaining risk still requires examining particular mechanisms of harm, the safeguards that could interrupt them, and the conditions under which their consequences could become permanent.}
\end{quote}

The paper does not predict that AI will eliminate humanity. It proposes a framework for examining possible routes to existential harm and identifying where those routes depend on evidence that remains incomplete. Systematic study is justified by plausible mechanisms and consequential unanswered questions; it does not require treating speculative endpoints as established outcomes.
\section{What Would Have to Go Wrong?}
\label{sec:what_wrong}

Advanced AI does not inevitably lead to human extinction or another existential outcome. A civilization-scale catastrophe requires a mechanism capable of producing sufficiently large effects and an inadequate response from the safeguards intended to prevent or contain them. An existential outcome requires something further: at least one of the permanent endpoints defined in Section~\ref{sec:definitions}, under the scope assumptions adopted there.

Several barriers can separate capability from irreversible harm. A poorly aligned system may remain constrained by its permissions. A dangerous capability may never be deployed. A serious failure may be contained, and even widespread catastrophe may leave enough people, knowledge, infrastructure, and institutional capacity to preserve a non-existential future.

Which barriers matter depends on the pathway. Autonomous misalignment, harmful human use, organizational failure, and competitive deployment involve different mechanisms and need not follow the same sequence. The task is to identify how a particular process could cause harm, which safeguards could interrupt it, and why those safeguards might prove insufficient. This section concentrates on acute or propagating catastrophic pathways; gradual permanent disempowerment may have a different structure and need not involve a discrete catastrophe.

\subsection{An Illustrative Barrier-Failure Sequence}
\label{subsec:necessary_chain}

One possible progression within an autonomous-misalignment pathway is

\begin{equation}
\boxed{
\begin{aligned}
&\text{misalignment}\\
&\quad\longrightarrow\text{inadequate evaluation}\\
&\quad\longrightarrow\text{inadequate containment}\\
&\quad\longrightarrow\text{governance or intervention failure}\\
&\quad\longrightarrow\text{civilization-scale catastrophe}\\
&\quad\longrightarrow\text{failure of preservation or recovery}.
\end{aligned}
}
\end{equation}

The sequence illustrates protective barriers rather than establishing necessary stages. Evaluation may correctly identify a danger while governance fails to prevent deployment. Containment and intervention may operate together, and some pathways may bypass individual barriers. Human misuse, for example, can begin with a system that follows its instructions accurately.

The arrows therefore represent a possible progression, not a claim that one failure inevitably causes the next. A sufficiently effective barrier can interrupt the pathway, provided it acts on the relevant mechanism before consequences exceed its reach.

\subsection{Failure of Technical Alignment}
\label{subsec:failure_alignment}

An autonomous-misalignment pathway may begin when a system's behavior fails to respect the objectives or constraints its designers intended. Hostility and consciousness are unnecessary: pursuing a specified objective can produce outcomes that its designers would reject if the specification omits important considerations \citep{amodei2016concrete,ngo2022alignment}.

Misalignment alone does not establish a civilization-scale danger. Limited permissions, restricted autonomy, and effective supervision may contain its consequences. The concern becomes more substantial when the behavior can affect important systems and other safeguards are inadequate.

Technical misalignment is also unnecessary for some catastrophic pathways. A harmful user may obtain precisely the assistance requested, while organizational mistakes or competitive pressures may make an otherwise functional system unsafe to deploy. These cases require attention to human objectives, institutional incentives, and operating conditions alongside model behavior.

\subsection{Failure of Evaluation}
\label{subsec:failure_evaluation}

Evaluation helps developers and responsible institutions understand what a system can do, how reliably it behaves, and how its capabilities change when combined with tools, memory, or greater autonomy. It can fail when tests omit relevant abilities, cover too narrow a range of conditions, or assess components separately when consequential behavior emerges from their combination.

Dangerous-capability evaluations address part of this problem \citep{shevlane2023model,bengio2026aisafety}. Their practical value depends on whether findings inform deployment decisions. A capability may remain undetected, its significance may be misunderstood, or accurate results may be overridden by other priorities.

Evaluation also matters beyond autonomous misalignment. A deployment assessed under restricted access may present different risks when connected to more powerful tools, offered to different users, or incorporated into another institutional setting. The relevant assessment must account for these changes. Evaluation provides evidence for decisions and safeguards; it cannot ensure safety if subsequent actions disregard that evidence.

\subsection{Failure of Containment}
\label{subsec:failure_containment}

Large external effects require a channel through which outputs or actions influence the world. Containment concerns the boundaries around those channels, including permissions, network access, tools, operational autonomy, and the use of outputs by people or other systems.

These boundaries may be inadequate without any technical ``escape.'' Operators can grant broad access because it makes a deployment more useful or convenient, while users can act on information produced through a text-only interface. A containment assessment must therefore consider both direct execution and indirect influence.

The same principle applies to misuse. Harmful users need some means of converting assistance into consequential activity, although that means may lie outside the AI deployment itself. Restrictions are effective when they address the relevant route to harm and remain enforceable as the system's role changes.

\subsection{Failure of Governance}
\label{subsec:failure_governance}

Governance shapes which systems are developed and deployed, what restrictions apply, and how identified hazards are handled. Its effectiveness depends on the decisions of companies, governments, regulators, research institutions, professional communities, and infrastructure operators.

Delay, fragmented responsibility, incomplete information, weak enforcement, and competitive pressure can undermine those decisions. An organization may recognize a risk yet deploy because it expects competitors to proceed. Safeguards adopted by one actor may also leave important exposure elsewhere, creating a need for compatible measures among relevant institutions \citep{han2019regulate,bengio2026aisafety}.

Governance failure need not mean institutional collapse. Oversight and authorization can remain operational while proving inadequate for the deployments they regulate. Such weaknesses may affect several pathways at once, increasing exposure to misuse, autonomous failures, organizational mistakes, and unsafe deployment practices.

\subsection{Failure of Intervention}
\label{subsec:failure_intervention}

Once hazardous activity begins, intervention may still prevent or limit harm. For responses that depend on recognizing a problem, warning signals must reach actors able to act, and the response must be effective within the available time. Automated limits and precautionary procedures can also interrupt activity before its causes are fully understood.

As discussed in Section~\ref{sec:switch_off}, practical control depends on more than formal authority. Distributed activity, incomplete information, fragmented responsibility, and dependence on essential services can make intervention difficult. Stopping one component may leave others operating, while delayed action may prevent additional harm without reversing effects that have already propagated.

Evaluation, containment, governance, monitoring, and intervention contribute to the preventive control architecture represented schematically by $W$ in Section~\ref{sec:formal_model}. They can prevent a catastrophe or limit an unfolding one. Their effectiveness must be assessed against the timing and structure of the particular hazardous process.

\subsection{Failure of Preservation or Recovery}
\label{subsec:failure_recovery}

A civilization-scale catastrophe is not automatically existential. Its longer-term consequences depend on surviving populations, knowledge, infrastructure, institutions, and the capacity to coordinate preservation or reconstruction.

For some endpoints, preservation matters more directly than rebuilding. Biological human extinction has not occurred at an assessment time if living humans remain, but their presence alone does not establish continued survival. Reproductive viability and the conditions supporting it matter for that prospect. Similarly, avoiding permanent disempowerment requires meaningful human agency to remain intact or to be restored under the criteria adopted in Section~\ref{sec:definitions}.

Let $\mathcal{C}$ denote the specified AI-mediated civilization-scale catastrophe, and let
$E=E_{\mathrm{bio}}\cup E_{\mathrm{civ}}\cup E_{\mathrm{agency}}$
denote the union of the three existential endpoint events. With all events defined over the same assessed trajectories, context, and temporal criteria, and assuming $P(\mathcal{C})>0$, define

\begin{equation}
Q_{\mathrm{cat}}
=
P(E^{c}\mid\mathcal{C}).
\end{equation}

This is the probability of avoiding all included existential endpoints, conditional on the catastrophe. It gives an explicit catastrophe conditioning to the preservation or recovery quantity $Q$ used earlier. Biological survival alone is insufficient if permanent civilizational loss or disempowerment nevertheless occurs.

The corresponding endpoint-specific probabilities are

\begin{equation}
\begin{aligned}
Q_{\mathrm{bio},\mathrm{cat}}
&=P(E_{\mathrm{bio}}^{c}\mid\mathcal{C}),\\
Q_{\mathrm{civ},\mathrm{cat}}
&=P(E_{\mathrm{civ}}^{c}\mid\mathcal{C}),\\
Q_{\mathrm{agency},\mathrm{cat}}
&=P(E_{\mathrm{agency}}^{c}\mid\mathcal{C}).
\end{aligned}
\end{equation}

Unlike the $Q_j$ quantities in Section~\ref{sec:definitions}, which condition on a hazardous process $F$, these quantities condition specifically on $\mathcal{C}$. Transferring between them requires either setting $F=\mathcal{C}$ under consistent definitions or specifying the relationship in an explicit model. Moreover, $Q_{\mathrm{cat}}$ concerns the joint event
$E_{\mathrm{bio}}^{c}\cap E_{\mathrm{civ}}^{c}\cap E_{\mathrm{agency}}^{c}$.
It cannot generally be calculated by multiplying the three endpoint-specific probabilities, because conditional independence has not been established.

Within a consistently specified assessment,

\begin{equation}
Q_{\mathrm{cat}}\rightarrow 0
\end{equation}

means that, conditional on the catastrophe, avoidance of every included endpoint becomes increasingly unlikely. The corresponding version of the earlier proxy $I=1-Q$ is

\begin{equation}
I_{\mathrm{cat}}
\equiv
1-Q_{\mathrm{cat}}
=
P(E\mid\mathcal{C}),
\end{equation}

so that

\begin{equation}
I_{\mathrm{cat}}\rightarrow 1.
\end{equation}

These are statements about conditional probabilities, not measurements of physical irreversibility or evidence that an endpoint has occurred. Establishing a realized endpoint requires examining the outcome and the mechanisms governing survival or restoration. Likewise, non-recovery over a finite observation period does not by itself establish permanent impossibility. If $P(\mathcal{C})=0$, the catastrophe branch contributes no probability, and these conditional quantities need not be defined.

Total existential risk includes outcomes outside this branch. For $P(\mathcal{C})>0$, the probability identity is
$P(E)=P(\mathcal{C})(1-Q_{\mathrm{cat}})+P(E\cap\mathcal{C}^{c})$.
The first term describes trajectories containing both the specified catastrophe and an existential endpoint; showing that the catastrophe causes the endpoint requires a further causal argument. The second includes existential outcomes without that catastrophe, potentially including gradual permanent disempowerment. A low $Q_{\mathrm{cat}}$ therefore does not determine total risk on its own.

The basic distinction remains

\begin{equation}
\text{advanced AI}
\not\Rightarrow
\text{existential catastrophe}.
\end{equation}

For the catastrophic pathways considered here, a possible progression is

\begin{equation}
\boxed{
\begin{aligned}
&\text{AI-mediated process capable of large effects}\\
&\quad+\text{failure of relevant pre-catastrophe safeguards}\\
&\quad\longrightarrow\text{civilization-scale catastrophe}\\
&\quad+\text{permanent failure of preservation or restoration}\\
&\quad\longrightarrow\text{existential outcome}.
\end{aligned}
}
\end{equation}

Permanent failure here concerns at least one specified existential endpoint, rather than the loss of an individual institution or technology. The arrows summarize a possible causal progression whose mechanisms still need to be demonstrated. Existential loss may occur during the catastrophe, rather than in a separate later stage.

The relevant barriers vary across pathways. Autonomous misalignment places particular demands on alignment, evaluation, containment, and intervention. Misuse may involve accurate instruction-following, while organizational and competitive pathways may depend more heavily on authorization, incentives, coordination, and deployment practices.

\begin{quote}
\emph{An existential conclusion requires more than identifying a dangerous capability or a severe failure. For a catastrophic pathway, the analysis must explain how harm reaches civilization scale, why relevant safeguards fail, and how the consequences produce at least one permanent existential endpoint.}
\end{quote}

This framework distinguishes prevention from preservation and recovery while recognizing their practical connections. Preparations made before a catastrophe can protect what survives and support later restoration. Intervention during disruption may also limit damage to that capacity.

The layers should not be assumed statistically independent. Shared infrastructure, institutions, information, models, or technical assumptions can cause several safeguards to fail together. Resources supporting prevention may also be needed for preservation and recovery. Assessing these dependencies is therefore part of understanding what would have to go wrong, and where protective measures could still interrupt the pathway.
\section{Preventing AI-Driven Existential Catastrophe}
\label{sec:prevention}

Because severe AI-mediated harm can arise through different pathways, prevention requires safeguards suited to the mechanisms involved. A technical solution, regulation, or institutional procedure may address an important vulnerability while leaving others unresolved. A credible safety architecture must therefore combine protections around the system, its deployment, and the institutions responsible for its use.

The aim is to reduce the likelihood that failures spread and to preserve the conditions for a non-existential future if severe disruption occurs. Evaluation, restrictions on autonomy and access, authorization, monitoring, governance, and intervention can prevent hazardous activity or contain its consequences. Preservation and recovery measures protect what survives and support restoration. Gradual permanent disempowerment requires additional attention to decision rights, institutional dependence, and the practical reversibility of delegation.

These functions overlap in practice, but their different purposes matter. Preventing a dangerous action, interrupting an unfolding cascade, and maintaining the capacity to recover each require their own evidence of effectiveness.

\subsection{Capability Evaluation}

Conventional benchmarks reveal performance on specified tasks, but may leave important safety questions unanswered. Evaluation should also examine capabilities that change the consequences of deployment, such as sustained planning, tool use, operation with reduced supervision, and performance under unfamiliar conditions.

The assessment must consider how these abilities interact with the tools, permissions, and support available in use. An improvement that appears modest on a general benchmark may become consequential in a particular operational setting. Evaluations designed for extreme AI risks address this connection between capability and deployment \citep{shevlane2023model,bengio2026aisafety}.

Assessment should continue as systems, users, and operating environments change. A deployment may acquire new tools or responsibilities after its initial evaluation, making earlier findings incomplete. Risk-management frameworks accordingly treat measurement and reassessment as continuing responsibilities throughout the lifecycle \citep{tabassi2023airmf}. The value of evaluation ultimately depends on whether its findings influence deployment decisions and the safeguards surrounding them.

\subsection{Limiting Autonomy}

A capable system can analyze problems and recommend actions without being authorized to execute every recommendation independently. Separating reasoning from execution provides an opportunity to retain useful performance while constraining particular routes to harm:

\begin{equation}
\text{high capability}
\not\Rightarrow
\text{high autonomy}.
\end{equation}

Operational authority should be justified by the application and its safeguards, rather than increased automatically as cognitive performance improves. Consequential decisions may require human or institutional authorization, bounded execution, or additional verification.

Such restrictions address risks arising from unsupervised action. Their effectiveness depends on whether they are enforced and whether reviewers can assess the proposed actions. They do not resolve every pathway: harmful users can exploit a system with little autonomy, and well-intentioned users can act on unreliable advice.

\subsection{Limiting Access}

Access determines which resources and operations a deployed system can use. Permissions can be tailored to the task, allowing information to be inspected without modified records, or permitting recommendations while reserving execution for separately authorized actors.

For actions with substantial or difficult-to-reverse consequences, narrower permissions and checks before execution may reduce exposure. The relevant boundaries must cover indirect channels as well: people and other software can convert outputs into actions even when the model has no direct operational access.

The objective is to grant access on the basis of demonstrated need and assessed consequences. Narrow permissions alone do not establish safety, while broader access may be justified where effective controls are available. Both judgments require evidence about the deployment and the pathways those permissions enable.

\subsection{Layered Authorization}

A single approval step may leave a consequential action dependent on one fallible judgment. Additional checks can improve protection when they examine relevant issues effectively and have sufficiently different failure modes. Depending on the application, these checks may involve human review, technical verification, organizational approval, or separate systems assessing different aspects of the decision.

Their number alone provides little assurance. Checks that share a model, data source, infrastructure, or mistaken assumption may fail together. Systems safety therefore requires attention to the interactions and shared vulnerabilities of protective measures \citep{leveson2012safer}.

A useful design principle is

\begin{equation}
\begin{aligned}
&\text{greater potential consequence}\\
&\qquad\longrightarrow
\text{stronger authorization with fewer shared failure modes}.
\end{aligned}
\end{equation}

This is a qualitative principle whose application must also account for urgency and the harm that delay could cause. Reviewers need sufficient information and a practical ability to refuse or interrupt the action. Strong benchmark performance does not, by itself, justify allowing an output to trigger a high-consequence operation.

\subsection{Monitoring and Auditing}

Deployment exposes systems to conditions that testing cannot fully reproduce. Monitoring can reveal unexpected behavior, changes in performance, misuse, and failures of safeguards. Auditing complements this work by making decisions and incidents available for reconstruction and examination, including by independent parties where appropriate.

Design intentions are insufficient evidence of compliance:

\begin{equation}
\text{intended compliance}
\not\Rightarrow
\text{actual compliance}.
\end{equation}

A safety assessment should draw on observed behavior alongside pre-deployment testing and analysis. Continuing measurement, review, and reassessment are consistent with lifecycle risk management \citep{tabassi2023airmf}.

Monitoring must connect warning signals to actors or mechanisms capable of responding. Detection without usable intervention authority may leave a hazard unaddressed. At the same time, enforced permissions and automatic containment can prevent some harmful actions without requiring every dangerous behavior to be identified individually.

\subsection{Preservation and Recovery Engineering}
\label{subsec:recovery_engineering}

Preservation and recovery capacity should be developed before a crisis. Even well-designed safeguards can fail, and unexpected combinations of failures may affect several protections at once.

Essential functions may require independent backups, alternative communications, replacement capacity, and procedures for operating under degraded conditions. Organizations also need workable ways to continue when AI-dependent systems become unavailable or unsafe. Resilience engineering addresses this broader capacity to anticipate disruption, withstand its effects, recover, and adapt \citep{ross2021cyberresilience}.

Let $\mathcal{C}$ denote the specified AI-mediated civilization-scale catastrophe, with $p_{\mathrm{cat}}=P(\mathcal{C})$, and let
$E=E_{\mathrm{bio}}\cup E_{\mathrm{civ}}\cup E_{\mathrm{agency}}$
denote the existential endpoint events defined in Section~\ref{sec:definitions}. Using consistent trajectory, deployment, and temporal criteria, and assuming $P(\mathcal{C})>0$, define

\begin{equation}
\begin{aligned}
Q_{\mathrm{cat}}
&=
P(E^{c}\mid\mathcal{C})\\
&=
P\!\left(
E_{\mathrm{bio}}^{c}
\cap E_{\mathrm{civ}}^{c}
\cap E_{\mathrm{agency}}^{c}
\mid\mathcal{C}
\right).
\end{aligned}
\end{equation}

For a fixed catastrophe assessment, prevention and resilience pursue complementary objectives:

\begin{equation}
\boxed{
\begin{aligned}
&\text{reduce }p_{\mathrm{cat}}\\
&\text{and improve endpoint avoidance given }\mathcal{C}.
\end{aligned}
}
\end{equation}

The second objective requires continued biological human existence, sufficient civilizational capacity, and meaningful human agency to be preserved or, where applicable, restored. Biological extinction must be prevented; rebuilding institutions cannot reverse it under the definition adopted here. A surviving population at one time also does not establish its continued survival. Avoidance of the aggregate existential event requires avoiding every included endpoint, without assuming that their probabilities are independent.

These objectives must be assessed within the complete probability accounting,
$P(E)=p_{\mathrm{cat}}(1-Q_{\mathrm{cat}})+P(E\cap\mathcal{C}^{c})$.
The final term includes existential outcomes without the specified catastrophe, potentially including gradual permanent disempowerment. This identity describes event probabilities; attributing an endpoint causally to the catastrophe requires further analysis.

Interventions may also change which catastrophes remain possible. If a measure preferentially prevents more recoverable events, $Q_{\mathrm{cat}}$ can decrease among the remaining catastrophes even while total existential risk falls. Conditional recovery probabilities should therefore inform a complete assessment rather than serve as an isolated measure of success. If the catastrophe has zero probability under the assessment, its branch contributes nothing and the conditional quantity need not be defined.

Evidence from recovery exercises and local incidents can help assess particular mechanisms, but extrapolation to civilization-scale disruption requires assumptions about surviving resources, shared dependencies, and the feasibility of restoration. Recovery by a specified date must also be distinguished from avoidance of a permanent endpoint.

For gradual disempowerment, protective measures may need to act well before a crisis. Retained expertise, contestable decision rights, independent infrastructure, and workable alternatives can preserve the possibility of revising delegated authority. Their purpose is to keep meaningful human agency practically recoverable as dependence grows.

\subsection{International Coordination}

AI deployments can cross national and organizational boundaries, making some hazards difficult for a single actor to manage. Competitive incentives can compound the problem when institutions favor stronger safeguards but fear losing an advantage to others that proceed more quickly.

Shared evaluation practices, technical standards, incident reporting, and information exchange may help address these collective-action problems \citep{bengio2026aisafety}. Their effectiveness depends on participation, implementation, and whether they cover the actors and activities relevant to the hazard.

Coordination does not require identical regulation everywhere, but it does require attention to gaps created by cross-border development, deployment, and use. It supports several functions simultaneously, including evaluation, authorization, monitoring, incident response, and preservation or recovery. It is therefore a continuing feature of the safety architecture rather than a final step in a defensive sequence.

\subsection{A Layered Safety Architecture}
\label{subsec:layered_safety_architecture}

The measures discussed above form interacting layers. One schematic representation of preventive control is

\begin{equation}
\boxed{
\begin{aligned}
\text{evaluate}
&\rightarrow
\text{limit autonomy and access}\\
&\rightarrow
\text{authorize}
\rightarrow
\text{monitor}
\rightarrow
\text{intervene}.
\end{aligned}
}
\end{equation}

Governance and coordination support these functions across stages. Where severe disruption occurs, preservation and recovery become especially important:

\begin{equation}
\boxed{
\begin{aligned}
&\text{preserve essential functions}\\
&\qquad\rightarrow
\text{stabilize}
\rightarrow
\text{recover or restore}.
\end{aligned}
}
\end{equation}

The diagrams organize functions rather than prescribe universal sequences. Evaluation and monitoring repeat, intervention can limit an unfolding catastrophe, and preservation may operate throughout it. Recovery preparations must precede disruption, while the same institutions and resources may support several layers.

Each layer has limitations. Evaluations can miss capabilities, access controls can fail, authorization can become superficial, and monitoring can overlook important signals. Recovery arrangements may depend on the infrastructure they are intended to replace. Additional layers improve protection only when they address relevant failure modes and avoid unnecessary shared vulnerabilities. Independence must be investigated rather than inferred from the number of checks or institutions involved.

\begin{quote}
\emph{Reducing AI-mediated existential risk requires safeguards that address the relevant pathways to harm, remain usable when failures occur, and preserve the conditions for avoiding permanent existential endpoints. Layering can strengthen this architecture, but its effectiveness depends on the protections provided and the vulnerabilities they share.}
\end{quote}

For gradual disempowerment, this means preserving meaningful human agency through decision rights, institutional adaptability, fallback capacity, and practical mechanisms for reversing excessive delegation. Across pathways, the objective is a sociotechnical system that can evaluate consequential behavior, constrain and authorize action, interrupt hazards, and preserve or restore the conditions for a non-existential future.
\section{From Control to Human--AI Coexistence}
\label{sec:coexistence}

Effective control remains central to AI safety, but direct supervision of individual actions provides an incomplete basis for long-term protection. As AI-mediated activity becomes faster, more extensive, and more deeply embedded in society, people may increasingly govern the conditions under which decisions are made rather than review every decision themselves.

The broader challenge is to preserve beneficial interaction and meaningful human agency as capabilities, incentives, institutions, and authority change. Here, \emph{coexistence} describes an enduring sociotechnical arrangement in which AI-mediated activity remains subject to effective correction, restriction, and replacement. The term does not require treating AI systems as persons or attributing independent interests to them.

\subsection{Control Is Not the Final Objective}
\label{subsec:control_not_final}

Human supervision takes several forms. People define objectives, evaluate outputs, grant permissions, and respond to problems, while access restrictions and automated checks can constrain activity without requiring individual review. Their effectiveness depends on the deployment and the consequences of failure.

As automation expands, meaningful agency may depend less on approving each action and more on retaining effective authority over objectives, operating boundaries, and intervention procedures. Delegation can remain accountable when people have the information, competence, resources, and authority needed to challenge decisions and revise or terminate unacceptable arrangements.

The longer-term objective is therefore beneficial and acceptably safe interaction without continuous micromanagement. Control remains essential to that objective: it helps maintain the conditions under which delegated activity can be scrutinized, constrained, and changed.

\subsection{Cooperative Stability}
\label{subsec:cooperative_equilibrium}

A durable arrangement would keep AI-mediated activity compatible with specified human interests and safety requirements as conditions change. Technical safeguards and institutional rules would need to constrain harmful behavior, address conflicts, and sustain beneficial uses.

Research on Cooperative AI examines cooperation involving humans and machine agents through game theory, multi-agent systems, social choice, and human--machine interaction \citep{dafoe2020cooperative}. Cooperation in this context can describe patterns produced by objectives, incentives, permissions, and rules, without implying goodwill or conscious intention. It also requires qualification: coordinated activity can serve harmful purposes. The objective here is cooperation consistent with meaningful human agency and the safety criteria adopted in this paper.

The relevant dependencies can be organized schematically as

\begin{equation}
\mathcal{S}_{\mathrm{coop}}
=
\Psi
\left(
H_{\mathrm{inc}},
A_{\mathrm{comp}},
I_{\mathrm{inst}},
R_{\mathrm{auth}}
\right),
\end{equation}

where $H_{\mathrm{inc}}$ represents relevant human incentives and preferences, $A_{\mathrm{comp}}$ the compatibility of AI-mediated behavior with specified human interests and safety criteria, $I_{\mathrm{inst}}$ the institutional environment, and $R_{\mathrm{auth}}$ the distribution of resources, decision rights, and authority.

These categories can interact and need not exhaust the determinants of coexistence. Human interests may also conflict. Any application must specify whose interests are considered, how disagreements are handled, and which outcomes are unacceptable.

The expression provides a conceptual organization of these dependencies. Neither $\Psi$ nor an empirical scale for $\mathcal{S}_{\mathrm{coop}}$ is defined, so it establishes no calibrated index, causal model, or game-theoretic equilibrium. Its notation remains distinct from the capability--control gap $G(t)$ and the irreversibility proxy $I$ in Section~\ref{sec:formal_model}.

Robustness would concern whether an arrangement remains within acceptable conditions under specified changes. Resilience would additionally concern its capacity to preserve or restore essential functions and meaningful human agency following disruption. Both require evidence about the arrangement and the conditions being assessed.

\subsection{Stability Under Changing Conditions}
\label{subsec:evolutionary_stability}

Evolutionary game theory offers a limited analogy. An evolutionarily stable strategy resists invasion by sufficiently rare alternatives under the assumptions of a specified game and population model \citep{maynardsmith1973logic}. The present framework specifies no payoffs, population dynamics, or invasion criterion, and evolutionary stability within a fixed game does not establish robustness when the game itself changes.

The useful lesson is to examine an arrangement in the presence of alternatives and disturbances. A human--AI system that functions only while its initial incentives, capabilities, and institutions remain unchanged may be fragile. Assessing coexistence requires considering new systems, changing economic incentives, institutional responses, and shifts in practical authority. These changes can strengthen safeguards or undermine them; their effects must be investigated.

The initial question is

\begin{equation}
\begin{gathered}
\text{Does cooperation meet the specified safety and agency criteria}\\
\text{under present conditions?}
\end{gathered}
\end{equation}

The broader question is

\begin{equation}
\begin{gathered}
\text{Does cooperation continue to meet those criteria}\\
\text{under specified changes in the surrounding system?}
\end{gathered}
\end{equation}

This extends the range of assessment without demanding stability under every conceivable disturbance. Successful testing over a finite range provides evidence within that range, rather than proof of permanent stability.

\subsection{Testing Coexistence}
\label{subsec:testing_coexistence}

Evidence of beneficial interaction concerns particular systems, users, permissions, institutions, and observation periods. Applying it to substantially different arrangements requires an account of which findings are transferable and why.

An assessment might examine whether greater autonomy weakens oversight, whether institutional dependence reduces replacement options, or whether changes in resources and authority make consequential decisions harder to contest. It should also examine whether improved safeguards and institutional adaptation can preserve agency under those changes.

A minimally acceptable condition can be represented conceptually as

\begin{equation}
\mathcal{S}_{\mathrm{coop}}(t)
>
\mathcal{S}_{\mathrm{min}},
\end{equation}

while relevant properties change over time:

\begin{equation}
C_{\mathrm{AI}}(t),\;
C_{\mathrm{human}}(t),\;
I_{\mathrm{inst}}(t),\;
R_{\mathrm{auth}}(t).
\end{equation}

Here, $C_{\mathrm{AI}}(t)$ and $C_{\mathrm{human}}(t)$ refer to capabilities relevant to the assessed interaction, without assuming a universal scale for comparing humans and AI. The boundary $\mathcal{S}_{\mathrm{min}}$ represents acceptable conditions under the chosen criteria. The strict inequality is a notational convention, not an empirical threshold.

Because no scale is specified, the inequality is not directly testable. Empirical work requires operational definitions, observable indicators, specified disturbances, observation periods, and criteria for unacceptable losses of safety or agency. Separate measures may be more informative than a single score: economic benefit, oversight, contestable authority, and recovery capacity can change in different directions.

Experiments, simulations, and deployment studies could examine whether supervisors retain effective intervention options, safeguards respond to identified failures, and institutions can revise or replace consequential deployments. Such evidence would constrain particular claims about coexistence within the scope of the study. Extending those conclusions to future technologies or indefinite periods would require further justification.

\subsection{From Alignment to Coevolution}
\label{subsec:alignment_to_coevolution}

Alignment concerns whether AI behavior remains compatible with the objectives, constraints, or values specified in an assessment. Long-term safety also depends on changes in the environment where systems are developed and used.

AI deployment can alter work, markets, decision-making, and institutional practices. Those changes can influence demand for new systems, training priorities, deployment incentives, and governance. A possible feedback sequence is

\begin{equation}
\begin{aligned}
\text{human institutions}
&\rightarrow \text{AI development}\\
&\rightarrow \text{social and economic change}\\
&\rightarrow \text{institutional adaptation}\\
&\rightarrow \text{new AI development}.
\end{aligned}
\end{equation}

The arrows describe possible influences across successive periods. They do not prescribe a necessary order or a complete causal model. Institutions can shape several stages simultaneously, and feedback may improve protection, introduce vulnerabilities, or produce mixed effects.

In this limited sense, \emph{coevolution} describes reciprocal institutional--technological adaptation. It makes no claim about biological populations or Darwinian dynamics. Its relevance to safety lies in recognizing that objectives, deployment conditions, institutions, and human expectations can change together. Protective arrangements must be assessed within that changing environment.

\begin{quote}
\emph{Evidence for safe human--AI coexistence should address beneficial interaction under present conditions and the preservation of meaningful human agency under specified changes and disruptions. Durable safety requires practical mechanisms for correcting, constraining, or replacing arrangements that cease to meet those criteria.}
\end{quote}

This perspective complements alignment and control. Evaluation, authorization, access restrictions, monitoring, and intervention address immediate hazards, while institutional adaptation and preservation or recovery capacity help sustain protection through change and disruption.

The proposed objective is an enduring arrangement in which beneficial AI-mediated activity remains compatible with meaningful human agency. Whether a deployment achieves that objective requires explicit criteria and evidence about its operation, its vulnerabilities, and the practical options available when correction becomes necessary.
\section{Discussion}
\label{sec:discussion}

The analysis supports a cautious conclusion. Several mechanisms relevant to AI risk can already be studied empirically, but their connection to biological extinction, permanent civilizational loss, or irreversible human disempowerment remains uncertain. Evidence that a mechanism exists must be distinguished from evidence that it can produce a specified existential endpoint.

AI systems can use tools, perform sequences of actions with varying supervision, and operate within complex technical and institutional settings. Research on accidents and interdependent networks also identifies mechanisms through which failures can propagate \citep{perrow1999normal,buldyrev2010cascade}. Applying these findings to existential AI risk requires an account of the dependencies involved, the barriers that could limit escalation, and the conditions supporting survival or restoration.

The unresolved question is whether an AI-mediated process could eliminate living humanity, permanently destroy sufficient civilizational capacity, or make meaningful human agency unrecoverable. Severe disruption alone establishes none of these outcomes. Research must therefore distinguish observed mechanisms, justified extrapolations, and speculative transitions, making clear what evidence supports each part of the argument.

\subsection{What Is Established and What Remains Speculative?}
\label{subsec:established_speculative}

AI performance has improved on many measured tasks, and some deployments can undertake extended sequences of tool-assisted action. Progress remains uneven, however. Success in one setting does not establish reliability in another, and practical consequences depend on the surrounding software, permissions, resources, and users \citep{bengio2026aisafety}.

Evaluation provides incomplete coverage of those conditions. Benchmarks sample particular tasks, while deployment can introduce new incentives, tools, and combinations of components. Assessing whether a system possesses a dangerous capability also differs from assessing its propensity to use that capability harmfully \citep{shevlane2023model}. These limitations justify further testing; they do not demonstrate the existence of an unobserved capability or establish an existential pathway.

The greatest uncertainty concerns the transitions to permanent outcomes. Future autonomy, delegated authority, institutional dependence, and the effectiveness of safeguards remain open questions. So do the circumstances under which severe disruption could remove the conditions for continued human existence or eventual restoration of civilizational capacity and agency.

Three epistemic levels help organize the discussion:

\begin{equation}
\begin{aligned}
\text{observed mechanisms}
&\rightarrow \text{plausible extrapolations}\\
&\rightarrow \text{existential outcomes}.
\end{aligned}
\end{equation}

The arrows indicate increasing inferential distance from direct evidence, rather than an established causal progression. Each transition requires additional assumptions and argument. Intermediate mechanisms can be investigated without experimentally observing an existential endpoint, but the resulting evidence has limits that must remain explicit.

\subsection{Which Assumptions Matter Most?}
\label{subsec:key_assumptions}

Some assumptions are essential to a pathway, while others affect its likelihood or severity without changing its basic structure. Research should identify these differences and ask which observations could strengthen or weaken the proposed causal links.

Relevant assumptions concern autonomy, access, delegated authority, harmful human use, oversight, containment, institutional response, and preservation or recovery capacity. Their importance varies across autonomous misalignment, misuse, organizational failure, and competitive deployment. A useful comparison between assessments should therefore examine their underlying pathways before comparing their final probabilities.

The framework in Section~\ref{sec:formal_model} separates a specified civilization-scale catastrophe from the existential endpoints that may accompany or follow it. Let $\mathcal{C}$ denote the catastrophe, $p_{\mathrm{cat}}=P(\mathcal{C})$, and $E$ the union of the three endpoint events. All events must refer to the same assessed trajectories, deployment context, and temporal criteria. For $p_{\mathrm{cat}}>0$,

\begin{equation}
p_{\mathrm{existential}}
=
P(E)
=
p_{\mathrm{cat}}(1-Q_{\mathrm{cat}})
+
P(E\cap\mathcal{C}^{c}),
\end{equation}

where

\begin{equation}
Q_{\mathrm{cat}}
=
P(E^{c}\mid\mathcal{C}).
\end{equation}

Here, $Q_{\mathrm{cat}}$ concerns avoidance of every included existential endpoint. Continued human existence, sufficient civilizational capacity, and meaningful agency must all be preserved or, where applicable, restored under the adopted criteria. Biological survival alone does not establish avoidance of permanent civilizational loss or disempowerment.

The first term equals $P(E\cap\mathcal{C})$: it describes trajectories containing both events. Establishing that the catastrophe contributes causally to the endpoint requires further analysis. The second term includes existential outcomes without the specified catastrophe, potentially including gradual permanent disempowerment. It can be omitted only under the explicit assumption that its probability is zero. If $p_{\mathrm{cat}}=0$, only the second branch remains, and $Q_{\mathrm{cat}}$ need not be defined.

These are probability identities, requiring no independence assumption and supplying no numerical estimate. The aggregate avoidance probability also cannot generally be obtained by multiplying endpoint-specific probabilities, since the endpoints may overlap or share causes.

Assessment must account for how interventions change the distribution of possible failures. A change in $Q_{\mathrm{cat}}$ can reflect a different mixture of remaining catastrophes rather than a change in total safety. Conditional recoverability should therefore be interpreted within the full accounting.

Where pathway-specific models become operational and empirically constrained, sensitivity analysis could identify assumptions that strongly affect their estimates. The present framework provides neither calibrated sensitivities nor a ranking of the assumptions dominating total risk. Its contribution is to make the questions and distinctions needed for such analysis explicit.

\subsection{Making Existential Risk Empirically Testable}
\label{subsec:empirical_xrisk}

Human extinction cannot serve as a controlled experimental endpoint, and failure to recover during a finite observation period does not establish permanent loss. Empirical work must focus on mechanisms and intermediate transitions that can be observed under specified conditions.

Relevant targets include the reliability and duration of autonomous operation, timely detection of hazardous behavior, the effectiveness of intervention, changes associated with tools and permissions, institutional response times, and preservation or restoration of essential functions after disruption. Instead of asking only

\begin{equation}
P(\text{AI causes human extinction}) = ?,
\end{equation}

research can examine narrower quantities such as

\begin{equation}
P\!\left(
\text{timely detection}
\,\middle|\,
\text{specified dangerous behavior}
\right),
\end{equation}

\begin{equation}
P\!\left(
\text{successful intervention}
\,\middle|\,
\text{danger detected in time}
\right),
\end{equation}

and

\begin{equation}
P\!\left(
\begin{gathered}
\text{successful preservation}\\
\text{or recovery}
\end{gathered}
\,\middle|\,
\begin{gathered}
\text{specified AI-mediated}\\
\text{disruption}
\end{gathered}
\right).
\end{equation}

Estimating these probabilities requires a task distribution, deployment configuration, observation period, and success criterion. Detection should count as timely only while an effective response remains possible. False alarms, missed detections, and cases where intervention cannot be attempted also matter. An association between intervention and a favorable outcome does not by itself establish that the intervention prevented harm.

The measurements remain context-dependent. Findings from current systems may not transfer to more capable deployments, and recovery rates from local incidents or simulations do not directly estimate $Q_{\mathrm{cat}}$ for civilization-scale disruption. They can constrain parts of a model when the assumptions needed for extrapolation are stated. Recovery within a specified period must also remain distinct from avoidance of a permanent endpoint.

Pathway-specific studies can examine how agent architecture changes autonomous performance, how access alters the consequences of errors, how AI assists different users, and how organizational safeguards perform under pressure. These studies assess the interaction between deployed systems and their environments, extending beyond model performance in isolation.

Measurements of evaluation, monitoring, intervention, capability, and institutional response can inform the capability--control analysis in a specified context. They do not provide a validated numerical decomposition of the gap. Here, $C_{\mathrm{control}}$ concerns control before or during escalation to catastrophe, while preservation and recovery are represented separately through $Q_{\mathrm{cat}}$ and endpoint-specific avoidance probabilities. Shared institutions and infrastructure can support both functions, so their analytical separation implies no causal independence.

Structured evaluation and continuing risk management provide a basis for revising intermediate claims as evidence accumulates \citep{tabassi2023airmf,shevlane2023model}. The final existential-risk estimate still depends on additional assumptions about future conditions and unobserved transitions.

\subsection{Extinction Probability Is Not the Only Relevant Metric}
\label{subsec:not_only_probability}

A single extinction probability can be informative when its time horizon, deployment assumptions, evidence, and endpoint definition are clear. Without that context, apparent precision can obscure substantial differences between assessments. Estimates may disagree because they concern different circumstances or make different assumptions about autonomy, misuse, safeguards, institutions, and recovery.

Risk analysis also considers consequences. To express the familiar probability--loss relationship precisely, let $L$ be a nonnegative random loss that is zero outside a specified event $E$ and has a finite expectation. For $P(E)>0$,

\begin{equation}
\operatorname{E}[L]
=
P(E)\operatorname{E}[L\mid E].
\end{equation}

The second factor is the conditional mean loss. When loss is constant on $E$, the expression reduces to $P(E)L_E$. If the assessment includes losses outside $E$, their contribution must be added.

This identity does not assign an agreed value to humanity's future. Quantifying existential loss requires explicit choices about consequences, values, and their distribution. The equation distinguishes likelihood from severity without calibrating either.

The endpoints considered in this paper extend beyond immediate damage. Extinction eliminates living humanity, while permanent civilizational or agency loss removes capacities needed for humans to shape their future. Severe disruption that leaves sufficient survival and restoration possible need not meet those definitions. This concern with future potential is central to the existential-risk literature \citep{bostrom2013existential,ord2020precipice}.

It does not follow that every imaginable scenario deserves equal priority. Evidence quality, causal plausibility, competing risks, and opportunity costs remain relevant. Even where a terminal probability cannot be estimated precisely, research can ask which assumptions matter, what evidence would change the assessment, and which interventions remain effective across plausible conditions.

\subsection{The Key Research Question}
\label{subsec:key_research_question}

AI capabilities can support beneficial uses while changing the consequences of error, misuse, unsafe deployment, and institutional dependence. Their significance depends on how they acquire practical influence. Increasing capability alone therefore supplies no universal boundary between useful assistance and existential danger.

The research task is to examine how capability interacts with autonomy, access, persistence, human use, and institutional arrangements. It must also assess whether evaluation, governance, and intervention remain adequate, and what preservation or restoration would be possible if preventive control failed.

One desirable aim is increasing social benefit alongside a non-widening capability--control gap:

\begin{equation}
\Delta B > 0,
\qquad
\frac{dG}{dt}\leq 0,
\end{equation}

where $\Delta B$ denotes a favorable change under a specified assessment of social benefit, and

\begin{equation}
G(t)
=
C_{\mathrm{AI}}(t)
-
C_{\mathrm{control}}(t).
\end{equation}

The benefit criterion requires choices about consequences and their distribution; it is not assumed to share a scale with capability or control. The numerical gap requires fixed, commensurate definitions of $C_{\mathrm{AI}}$ and $C_{\mathrm{control}}$ for the assessed context. Without those definitions, it remains conceptual.

Where a differentiable approximation is appropriate, a non-widening gap is equivalent within this representation to

\begin{equation}
\frac{dC_{\mathrm{control}}}{dt}
\geq
\frac{dC_{\mathrm{AI}}}{dt}.
\end{equation}

This condition does not establish safety. A stable or narrowing gap can coexist with inadequate initial safeguards, while a widening gap may remain within an adequate margin for some period. Both the starting conditions and subsequent changes matter.

A widening gap is represented by

\begin{equation}
\frac{dG}{dt}>0,
\end{equation}

or, under the same assumptions,

\begin{equation}
\frac{dC_{\mathrm{AI}}}{dt}
>
\frac{dC_{\mathrm{control}}}{dt}.
\end{equation}

Such a change may make particular pathways harder to manage. It is neither a catastrophe nor a sufficient condition for an existential outcome. Its importance depends on permitted actions, existing safeguards, and the mechanisms through which failures could spread or become permanent.

Different domains may have different transitions. The same capability can have different consequences depending on connected infrastructure, available users, delegated authority, and practical intervention options. Benefits and serious risks can also coexist within a deployment.

\begin{quote}
\emph{Under what conditions does increasing AI capability, combined with autonomy, access, human use, and institutional dependence, contribute to processes that exceed effective human control or permanently undermine the conditions for continued human existence, sufficient civilizational capacity, or meaningful agency?}
\end{quote}

Progress does not require a precise forecast of future architectures or an agreed extinction probability today. It requires evidence about how capabilities become consequential influence, how hazards arise and propagate, and which safeguards preserve intervention and restoration options. Reporting each finding with its scope and assumptions allows broader risk arguments to be revised through smaller causal questions that can increasingly be tested.
\section{Conclusions}
\label{sec:conclusions}

The central argument of this paper is that existential AI risk need not depend on consciousness, hostility, or a deliberate intention to harm humanity. Poorly specified objectives, harmful human use, excessive autonomy, weak oversight, institutional dependence, competitive deployment, and interconnected failures provide possible mechanisms of severe harm. Establishing an existential pathway requires a further argument: how those mechanisms could reach a permanent endpoint defined in Section~\ref{sec:definitions}, despite safeguards and remaining opportunities for survival or restoration.

Advanced AI therefore does not imply human extinction. Cognitive capability, operational influence, and failure of preventive control are distinct from biological extinction, permanent civilizational loss, and irreversible human disempowerment. A severe catastrophe may leave recovery possible, while permanent loss of agency could develop without an acute catastrophe. Assessment must specify the endpoint, its temporal criteria, and the assumptions connecting it to humanity's future.

There is no universal sequence from capability to existential harm. Autonomous misalignment may involve interactions among capability, autonomy, access, persistence, strategic competence, and inadequate safeguards. These dimensions need not all be present: a single consequential action may initiate a process that continues without further AI involvement. Misuse can involve accurate instruction-following, while organizational and competitive pathways depend on responsibility, incentives, coordination, authorization, and deployment decisions.

The paper distinguishes two broad forms of possible transition:

\begin{equation}
\boxed{
\begin{aligned}
\text{AI-mediated pathways}
&\longrightarrow
\left\{
\begin{array}{l}
\text{catastrophe-mediated route},\\
\text{gradual loss-of-agency route}.
\end{array}
\right.
\end{aligned}
}
\end{equation}

These categories are neither exhaustive nor mutually exclusive. Institutional dependence can weaken protection against catastrophe, and catastrophic disruption can accelerate disempowerment. The arrows describe possibilities requiring causal investigation.

An illustrative catastrophe-mediated pathway is

\begin{equation}
\boxed{
\begin{aligned}
&\text{AI-mediated hazardous process}\\
&\quad\longrightarrow
\text{civilization-scale catastrophe}\\
&\quad\longrightarrow
\text{permanent loss of at least one}\\
&\qquad\phantom{\longrightarrow}
\text{condition required for a non-existential future}.
\end{aligned}
}
\end{equation}

The final condition concerns continued human existence, sufficient civilizational capacity, or meaningful agency. Under the adopted classification, permanent loss of any one is sufficient; all three need not disappear. The endpoint may be reached during the catastrophe rather than in a separate later stage.

Permanent disempowerment may instead follow a progression such as

\begin{equation}
\boxed{
\begin{aligned}
&\text{progressive delegation and dependence}\\
&\quad\longrightarrow
\text{erosion of meaningful human agency}\\
&\quad\longrightarrow
\text{loss of realistic restoration mechanisms}\\
&\quad\longrightarrow
\text{permanent human disempowerment}.
\end{aligned}
}
\end{equation}

Delegation alone does not establish this outcome. People may retain effective decision rights, replacement options, and institutions through which arrangements can be revised. An existential argument must explain how those options become permanently unavailable. Difficulty in restoring agency, or its absence over a finite period, does not establish permanence.

Probability accounting must retain these distinctions. Let $\mathcal{C}$ denote the specified AI-mediated civilization-scale catastrophe, with $p_{\mathrm{cat}}=P(\mathcal{C})$, and let
$E=E_{\mathrm{bio}}\cup E_{\mathrm{civ}}\cup E_{\mathrm{agency}}$
denote the union of the assessed endpoints. With consistent trajectory, deployment, and temporal criteria, and assuming $P(\mathcal{C})>0$,

\begin{equation}
p_{\mathrm{existential,cat}}
=
P(E\cap\mathcal{C})
=
p_{\mathrm{cat}}\left(1-Q_{\mathrm{cat}}\right),
\end{equation}

where

\begin{equation}
\begin{aligned}
Q_{\mathrm{cat}}
&=
P(E^{c}\mid\mathcal{C})\\
&=
P\!\left(
E_{\mathrm{bio}}^{c}
\cap E_{\mathrm{civ}}^{c}
\cap E_{\mathrm{agency}}^{c}
\mid\mathcal{C}
\right).
\end{aligned}
\end{equation}

Avoiding the aggregate existential event requires avoiding every included endpoint. Continued human existence, sufficient civilizational capacity, and meaningful agency must all be preserved or, where applicable, restored. Biological survival alone is insufficient, and endpoint-specific avoidance probabilities cannot generally be multiplied because the events may overlap or share causes.

The catastrophe-branch expression is a probability identity, supplying neither a numerical estimate nor a causal demonstration. A catastrophe-mediated interpretation requires an account of how $\mathcal{C}$ contributes to $E$. Total risk also includes the branch without the specified catastrophe:
$P(E)=p_{\mathrm{cat}}(1-Q_{\mathrm{cat}})+P(E\cap\mathcal{C}^{c})$.
That branch cannot be omitted without justification. If $P(\mathcal{C})=0$, it contains the entire assessed risk and $Q_{\mathrm{cat}}$ need not be defined.

Prevention and resilience must consequently be assessed together. Comparisons between interventions should consider the full risk, since preventing some catastrophes changes the mixture of those that remain possible. Conditional recoverability alone can therefore give a misleading impression of an intervention's effect.

No universal capability threshold separating useful assistance from existential danger has been established here. Consequences depend on how capabilities acquire influence through tools, human reliance, institutional roles, resources, and delegated authority. The capability--control gap organizes these relationships conceptually. A numerical interpretation requires fixed, commensurate definitions, and the gap alone establishes neither inadequate control nor a reachable existential endpoint.

AI safety accordingly extends beyond technical alignment. Permissions, authorization, access restrictions, monitoring, institutional design, human decisions, international coordination, and infrastructure resilience address different parts of the problem. Their effectiveness must be examined within the deployed sociotechnical system. Limits on autonomy may constrain one pathway while leaving misuse possible; approval procedures may offer little protection if reviewers lack information, time, competence, or effective authority.

Preservation and recovery require advance preparation and usable capacity when disruption occurs. Their success depends on surviving populations, resources, knowledge, and institutions, rather than on the existence of a plan. They may share infrastructure with preventive control, making common vulnerabilities important across both functions. Protective layers should address different failure modes and reduce avoidable dependencies; adding checks does not itself establish adequate or independent protection.

Over longer periods, safety also requires attention to human--AI coexistence. Extensive automation may make direct supervision of every action impractical. Meaningful agency can instead rest on enforceable boundaries, contestable authority, institutional adaptation, and realistic options for correction and replacement. The framework proposed here identifies these requirements without establishing a formal equilibrium or a validated measure of long-term stability.

Substantial uncertainty remains. This paper provides neither calibrated existential-risk probabilities nor an empirically validated chain to a permanent endpoint. Its contribution is to separate mechanisms, expose assumptions, and distinguish evidence about intermediate failures from claims about terminal outcomes.

Research can make progress through narrower questions. Studies can examine how agent architecture and permissions affect autonomous performance, how tools and human use alter consequences, whether evaluations transfer to deployment, and whether hazards can be detected while intervention remains effective. Further work can assess shared vulnerabilities, institutional response, preservation of essential functions, restoration after specified disruptions, and the practical reversibility of delegated authority.

Results should identify their task distributions, deployment configurations, comparison conditions, success criteria, and observation periods. Evidence from current systems, local incidents, and simulations can constrain parts of a pathway, but extending it to future systems or civilization-scale disruption requires explicit assumptions. Finite observations of failure or non-recovery cannot establish permanent loss on their own.

\begin{quote}
\emph{A claim of existential AI danger requires a credible account of how an AI-mediated process could reach biological human extinction, irreversible loss of sufficient civilizational capacity, or permanent loss of meaningful human agency. The argument must explain why relevant safeguards and remaining opportunities for survival or restoration would fail to prevent that endpoint.}
\end{quote}

The framework supports investigation without predicting inevitable catastrophe. Its practical purpose is to identify how capabilities become consequential, which failures could spread, where safeguards could interrupt them, and what would preserve a non-existential future if prevention failed. Those questions provide a basis for assessments that can be examined and revised as evidence develops.

\section*{Declaration of AI Use}

The authors used ChatGPT to assist with language editing, \LaTeX formatting and references. The authors reviewed and revised the AI-assisted material and take full responsibility for the manuscript's arguments, accuracy, references, and conclusions. AI-generated suggestions were not treated as independent scientific evidence.

\clearpage
\phantomsection
\addcontentsline{toc}{section}{References}

\bibliographystyle{plainnat}
\bibliography{references}

\end{document}